\documentclass[12pt]{article}

\usepackage{amsmath,amssymb}
\usepackage{iftex}
\usepackage{algorithm}
\usepackage{algorithmic}
\usepackage{siunitx}
\usepackage{threeparttable}
\usepackage{caption}
\usepackage{multirow}
\usepackage{subcaption}
\usepackage[shortlabels]{enumitem}
\usepackage{newtxtext}
\usepackage{bm}
\usepackage{pgfplots}
\usepackage{xr-hyper}
\usepackage{setspace}
\usepackage{url}
\usepackage{booktabs}
\usepackage[margin=1in]{geometry}
\usepackage{natbib}
\usepackage{adjustbox}
\usepackage{bibunits}
\usepackage{tikz}
\usetikzlibrary{positioning,calc,arrows.meta,fit,matrix}
\usepackage{xcolor}
\usepackage{longtable,array,fvextra}
\pgfplotsset{compat=1.17}
\newtheorem{theorem}{Theorem}

\newtheorem{assumption}{Assumption}
\newtheorem{definition}{Definition}
\newtheorem{corollary}{Corollary}
\newcommand{\Pa}{\operatorname{Pa}}

\newcommand{\sig}{\sigma}

\title{Structure-Aware Transformers for Hierarchical Authorization in Personal AI Agents}

\author{
  Dongyu Cui\thanks{School of Statistics, University of Minnesota}
  \and
  Jie Ding\thanks{School of Statistics, University of Minnesota}
  \and
  Xuan Bi\thanks{Carlson School of Management, University of Minnesota, email: \texttt{xbi@umn.edu}}
}
\date{}

\begin{document}
\doublespacing
\maketitle

\begin{abstract}
\noindent Personal AI agents can carry out complex, multi-step tasks on users’ behalf, but doing so often involves numerous permission requests. Manually reviewing each request is time-consuming, while indiscriminate approval poses privacy and security risks. Predicting users’ authorization preferences offers a way to reduce this burden while preserving user control. However, authorization prediction is challenging because permission requests are often hierarchical and observed user decisions are sparse and potentially correlated. To address these challenges, We propose the Parent-gated Authorization Transformer for Hierarchies (PATH) to predict users’ authorization preferences for personal AI agents. PATH is a novel neural network architecture that combines hierarchy-aware attention masks with a parent-gated probability formulation to ensure hierarchy-consistent predictions. Hierarchy-level embeddings further support learning from sparse observations and patterns shared within each authorization level. We establish finite-sample risk bounds and provide sufficient conditions for consistent authorization probability estimation and algorithmic convergence. Experiments on simulated data, AI-generated user profiles, and real user authorization data demonstrate strong predictive performance with zero hierarchy violations.
\end{abstract}


\section{Introduction}
\label{sec:intro}

The rise of agentic AI systems enables personal AI agents to manage users’ everyday digital tasks in an increasingly autonomous manner. AI products are often able to accomplish a wide range of workflows on behalf of users through self-directed behavior. 

The high degree of delegation and growing reliance on AI agents in human activities also introduce substantial privacy, security, and safety concerns.
For example, there have been reports of AI agents deleting important user emails or making unauthorized purchases without appropriately informing the user \citep{martindaleMetaSecurityResearchers2026,mchughWhosReallyShopping2026}. In another example, AI coding agents with terminal or repository access have executed unintended commands \citep{GooglesAIpoweredAntigravity2025,liu2026youraishelldemystifying,AnthropicSaysIts2026}, modified critical files \citep{liuYourAICoding2026}, or exposed sensitive credentials due to overly broad permissions or prompt injection attacks \citep{mehtaEvenBestAI2026}. These incidents highlight the risks associated with granting AI agents extensive autonomy.
On the other hand, overly restrictive permissions can require users to approve most AI actions manually, which could increase user burden and reduce operational efficiency. Therefore, it is crucial to strike a balance where AI agents are granted the appropriate level of authorization to accomplish desired tasks, while minimizing the potential risks for each user.

\begin{figure}[htbp]
    \centering
    \includegraphics[width=0.8\linewidth]{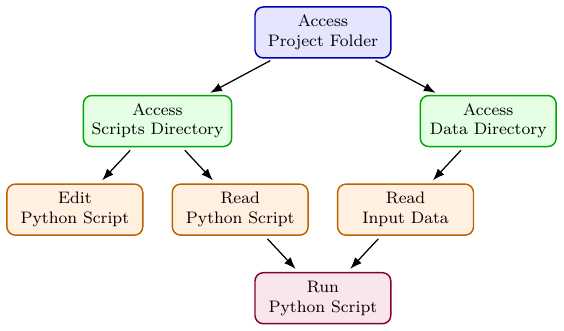}

    \caption{An illustrative hierarchical authorization DAG
    for a coding agent. Nodes represent actions, and directed
    edges indicate authorization dependencies. Authorizing a
    child action requires authorization for all its parent
    actions. In this example, running the Python script requires
    permission to read both the script and its input data,
    while editing permissions are not required.}
    \label{fig::authorization_challenge}
\end{figure}

Meanwhile,
possible AI actions are often structurally dependent, making it hard for statistical formulation and inefficient to grant authorizations independently. Here the dependencies among AI actions may constitute a hierarchy, or a directed acyclic graph (DAG) in general, where authorizing an action may require parent authorizations. Figure \ref{fig::authorization_challenge} provides an illustrative example, where a node denotes a possible AI action, and an edge denotes a structural dependency. For instance, authorizing an AI agent to ``Place Order'' may require authorizations for both ``Add to Cart'' and ``Access Credit Card Info.''
In other words, granting permission for a child action infers permission for their ``parent'' actions. 

Several existing works have attempted to facilitate user authorization. For example, Progent \citep{shi2026progentsecuringaiagents} provides a tool to restrict the behavior of agents when they want to perform some potentially risky operations. Similarly, AgentSpec \citep{Wang_2026} enables users to specify their safety requirements and constrains agent behavior within these user-defined boundaries. 
However, most of these solutions are rule-based, where the same authorization criteria are applied to all users. 
In fact, users may differ substantially in the actions they are willing to delegate to AI agents. For example, one user may allow an agent to complete purchases within a predefined budget, while another may require explicit approval for every transaction. Such differences make personalization essential for predicting authorization decisions and determining appropriate levels of agent autonomy.

In this paper, instead of setting user-agnostic rules, \emph{we propose to learn user authorization preferences through data and achieve personalized user authorization prediction for agentic AI systems.} Specifically, we intend to develop a novel structured prediction model that uses extracted user features to predict authorization preferences. We envision that, based on such a model, each user can be presented with a personalized batch of actions predicted to align with their authorization preferences, which enables users to approve multiple actions collectively and streamline the authorization process.

Two lines of research are related to our work. One is hierarchical multi-label classification, where labels are organized according to a pre-specified tree or DAG structure \citep{rousu2006kernel,wang2009large,7118216}. For example, \cite{chow1968approximating} proposed a method to approximate the probability of each node in a tree, \cite{lafferty2001conditional} proposed conditional random fields, and \cite{taskar2003max} proposed Markov networks. These methods aim to model the dependencies among labels using a structured probabilistic framework. There are also methods that train independent classifiers for one or several nodes, and then predict the tree structure by combining the classifiers \citep{barbedo2006automatic,fagni2007selection,qiu2009hierarchical,silla2011survey,read2021classifier}. Nevertheless, these methods usually require post-processing steps to enforce hierarchical consistency. In the meantime, many of the traditional hierarchical multi-label classification methods do not jointly model all nodes, which may not be able to capture dependencies and result in suboptimal performance.

The second line of research focuses on the development of deep learning architectures for structured prediction tasks, which has drawn a growing interest in recent years \citep{8933476}. For example, \cite{tai2015improved} proposed tree-structured long-short-term memory networks and \cite{socher2013recursive} proposed recursive neural networks for sentiment treebanks. These methods could capture the structural dependencies among nodes in a tree or graph. Furthermore, \cite{wehrmann2018hierarchical} proposed neural architectures HMCN-F and HMCN-R that combine both local and global information to predict hierarchical labels, and a new penalty function is designed to enforce the hierarchical consistency. \cite{giunchiglia2020coherent} proposed C-HMCNN, which adds a constraint layer on top of the neural network to ensure that the predictions satisfy the hierarchical constraints. 

Despite these progress, none of the existing prediction models is designed for personalized user authorization in agentic AI systems. Therefore, several issues remain unresolved. First, in an agentic AI system, users usually do not have the bandwidth to make decisions for every AI action. For example, it is unrealistic for a user to specify access permissions for every file on a device.
As a result, the authorization decisions collected in practice are usually sparse, indicating only the subset of actions that users have previously encountered and responded to. This poses a challenge for learning and inference, as a prediction model must infer authorization decisions based on sparse observations while still respecting the underlying hierarchical structure \citep{9568738}. To the best of our knowledge, most of the existing methods 
typically assume that the complete hierarchical outcomes are fully observed during training, which is often not the case in agentic AI systems. 


Second, some methods estimate node-level marginal probabilities independently, which can produce structural violations and require post-processing to enforce hierarchical consistency. A structural violation occurs when a child node is predicted to be authorized while at least one of its parents is predicted to be unauthorized. 
Resolving violations after prediction also introduces ambiguity: it may be unclear whether to authorize the child’s unauthorized ancestors or revoke authorization for the child. Third, existing methods may struggle to capture complex dependencies, including correlated nodes within the same hierarchy.

To address these issues, we propose a novel neural network architecture called PATH (Parent-gated Authorization Transformer for Hierarchies). Specifically, instead of estimating marginal probabilities and enforcing consistency afterward, PATH adopts a parent-gated probability formulation and directly models conditional probabilities to take into account the chain-rule nature of the hierarchical structure. To estimate these conditional probabilities, PATH further incorporates the hierarchical structure into a masking mechanism of self-attention. 
As a result, the proposed design ensures hierarchical consistency, and hence avoids the need for post-processing or additional penalty terms. 
We further introduce depth-level embeddings.
By sharing information across these nodes, this design simultaneously addresses the sparsity caused by partially observed authorization decisions and captures shared patterns among nodes at the same hierarchical depth. We also derive the theoretical results and conduct experiments, both of which confirm the effectiveness of our method.

The remainder of this article is organized as follows. Section \ref{sec:methods} presents the problem formulation and our proposed method. Section \ref{sec:theory} establishes its theoretical guarantees. Section \ref{sec:experiments} evaluates its performance against competing methods using simulated data, AI-generated user profiles, and real user authorization data. Finally, Section \ref{sec:discussion} concludes the article.

\section{Proposed Problem and Method}
\label{sec:methods}
\subsection{Problem Setup}
\label{subsec: problem}
\begin{figure}[htbp]
    \centering
\includegraphics[width=\linewidth]{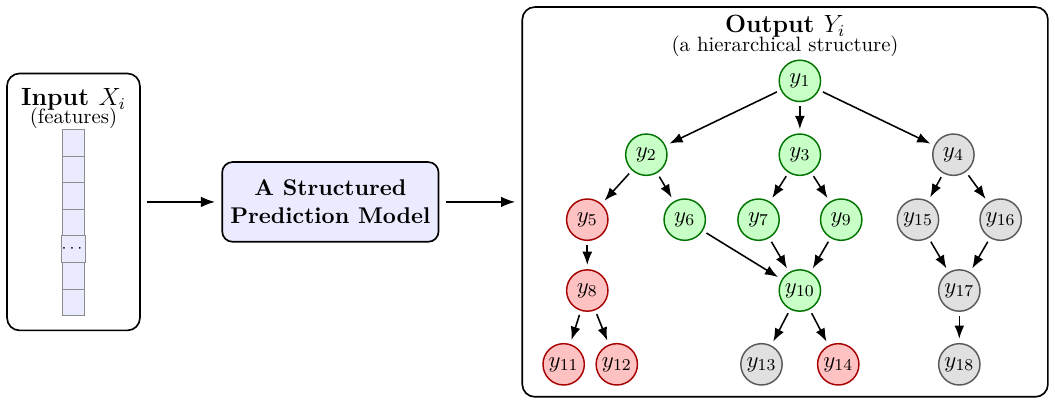}

    \caption{An illustration of a structured prediction problem:
    learning hierarchical authorization output ($Y_i$) from observed
    user features ($X_i$). A green node denotes an authorized AI action
    (i.e., $Y_{ij}=1$), a red node denotes a denied AI action
    (i.e., $Y_{ij}=0$), and a gray node denotes
    an unobserved authorization decision. If a node is denied, all its descendants must also be denied.}
    \label{fig::authorization_tree}
\end{figure}

We consider the problem of user authorization prediction as predicting a hierarchy-structured output given a set of user features. See Figure \ref{fig::authorization_tree} for an illustration. Let $\mathcal{X} \subseteq \mathbb{R}^d$ denote the input space, where each $x \in \mathcal{X}$ represents the $d$-dimensional observed features of a user. Let $\mathcal{V} = \{1, \dots, m\}$ denote a finite set of authorization decisions, where each element corresponds to a permission associated with a potential AI action, such as folder access or system setting modification. 
For each user $i$, our goal is to predict the corresponding output $Y_i=(Y_{i1}, \dots, Y_{im})$, where for any $j \in \mathcal{V}$, $Y_{ij} = 1$ indicates that permission $j$ is granted and $Y_{ij} = 0$ otherwise. Let $\mathcal{Y} = \{0, 1\}^m$ denote the output space.

Following this setup, we assume that the elements in $\mathcal{V}$ are organized according to a hierarchical structure. This structure is shared and fixed across all users, although user authorization outcomes may differ. We represent the hierarchical structure as a DAG 
$$
\mathcal{G} = (\mathcal{V}, \mathcal{E}),
$$
where $\mathcal{E} \subseteq \mathcal{V} \times \mathcal{V}$ denotes the set of directed edges encoding dependency relationships among authorization decisions. For any node $j \in \mathcal{V}$, we denote the set of its parents by $\mathrm{pa}(j)$ and the set of ancestors by $\mathrm{Anc}(j)$.
Then a valid authorization structure for a DAG satisfies the following hierarchical consistency constraint:
$$
Y_{i,j_2} = 1 \implies Y_{i,j_1} = 1, \quad \mbox{for any } j_1 \in \mathrm{Anc}(j_2).
$$
That is, if a permission $j_2$ is granted, then all of its ancestors must also be granted. This captures the hierarchical dependencies among permissions, where granting a child permission implies granting all of its parent permissions.

Based on the above setup, the problem of predicting authorization can be formulated as learning a function $g: \mathcal{X} \to \mathcal{Y}$ that maps observed features to a valid DAG-structured output while satisfying the hierarchical consistency constraint. Under this setting, we also seek to optimize function $g$ that achieves high prediction performance while being able to handle incomplete observations effectively during training.
In this paper, we only focus on the setting where the DAG structure $\mathcal{G}$ is known and fixed, as it represents the authorization rules of the underlying agentic AI system. However, we allow each $Y_i$ to be partially observed.

\subsection{The Proposed Method}
\label{subsec: proposed}
In this subsection, we present the proposed PATH, which directly incorporates the hierarchical dependencies. 
Recall that the core component of a Transformer \citep{vaswani2017attention} is the self-attention mechanism, which allows the model to capture dependencies among input tokens. In the proposed method, we re-arrange the whole hierarchical structure as a vector with a given ordering, such that the self-attention mechanism can be used to capture the dependencies among nodes. 

Specifically, we first encode the user features through a multi-layer perceptron (MLP) to obtain a user embedding $u_i = f_\theta(X_i) \in \mathbb{R}^l$ for each user $i$, where $\theta$ denotes the learnable parameters associated with the MLP and $l$ is the dimension of the user embedding. Meanwhile, we assign each node $j$ a learnable embedding $e_j \in \mathbb{R}^l$, along with a depth-level embedding $d_j \in \mathbb{R}^h$ corresponding to the depth of node $j$ in the hierarchy (i.e., the shortest path distance from $j$ to the root node). For each user-node pair $(i,j)$, we construct an initial token representation as
\[
t_{ij}^{(0)} = u_i + e_j + d_j.
\]
Stacking all node tokens yields a sequence $T_i^{(0)} \in \mathbb{R}^{m \times h}$ for each user $i$.

\begin{figure}[htbp]
    \centering
    \includegraphics[width=\linewidth]{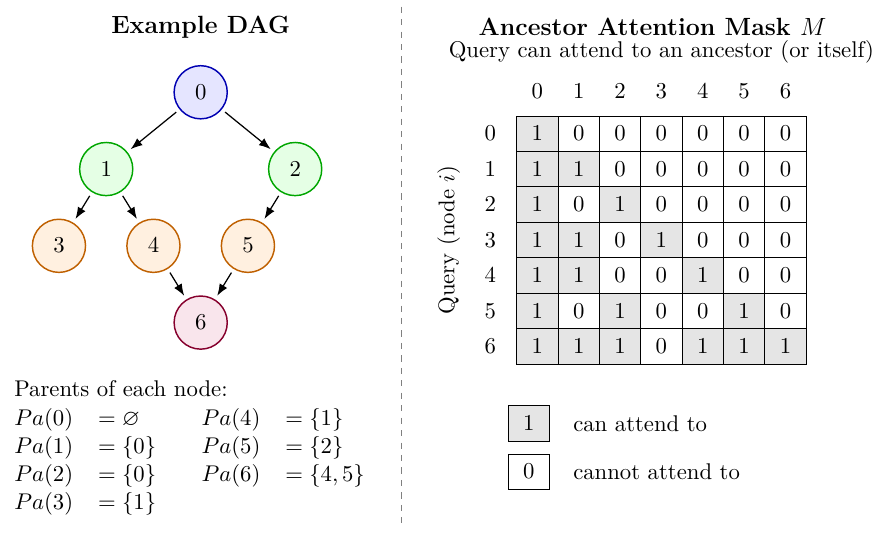}
    \caption{An illustration of the proposed attention mask $M$ corresponding to
    Figure~\ref{fig::authorization_challenge}. The proposed mask is designed such that each node can only attend to itself and its ancestors, ensuring that the model captures the hierarchical dependencies in the authorization structure. In this example, node 5 can attend to nodes 2 and 0 (its ancestors) and itself, but not to nodes 1, 3, or 4.
}
    \label{fig::matrix_structure}
\end{figure}

To capture structural dependencies among nodes, we apply a Transformer encoder with a structure-based attention mask. Specifically, we restrict the self-attention mechanism such that node $j$ can only attend to its ancestors and itself. Formally, the attention mask $M \in \{0,1\}^{m \times m}$ is defined as
\[
M_{jk} =
\begin{cases}
1, & k \in \mathrm{Anc}(j) \cup \{j\}, \\
0, & \text{otherwise}.
\end{cases}
\]
Figure~\ref{fig::matrix_structure} illustrates the proposed attention mask matrix.

Then the Transformer encoder computes structure-aware node representations
\[
H_i = \mathrm{Transformer}(T_i^{(0)}, M),
\]
where $H_i = [h_{i1}, \dots, h_{im}]$ denotes the learned hidden representations for all authorization decisions, and $\mathrm{Transformer}()$ denotes the original Transformer encoder that uses $T_i^{(0)}$ as its input, and the proposed $M$ as its self-attention matrix.
For each node $j$, we then compute a base activation probability
\[
q_{ij} = \sigma(w^\top h_{ij}),
\]
where $\sigma(\cdot)$ denotes the sigmoid function and $w \in \mathbb{R}^h$ is a learnable weight vector. The quantity $q_{ij}$ represents the conditional probability that node $j$ is authorized by user $i$ (i.e., takes value 1), given that all of its ancestor nodes have been authorized by user $i$.

To enforce structural consistency, we introduce a parent-gated probability, where each marginal probability is formulated based on the probability chain rule. The probability for root node (denoted as node 0) to be authorized is $\hat{p}_{i0} = q_{i0}$. For any non-root node $j>0$, we recursively define the final probability as
\[
\hat{p}_{ij}
=
q_{ij}
\cdot
\prod_{p \in \mathrm{Pa}(j)} \hat{p}_{ip}.
\]
This gating mechanism ensures that the predicted marginal probability of each node is modulated by its parents. This guarantees zero structural violation. In particular, if any parent node has zero probability, the predicted probability of the child node is also forced to be zero.

\begin{figure}[htbp]
\centering
\includegraphics[width=\linewidth]{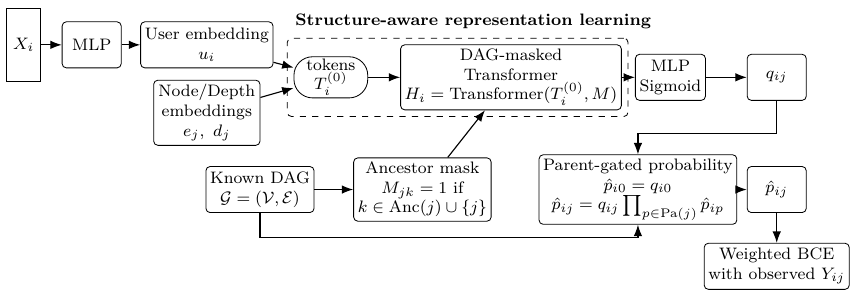}
\caption{Workflow of the proposed PATH architecture. User features are encoded into a user embedding and combined with node and depth embeddings to form node tokens. Here depth embeddings are used to accommodate data sparsity and incorporate correlations among nodes of the same hierarchical depth. The known authorization DAG provides both the ancestor attention mask for the Transformer encoder and the parent sets used by the parent-gated probability formulation. The final probabilities are trained end-to-end with weighted BCE.}
\label{fig:simple_workflow}
\end{figure}

This formulation integrates the DAG structure at two levels: the masked attention mechanism restricts information flow along valid dependency paths, while the parent-gated probability ensures that predictions satisfy hierarchical constraints by following the probability chain rule. As a result, the model produces structurally consistent outputs without requiring any post-processing or additional constraint penalties. Figure \ref{fig:simple_workflow} illustrates the flow of the proposed method.

For the loss function, we follow \cite{wehrmann2018hierarchical} and adopt a unweighted binary cross-entropy loss (BCE) \citep{gneiting2007strictly,reid2010composite,mohan2021graphical} for end-to-end model training
\[
\mathcal{L}
=
-\frac{1}{n}\sum_{i=1}^n \sum_{j=1}^m
\left[
Y_{ij}\log(\hat{p}_{ij})
+
(1-Y_{ij})\log(1-\hat{p}_{ij})
\right].
\]
In practice, we can also replace the term $Y_{ij}\log(\hat{p}_{ij})$ by $\alpha Y_{ij}\log(\hat{p}_{ij})$. Here $\alpha$ is a positive class weight used to address class imbalance in hierarchical multi-label classification tasks, especially when positive labels are sparse. Recall that the child nodes could be 1 only when their parents are 1, which means the marginal probability of child nodes being 1 decreases along the depth of the tree, possibly causing a lot more 0's than 1's in $Y$. The adoption of $\alpha$ gives more weight to the positive class, which encourages the model to pay more attention to correctly predicting active permissions while still maintaining overall performance \citep{5128907}.

\section{Theoretical Analysis}
\label{sec:theory}
In this section, we develop a theoretical analysis of the proposed PATH for learning hierarchical authorization probabilities from partially observed labels. We first establish finite-sample bounds for BCE risk and probability estimation error. The bounds separate model approximation, statistical estimation, optimization error, and selection bias, which allows us to construct explicit conditions for probability and risk consistency. The proof is provided in Appendix.

\subsection{Setup and Assumptions} \label{sec:assumption}
Let $G=(V,E)$ be the fixed authorization structure, with root set $V_0$ and
nonroot set $J=V\setminus V_0$, and $m=|V|$.
For each user, $X\in\mathcal X\subseteq\mathbb R^d$ denotes the features,
$Y\in\{0,1\}^{m}$ the complete authorization vector, and
$O\in\{0,1\}^{m}$ the observation indicators. Only the entries with
$O_j=1$ are observed. Then the sample is denoted by
$\mathcal D_n=(Z_1,\ldots,Z_n)$, where $Z_i=(X_i,Y_i,O_i)$ specifies
the complete data for user $i$, and $n$ is the number of users. For each nonroot node $j\in J$, we define the following
conditional probabilities:
\begin{align*}
\eta_j(x)&=\mathbb P(Y_j=1\mid X=x), &
\nu_j(x)&=\mathbb P(Y_j=1\mid X=x,O_j=1),\\
\pi_j(x)&=\mathbb P(O_j=1\mid X=x), &&j\in J.
\end{align*}
Here, $\eta_j(x)$ denotes the true authorization probability,
$\nu_j(x)$ denotes the authorization probability conditional on
the decision being observed, and $\pi_j(x)$ denotes the
observation probability.
We set $\eta_r=\nu_r=1$ for $r\in V_0$. 

Recall that $f_{\theta}$ is the user embedding function, where $\theta$ is the model parameter. Since the output is a function of $\theta$, and we only focus on the theoretical result of each individual $i$, we omit the subscript $i$ in our notation and use $(\theta,j)$ instead throughout this section for simplicity. Under our model setting, final predictions can be defined in topological order by
\begin{align} 
 q_{\theta,j}(x)&=\sig(f_{\theta,j}(x)),\quad j\in J,\qquad \sig(t)=(1+e^{-t})^{-1},\\
 p_{\theta,r}(x)&=1,\qquad r\in V_0,\\
 p_{\theta,j}(x)&=q_{\theta,j}(x)\prod_{k\in\Pa(j)}p_{\theta,k}(x),\qquad j\in J.\label{eq:pq}
\end{align}
Then we introduce some regularity conditions.
\begin{assumption}\label{ms:ass1}
The following statements hold:
\begin{enumerate}
    \item $Z_1,\ldots,Z_n$ are i.i.d.
    \item For $r\in V_0$, $Y_r=1$, and $Y_j\le Y_k$
for every $j\in J$ and $k\in \Pa(j)$, a.s.
    \item There exists a constant
$\pi_{\min}>0$ such that $\pi_j(X)\ge\pi_{\min}$ a.s. for
every $j\in J$.
\end{enumerate} 
\end{assumption}
Assumption \ref{ms:ass1}(ii) states that root nodes are
always authorized, and a node can be authorized only if all
its parents are authorized. In practice, an authorization structure usually has a finite number of nodes.
Assumption \ref{ms:ass1}(iii) ensures that every nonroot
node has a strictly positive probability to be observed. We further assume that the authorization structure and model architecture are fixed in the training procedure.

\begin{definition}\label{ms:ass2}
Let the parameter space be denoted as $\Theta\subseteq\mathbb R^D$
, and let $\theta$ be the model parameter. We assume that
$f_{\theta,j}:\mathcal X\to\mathbb R$ is the finite-valued measurable function in $(\theta,x)$ for each $\theta\in\Theta$ and $j\in J$.
Then the raw prediction class is
\begin{equation*}
 \mathcal F
 =\bigl\{p_\theta=(p_{\theta,j})_{j\in V}:
               \mathcal X\to[0,1]^m\;\big|\;\theta\in\Theta\bigr\},
 \label{ms:class-definition}
\end{equation*}
where $p_\theta$ is defined through the structure in \eqref{eq:pq}.
Thus $p_{\theta,j}=1$ at roots and $0<p_{\theta,j}<1$ at nonroots.

\end{definition}

Definition \ref{ms:ass2} gives the definition of $\mathcal F$, the actual training class, i.e. all the possible $p_{\theta}$ that can be fitted via our architecture.

\begin{assumption}\label{ms:ass3}
For every $j\in J$, $\nu_j(X)=\eta_j(X)$ a.s.
\end{assumption}
Assumption~\ref{ms:ass3} allows observation probabilities to depend on
user features \citep{rubin1976inference}. For binary labels with positive observation probabilities,
this assumption is equivalent to $O_j\perp Y_j\mid X$ for each node
separately \citep{dawid1979conditional}. This conditional independence assumption may not
hold in practice. Therefore, our main theoretical result does not
rely on it. Nevertheless, we will show that given Assumption \ref{ms:ass3}, we can further eliminate the upper bound for the error term, which gives us a consistent estimator.

Under the binary cross-entropy (BCE) loss, we define
\begin{align*}
 \ell(p,y)&=-y\log p-(1-y)\log(1-p),\notag\\
 \widehat R_O(p)
   &=\frac1{nm}\sum_{i=1}^n\sum_{j\in J}
          O_{ij}\ell(p_j(X_i),Y_{ij}),\notag\\
 R_O(p)
   &=\mathbb E\!\left[\frac1m\sum_{j\in J}
          O_j\ell(p_j(X),Y_j)\right],
 \qquad
 R(p)=\mathbb E\!\left[\frac1m\sum_{j\in J}
          \ell(p_j(X),Y_j)\right].
 \label{ms:risks}
\end{align*}
Here, $\widehat R_O(p)$ denotes the empirical risk computed
from observed authorization decisions, $R_O(p)$ denotes
its population counterpart, and $R(p)$ denotes the
population risk based on complete authorization outcomes.
We adopt the usual endpoint conventions, including
$0\log 0=0$, so that $R(\eta)$ is well defined.

Let $\widehat p\in\mathcal F$ be a measurable estimator satisfying
\begin{equation}
 \widehat R_O(\widehat p)
 \le\inf_{p\in\mathcal F}\widehat R_O(p)+\xi,
 \qquad \xi\ge0,
 \label{ms:erm}
\end{equation}
almost surely, where $\xi$ is a deterministic optimization
tolerance. This condition requires the optimization algorithm
to return a solution whose empirical risk is within $\xi$
of the minimum achievable over $\mathcal F$.
Thus, $\xi$ quantifies the optimization error and can be set
to zero when the empirical risk is minimized exactly.

Define the observed approximation error and empirical
deviation by
\begin{align*}
 A_{O,\mathcal F}
   &=\inf_{p\in\mathcal F}
        \{R_O(p)-R_O(\nu)\},\notag\\
 \Delta_n
   &=\sup_{p\in\mathcal F}
        |\widehat R_O(p)-R_O(p)|.
\end{align*}
By definition, $A_{O,\mathcal F}\ge0$ measures the
approximation error arising from the discrepancy between
the model class $\mathcal F$ and the underlying
data-generating mechanism. In particular,
$A_{O,\mathcal F}=0$ when $\nu\in\mathcal F$,
and it can still be small when $\mathcal F$ is sufficiently
large.
On the other hand, $\Delta_n$ measures the uniform
deviation of the empirical risk from its population
counterpart. We assume that, for any $0<\delta<1$,
there exists a deterministic bound $T_{n,\delta}$
such that
\begin{equation}
 \mathbb P_{\mathcal D_n}
 \{2\Delta_n\le T_{n,\delta}\}\ge1-\delta.
 \label{ms:deviation}
\end{equation}
Here, $\delta$ represents the probability of exceeding
the specified deviation bound. Under suitable
uniform convergence conditions, $T_{n,\delta}$
can be chosen to converge to zero as $n\to\infty$
for any fixed $\delta$.

\subsection{Theoretical Results}
Let
\begin{equation*}
 d_2(a,b)^2=\mathbb E_X\!\left[\frac1m\sum_{j\in J}
                       (a_j(X)-b_j(X))^2\right].
 \label{ms:distances}
\end{equation*}
The distance $d_2(a,b)$ is a normalized $L^2$ distance
that measures the discrepancy between two vector-valued functions, averaged over the nonroot nodes
and the feature distribution.
Now we can demonstrate our main theorem, which provides an upper bound for $R_O$ and $d_2$ of our method.

\begin{theorem}\label{ms:main-theorem}
Suppose Assumptions~\ref{ms:ass1} holds, and that
\eqref{ms:erm} and \eqref{ms:deviation} are satisfied. Then, with probability at
least $1-\delta$ over $\mathcal D_n$, the following inequalities hold:
\begin{align}
 &R_O(\widehat p)-R_O(\nu)
       \le C_{n,\delta},\label{ms:observed-bound}\\
&d_2(\widehat p,\eta)
       \le\sqrt{\frac{C_{n,\delta}}
                             {2\pi_{\min}}}+d_2(\nu,\eta),
       \label{ms:probability-bound}
\end{align}
where $C_{n,\delta}=A_{O,\mathcal F}+T_{n,\delta}+\xi$.
\end{theorem}
Theorem~\ref{ms:main-theorem} establishes upper bounds on
the observed risk and the $L^2$ estimation error,
which are governed by $C_{n,\delta}$ and the discrepancy
$d_2(\nu,\eta)$. The latter captures the bias induced by
the dependence between the observation mechanism and
authorization outcomes, conditional on user features.
In particular, if Assumption~\ref{ms:ass3} holds,
then $d_2(\nu,\eta)=0$.

On the other hand, $C_{n,\delta}$ consists of three
components: the approximation error $A_{O,\mathcal F}$,
the optimization error $\xi$, and the statistical
estimation error $T_{n,\delta}$. 
Under some conditions, these terms could be exactly 0 or asymptotically converge to 0, as we discuss
in the remainder of this section. We begin with the following
corollary, which shows the convergence of $T_{n,\delta}$.

\begin{corollary}
\label{ms:estimation-special}
Suppose Assumptions~\ref{ms:ass1} hold.
Assume that the authorization structure $\mathcal{G}$ and prediction class $\mathcal F$
are fixed independently of the training sample and that,
for some fixed $0<\varepsilon<1/2$,
\begin{equation*}
\varepsilon\le p_{\theta,j}(x)\le1-\varepsilon
\qquad
\text{for all }\theta\in\Theta,\ x\in\mathcal X,\ j\in J.
\label{ms:special-output-bound}
\end{equation*}
Suppose the corresponding gate class satisfies
\begin{equation*}
\log\mathcal N\left(
u,\,
\{(q_{\theta,j})_{j\in J}:\theta\in\Theta\},
\,\|\cdot\|_\infty
\right)
\le c\log(M/u),
\qquad 0<u\le1,
\label{ms:special-covering}
\end{equation*}
for fixed constants $c\ge1$ and $M\ge1$.
Here $\mathcal N$ denotes the covering number with centers
in the class, and
\begin{equation*}
\|q-q'\|_\infty
=
\sup_{x\in\mathcal X}\max_{j\in J}
|q_j(x)-q'_j(x)|.
\label{ms:gate-uniform-norm}
\end{equation*}
Let $K_G\ge1$ be the fixed product-gate sensitivity constant
satisfying
\begin{equation*}
\begin{aligned}
&\sup_{x\in\mathcal X}\max_{j\in J}
|p_{\theta,j}(x)-p_{\theta',j}(x)|\\
&\qquad\le
K_G
\sup_{x\in\mathcal X}\max_{j\in J}
|q_{\theta,j}(x)-q_{\theta',j}(x)|
\end{aligned}
\label{ms:special-gate-sensitivity}
\end{equation*}
for all $\theta,\theta'\in\Theta$.

Then the empirical deviation condition
\eqref{ms:deviation} holds with
\begin{equation*}
T_{n,\delta}
=
\frac4n+
2\log(1/\varepsilon)
\sqrt{
\frac{
c\log(MK_Gn/\varepsilon)+\log(2/\delta)
}{
2n
}
}.
\label{ms:explicit-estimation}
\end{equation*}
In particular, for $\delta_n=n^{-2}$,
\begin{equation*}
T_{n,n^{-2}}
=
O\left(\sqrt{\frac{\log n}{n}}\right)
\longrightarrow0.
\label{ms:estimation-rate}
\end{equation*}
\end{corollary}

Corollary~\ref{ms:estimation-special} provides an
upper bound on the statistical estimation error
$T_{n,\delta}$. Specifically, the boundedness condition
on the predicted probabilities ensures that the BCE loss
remains bounded, while the covering number condition
controls the complexity of the prediction class.
The product-gate sensitivity constant $K_G$ characterizes how perturbations in the gate functions propagate through the DAG structure. 
As a result, these conditions yield a convergence rate of
$O(\sqrt{\log n/n})$. In other words, the statistical
estimation error goes to 0 as the sample size increases.

The following corollary further shows the bound on the prediction error.
\begin{corollary}
\label{ms:fidelity}
Suppose the conditions of Theorem~\ref{ms:main-theorem} hold.
For a new independent feature vector $X$
and any $t>0$,
\begin{equation*}
\begin{aligned}
&\mathbb P_{\mathcal D_n,X}
\left\{
\max_{j\in V}
|\widehat p_j(X)-\eta_j(X)|\ge t
\right\}\\
&\qquad\le
\min\left\{
1,\,
\delta+\frac{m}{t^2}
\left(
\sqrt{\frac{C_{n,\delta}}{2\pi_{\min}}}
+d_2(\nu,\eta)
\right)^2
\right\}.
\end{aligned}
\label{ms:fresh-tail}
\end{equation*}
For $n\ge2$, applying the theorem with $\delta=n^{-2}$ also gives
\begin{equation*}
\mathbb P_{\mathcal D_n}
\left\{
d_2(\widehat p,\eta)
\ge
\sqrt{\frac{C_{n,n^{-2}}}{2\pi_{\min}}}
+d_2(\nu,\eta)+\frac1n
\right\}
\le n^{-2}.
\label{ms:training-tail}
\end{equation*}
\end{corollary}

Corollary~\ref{ms:fidelity} extends the $L^2$
estimation bound in \eqref{ms:probability-bound} to probabilistic guarantees on the prediction error for new users.
The first inequality bounds the probability that
the maximum prediction error across authorization
nodes exceeds a given threshold.
The second inequality provides a high-probability
bound on the overall $L^2$ estimation error.

According to Theorem \ref{ms:main-theorem}, we know that the $L^2$ distance between $\hat{p}$ and $\eta$ is controlled by $C_{n,\delta}$ and $d_2(\nu,\eta)$. Under some conditions, we can show that $d_2(\hat{p},\eta)$ can converge to 0 as $n\to\infty$, proving the consistency of our estimator. This is formalized in Corollary \ref{ms:consistency}.

\begin{corollary}
\label{ms:consistency}
Assume that estimators $\widehat p_n\in\mathcal F_n$ satisfies the conditions of 
Theorem~\ref{ms:main-theorem} for each sample size $n$.
Let $\nu_n$ and $\pi_{\min,n}$ denote the
corresponding observed-label target and positivity bound, respectively, that varies as $n$ increases. And let $C_{n,\delta_n}$ be defined as in Theorem ~\ref{ms:main-theorem}, with $\delta_n$ varying with $n$. If 
$
\delta_n\to0,
$
$
\frac{C_{n,\delta_n}}{\pi_{\min,n}}\to0,
$
then $$d_2(\widehat p_n,\nu_n)\xrightarrow{\mathbb P}0.$$ Additionally, if $d_2(\nu_n,\eta)\to0$,
then
\begin{align*}
d_2(\widehat p_n,\eta)
&\xrightarrow{\mathbb P}0,\\
\max_{j\in V}
|\widehat p_{n,j}(X)-\eta_j(X)|
&\xrightarrow{\mathbb P}0,
\end{align*}
where $X$ denotes an independent feature
vector from a new user.
\end{corollary}

As mentioned in Corollary \ref{ms:consistency}, a sufficient condition for estimator consistency is that $C_{n,\delta}\to 0$. Since
$C_{n,\delta_n}=A_{O,\mathcal F_n}
+T_{n,\delta_n}+\xi_n$, its convergence
to zero is equivalent to the convergence of each
component to zero. 
As discussed in Section \ref{sec:assumption}, $A_{O,\mathcal F_n} \ge 0$ measures the
approximation error, which satisfies
$A_{O,\mathcal F_n}=0$ when $\nu_n\in\mathcal F_n$, or can be small when $\mathcal F_n$ is sufficiently
large. Meanwhile, the convergence of $T_{n,\delta}$ has been illustrated in Corollary \ref{ms:estimation-special}.

We next discuss the convergence of the optimization
error $\xi$, which depends on the convergence properties
of the optimization algorithm. If the algorithm can achieve the optimized $\theta$ with the increase of number of iterations, then $\xi$ can converge to 0.
Specifically,
Theorem~\ref{thm:main} in the Supplementary Material establishes the convergence
rate of the optimization
algorithm used in our method (i.e., Adam \citep{kingma2014adam}) under certain conditions.

\section{Simulation Studies}
\label{sec:experiments}
We conduct three simulation studies to evaluate the performance of our method and compare it with other competing methods in different scenario. The first study considers conditionally independent node-level mechanisms, providing a baseline setting for evaluation. The second study introduces additional correlations among authorization decisions. The third study uses AI-generated synthetic user profiles and their responses, providing a more realistic setting.

Each dataset consists of user features and corresponding hierarchical authorization labels following a fixed authorization structure, with incomplete observations of structured outputs. Across all three studies, we evaluate both predictive accuracy and adherence to hierarchical constraints.
We compare PATH with four baselines. Notice that, all baselines were designed only for complete structured outputs, while details of model parameters and data structure are provided in Section \ref{app:experimental-details} of Supplementary Material: 
\begin{itemize}
  \item Independent Classifier \citep[IC;][]{cox1958regression, BOUTELL20041757}: An independent classifier is conducted at each node that ignores the hierarchical structure, treating each node as an independent binary classification problem. We use logistic regression as the independent classifier.
  \item Chain Classifier \citep[CC;][]{zhang2010multi, sucar2014multi}: A classifier for each node using only the samples for which all parent nodes are authorized. 
That is, the model estimates the conditional authorization probability of node $j$ given that all of its parent permissions are authorized. For prediction, it follows a top-down strategy, where the root node is predicted first, followed by its child nodes using the conditional probability recursively. We use logistic regression as the classifier.
\item HMCN-F \citep{wehrmann2018hierarchical}: A neural network architecture designed for hierarchical multi-label classification. It jointly learns global representations for the entire label hierarchy and local representations for individual hierarchy levels, and combines their predictions to improve classification performance while encouraging hierarchical dependency by applying a penalty term.
  \item C-HMCNN \citep{giunchiglia2020coherent}: A state-of-the-art hierarchical multi-label classification neural network that incorporates a constraint layer to ensure predictions satisfy the hierarchical constraints without post-processing.
\end{itemize}

\subsection{Simulation Study 1: Conditionally Independent Authorization Decisions.}
In this study, the authorization structure is represented by a directed hierarchical graph containing 171 permission nodes with maximum depth 5. The root node corresponds to global system access. Below the root, we assume that the structure consists of eight primary business domains, including Finance, HR, Research, IT, Teaching, Operations, Procurement, and StudentServices. Each domain further expands into operation-level permissions. There are also cross-domain permissions and governance/audit permissions, resulting in some nodes with multiple parent nodes. In total, the graph contains 52 multi-parent nodes, and the maximum number of parent nodes per node is 5. 

The user features are i.i.d generated from $N(0,I_d)$ with a fixed $d=10$ in this experiment, and the response, i.e. the hierarchical authorization labels, are generated as follows. The root node is always set to be active, i.e. $Y_{i0}=1$. For any non-root node $j$, we randomly sample a coefficient $\beta_j\in \mathbb{R}^d$ from $N(0,0.55^2I_d)$. In addition, an intercept term $b_j$ is generated from $N(\mu_j,0.35^2)$, where $\mu_j=-0.45 - 0.45 \cdot \mathrm{depth}(j) - 0.28 \bigl(n_{pa}(j)-1\bigr)$. Here $\mathrm{depth}(j)$ means the shortest path distance from node $j$ to root node, and $n_{pa}(j)$ is the number of parent nodes associated with node $j$. This setting ensures that those deep nodes and with more parent nodes are unlikely to be authorized. Such a design is also consistent with real-world authorization systems, where permissions with greater structural complexity are typically more sensitive and therefore less likely to be granted. 



The conditional probabilities are then computed as
$$
q_{ij} = \sigma\left(b_j + x_i^\top \beta_j\right),
$$
where $\sigma(\cdot)$ denotes the sigmoid function. After obtaining all probabilities $q_{ij}$, each non-root node $j$ is generated according to
$$
Y_{ij} \sim
\begin{cases}
\mathrm{Bernoulli}(q_{ij}),
& \text{if } Y_{ip}=1,\ \forall p \in \mathrm{Pa}(j), \\
0,
& \text{otherwise}.
\end{cases}
$$

Consequently, all generated labels satisfy the hierarchical consistency constraint
$$
Y_{ij} \leq Y_{ip},
\qquad
\forall p \in \mathrm{Pa}(j),
$$
which ensures that a child permission cannot be authorized unless all of its parent permissions are authorized.

Under this data generation mechanism, each user authorizes approximately $15.21$ permission nodes on average, while the median node-wise authorization rate is only $0.0375$. This indicates an extremely sparse multi-label prediction setting, and closely resembles real-world authorization systems in which only a small subset of permissions are typically granted by each individual user.
To simulate incomplete observations, we randomly mask the authorization outcomes of 30\% of nodes (i.e., set them to NA). Then all their child nodes of the masked nodes are also labeled as NA, leading to a total unobservation rate of 65\%.

Each simulated dataset contains $N = 2000$ user samples, which are randomly partitioned into training, validation, and test sets of sizes $1200$, $400$, and $400$, respectively. Models are trained on the training set, hyperparameters are selected based on validation performance, and the final evaluation results are reported on the test set.
The experiment is replicated $100$ times. 

We evaluate model performance at the node level by comparing each predicted label $\hat{Y}_{ij}$ with its corresponding ground-truth label $Y_{ij}$. The classification performance is assessed using standard metrics including AUC, the F1 scores, and their corresponding 95\% confidence intervals (CIs) over 100 replicates. We also conducted two sample $t$-test for each baseline with our method using AUC values, and report the $p$-values to see whether there is significant difference. In addition, we evaluate the structural validity of model predictions by computing the binary violation rate prior to post-processing procedure. This quantity is defined as the proportion of predicted node labels that violate the hierarchical consistency constraint, and a violation occurs when a child node is predicted to be active while at least one of its parent nodes is predicted to be inactive. Therefore, a lower violation rate indicates better adherence to the underlying hierarchical structure. For those methods with non-negative violation rates, we also evaluate their AUC and F1 score (using a classification threshold of 0.5) after post-processing, based on the rule that modifies all violated child nodes to be 0. This imposes minimal modifications to the original predicted outcomes and could enforce the sanctification of hierarchical structure. 


The experimental results are summarized in Table~\ref{tab:complex_authorization_dag_results}. PATH achieves the highest AUC value of $0.9013$ and a reasonably high F1 score, demonstrating superior predictive performance compared with all baseline approaches. Moreover, CC, C-HMCNN, and the proposed method attains a binary violation rate of $0$ without requiring any post-processing procedure or explicit structural penalty, indicating that all predicted labels satisfy the hierarchical constraints by construction. In contrast, IC produces a violation rate of $0.0570$, suggesting that independently modeling each node fails to adequately preserve the underlying hierarchical dependencies. HMCN-F also produces a non-zero violation rate, indicating that a penalty term could help preserve the hierarchical structure, but cannot ensure that is always followed. Notice that, although CC and C-HMCNN are able to eliminate structural violations, they primarily achieve this through a conditional generation procedure or additional constraint-enforcing layers, respectively. The former does not jointly learn from all nodes simultaneously, while the latter does not fully catch the chain-rule nature of the hierarchical structure. As a result, their predictive performance is generally no better than that of the proposed method. In addition, the 95\% CI and $p$-values show that our method is significantly better than others in AUC.

The results also highlight the limitations of enforcing hierarchical consistency through post-processing. For IC, post-processing reduces the AUC from $0.8801$ to $0.8427$, although its F1 score increases from $0.4282$ to $0.4956$. In contrast, the changes in AUC and F1 for HMCN-F are negligible, consistent with its already low violation rate. In other words, enforcing structural constraints through post-processing may come at the cost of predictive performance. Actually, in practice, the poor prediction performance also lead to more actions that must be taken by users.



\begin{table}[htbp]
\centering
\caption{Simulation study comparing the performance of different methods for predicting incomplete hierarchical authorization outcomes using AUC, F1 score, and the violation rate. Best results are highlighted in bold. The $p$-value is computed from a two-sample $t$-test comparing the AUC of each baseline against the proposed method over 100 replicates. The suffix ``(post)'' denotes results after post-processing to enforce hierarchical consistency.}
\label{tab:complex_authorization_dag_results}
\vspace{6pt}
\scalebox{0.9}{
\begin{tabular}{lcccccc}
\hline
Method  & AUC(95\% CI) & F1(95\% CI) & Violation Rate & $p$-value\\
\hline
IC & 0.8801 [0.8790, 0.8812]& 0.4282 [0.4259, 0.4304]& 0.0570 & $<10^{-4}$\\
HMCN-F & 0.8850 [0.8840, 0.8859] & 0.4438 [0.4409, 0.4466] & $<10^{-4}$ & $<10^{-4}$\\
IC(post) & 0.8427 [0.8409, 0.8444]& 0.4956 [0.4933, 0.4978]& 0.0000 & $<10^{-4}$\\
HMCN-F(post) & 0.8850 [0.8840, 0.8859] & 0.4438 [0.4409, 0.4466] & 0.0000 & $<10^{-4}$\\
CC &  0.8420 [0.8405, 0.8435]& \textbf{0.5334} [0.5312, 0.5357] & 0.0000 & $<10^{-4}$\\
C-HMCNN &  0.8657 [0.8647, 0.8666]& 0.4310 [0.4272, 0.4349] & 0.0000 & $<10^{-4}$\\
\textbf{PATH} & \textbf{0.9013} [0.9003, 0.9022] & 0.5276 [0.5252, 0.5300] & 0.0000 & --\\
\hline
\end{tabular}
}
\end{table}

\subsection{Simulation Study 2: Correlated Authorization Decisions.}
To further evaluate the performance of the proposed method in the presence of dependencies beyond the hierarchical structure, we consider a simulation setting where authorization decisions are correlated across nodes for each user. This is achieved through latent user-specific factors. The setting is identical to the first simulation scenario except that conditional probabilities $q_{ij}$ are computed through
$$
q_{ij}=\sigma\left(b_j+x_i^\top \beta_j+cs_i\gamma_j\right),
$$
where $s_i=U_i^\top v$. Here $U_i, v\sim N(0,I_2)$ and $\gamma_i\sim \mathrm{LogNormal} (0, 0.2^2)$. Under this setting, since all probabilities share the same $s_i$, there is universal correlation among nodes, including those at the same hierarchical depth. Hyperparameter $c$ could be used to adjust the correlation. Once $c=0$, it goes back to the first setting. In fact, since the nodes have binary labels, introducing a large $c$ would tend to make many nodes simultaneously take the value 1. This is often unrealistic in authorization systems. Therefore, we set $c=0.5$.

The results for this setting are reported in Table \ref{tab:correlated_dag_results}. We can see that PATH still achieves the highest AUC and a reasonably high F1 score. It also preserves a zero violation rate without the need for any post-processing procedure or a constraint layer. Furthermore, the $p$-values indicate that the improvement in AUC over all competing methods remains statistically significant. For IC and HMCN-F, we can also observe a decline in AUC after post-processing.

\begin{table}[htbp]
\centering
\caption{Simulation study comparing the performance of different methods for predicting incomplete hierarchical authorization outcomes using AUC, F1 score, and the violation rate when all nodes are correlated. Best results are highlighted in bold. The $p$-value is computed from a two-sample $t$-test comparing the AUC of each baseline against the proposed method over 100 replicates. The suffix ``(post)'' denotes results after post-processing to enforce hierarchical consistency.}
\label{tab:correlated_dag_results}
\vspace{6pt}
\scalebox{0.9}{
\begin{tabular}{lcccccc}
\hline
Method  & AUC(95\% CI) & F1(95\% CI) & Violation Rate & $p$-value\\
\hline
IC & 0.8633 [0.8621, 0.8645]& 0.4164 [0.4136, 0.4191]& 0.0644 & $<10^{-4}$\\
HMCN-F & 0.8695 [0.8686, 0.8704] & 0.4185 [0.4162, 0.4209] & $<10^{-4}$ & $<10^{-4}$\\
IC(post) & 0.8221 [0.8201, 0.8241]& 0.4793 [0.4767, 0.4818]& 0.0000 & $<10^{-4}$\\
HMCN-F(post) & 0.8695 [0.8686, 0.8704] & 0.4185 [0.4162, 0.4209] & 0.0000 & $<10^{-4}$\\
CC &  0.8209 [0.8191, 0.8227]& \textbf{0.5154} [0.5128, 0.5180] & 0.0000 & $<10^{-4}$\\
C-HMCNN &  0.8691 [0.8682, 0.8700]& 0.4166 [0.4141, 0.4191] & 0.0000 & $<10^{-4}$\\
\textbf{PATH} & \textbf{0.8828} [0.8816, 0.8840] & 0.4989 [0.4961, 0.5018] & 0.0000 & --\\
\hline
\end{tabular}
}
\end{table}

\subsection{AI-Generated Synthetic User Study}

For the third experiment, we conduct an AI-generated synthetic user study to evaluate the performance of the proposed method under incomplete observations. Specifically, we prompt ChatGPT 5.5 to generate $500$ synthetic users with diverse backgrounds. The prompt we used is provided in the Section \ref{app:experimental-details} of Supplementary Material. We then simulate their responses to a sequential authorization questionnaire regarding their preferences for authorization to agentic AI systems. The resulting dataset consists of user profiles $X$ and their corresponding structured authorization responses $Y$. The user profile $X$ includes a collection of categorical variables representing demographic and contextual factors that are likely to influence authorization preferences in real-world human–AI interactions. These factors include age group, occupation type, technical savviness, privacy concern, risk tolerance, prior experience with AI agents, task type, requested autonomy level, time pressure, and domain sensitivity. All categorical variables are transformed using one-hot encoding, resulting in a total of $48$ input features.


The response variable $Y$ follows a fixed hierarchical authorization structure. Each node in the hierarchy corresponds to a specific authorization action, including communication access, file operations, payment permissions, smart-home control, healthcare access, or software development operations, while edges represent the dependency relationships among authorization nodes. For each user, the response records whether an individual authorization node is granted or denied. To better reflect realistic authorization behavior, users are also allowed to skip questions during the sequential authorization process. Once a question is skipped, all subsequent descendant questions are automatically treated as skipped. These skipped nodes are regarded as missing labels in the final dataset. The resulting authorization hierarchy contains a total of $167$ authorization nodes, and its missing rate is 83.04\%. The complete dataset of $500$ synthetic users is randomly divided into training, validation, and test sets consisting of $400$, $50$, and $50$ users, respectively.


On this dataset, we train the proposed method together with the same four competing methods (IC, CC, C-HMCNN and HMCN-F), following the same training procedure used in the previous experiment. Model performance is then evaluated on the test set using the same set of evaluation metrics, including AUC and F1 scores, and the binary violation rate measured prior to post-processing procedure. We also evaluate the AUC and F1 scores after post-processing for methods with nonnegative violation rates (i.e., IC and HMCN-F).


The experimental results are summarized in Table~\ref{tab:synthetic_authorization_dag_results}. PATH achieves the highest AUC and F1 score among all competing approaches, demonstrating superior predictive performance. CC, C-HMCNN, and the proposed method maintains a binary violation rate of $0$, indicating that all predicted labels satisfy the hierarchical constraints. In contrast, IC and HMCN-F yield substantially higher violation rates of $0.1189$ and $0.1421$, respectively, both exceeding their corresponding rates in the simulation studies. Meanwhile, both methods have a decline in AUC following post-processing. 
These findings further demonstrate that enforcing zero structural violations through post-processing may come at the cost of predictive performance, highlighting the advantage of methods that guarantee hierarchical consistency by construction without requiring such adjustments.

\begin{table}[htbp]
\centering
\caption{AI-generated synthetic user study comparing the performance of different methods for predicting incomplete hierarchical authorization outcomes. Best results are highlighted in bold. The suffix ``(post)'' denotes results after post-processing to enforce hierarchical consistency.}
\label{tab:synthetic_authorization_dag_results}
\vspace{6pt}
\scalebox{0.9}{
\begin{tabular}{lccccc}
\hline
Method & AUC & F1 & Violation Rate\\
\hline
IC & 0.6867 & 0.6974 & 0.1189 \\
HMCN-F & 0.7474 & 0.7182 & 0.1421\\
IC(post) & 0.6571 & 0.6228 & 0.0000 \\
HMCN-F(post) & 0.7380 & 0.7011 & 0.0000\\
CC & 0.6665 & 0.6039 & 0.0000 \\
C-HMCNN &  0.7217 & 0.6981 & 0.0000 \\
\textbf{PATH} &  \textbf{0.7692} & \textbf{0.7560} & 0.0000 \\
\hline
\end{tabular}
}
\end{table}

\section{Real Data Analysis}
In this section, we apply our proposed method to a publicly available dataset of users' data-sharing preferences for AI agents\footnote{The dataset is publicly available at \url{https://github.com/llm-platform-security/ai-agent-permissions}.} \citep{wu2026towards} to evaluate the performance of our method in real scenario. This dataset includes permission responses from users collected across 65 scenarios in eight application domains, such as health and fitness, finance, shopping etc. The raw data contains 201 valid participant records. 
The dataset also includes participants' background information, such as demographic characteristics, familiarity with AI, frequency of AI use, trust in AI, and perceived importance of privacy. To protect participants' privacy, this information is processed into 19 input features. Details are provided in Section \ref{app:experimental-details} of Supplementary Material.

The original dataset contains four response categories: ``Yes, always share,'' ``Yes, but ask me first,'' ``No, but ask me first'' and ``No, never share.'' As our experiment focuses on automatic sharing permissions, we specifically assign a label of 1 to ``Yes, always share'', and a label of 0 to the remaining three categories. Accordingly, a negative label (0) indicates that the user has not granted permission for automatic sharing, but does not necessarily imply that the user would refuse to share if explicitly asked. Unobserved permission responses are treated as missing rather than negative labels. The resulting authorization hierarchy contains a total of 447 authorization nodes, with an overall missing rate of 78.18\%.

For comparison, we still compare PATH with the same four competing methods (IC, CC, C-HMCNN, HMCN-F). We partition the dataset into 150 training participants and 51 test participants. Using the 150 training samples, we generate an additional 500 synthetic participants via ChatGPT 5.5 (the prompt we used can be seen in Section \ref{app:experimental-details} of Supplementary Material), yielding an augmented training set of 650 samples for all methods. For evaluation, we report AUC and F1 scores on the same test set, using a fixed threshold of 0.5 for binary predictions. We also report the binary violation rate before post-processing and evaluate the AUC and F1 scores after post-processing for methods that have nonzero violation rates.



\begin{table}[htbp]
\centering
\caption{Real data analysis comparing the performance of different methods for predicting incomplete hierarchical authorization outcomes. Best results are highlighted in bold. The suffix ``(post)'' denotes results after post-processing to enforce hierarchical consistency.}
\label{tab:real_data_results}
\vspace{6pt}
\small
\begin{tabular}{lccc}
\hline
Method & AUC & F1  & Violation Rate\\
\hline
IC       & 0.5498 & 0.3596  & 0.1129\\
HMCN-F   & 0.6757 & 0.3027  & 0.0010\\
IC(post) & 0.5445 & 0.2997  & 0.0000\\
HMCN-F(post)   & 0.6751 & 0.3019  & 0.0000\\
CC       & 0.5209 & 0.2327  & 0.0000 \\
C-HMCNN  & \textbf{0.6778} & 0.2864  & 0.0000\\
\textbf{PATH} & 0.6773 & \textbf{0.4060}  & 0.0000\\
\hline
\end{tabular}
\end{table}

The results are reported in Table~\ref{tab:real_data_results}. PATH achieves the highest F1 score among all five methods and performs comparably to the best method in terms of AUC. These results show the effectiveness of the proposed method in predicting authorization outcomes in a real-world setting. In terms of hierarchical consistency, CC, C-HMCNN, and the proposed method all achieve zero binary violations. In contrast, IC has a substantially higher violation rate of 0.1129, while HMCN-F produces a small but nonzero violation rate of 0.0010. Although post-processing eliminates these violations, it also leads to reductions in both AUC and F1 scores, particularly for IC.

\section{Discussion}
\label{sec:discussion}

In this paper, we proposed a novel method, namely Parent-gated Authorization Transformer
for Hierarchies (PATH), for predicting hierarchical authorization structures in agentic AI systems. PATH leverages a Transformer architecture with a hierarchy-structured attention mask and a parent-gated probability formulation to ensure hierarchical consistency in the predictions. This design also ensures that the model actually learns the probability chain rule that forms the hierarchical structure. In addition, we also adopt a depth-level embedding that accommodate incomplete authorization observations and capture the correlation among nodes at the same hierarchical depth. 
Theoretically, we establish a prediction error bound that characterizes the effects of approximation error, estimation error, optimization error, and missingness-induced bias. We further establish statistical consistency and derive convergence guarantees for the proposed optimization algorithm. 
We evaluated our method on simulated, AI-generated user data and real-world data analysis. The results demonstrate strong performance in prediction while maintaining zero hierarchy violations.
For future work, we will extend the proposed framework to incorporate more complex inputs, such as multimodal data and users' interaction histories with AI agents. Furthermore, we will also incorporate industry-scale datasets and integrate the proposed approach into real-world products for deployment and valuation.

\newpage
\section*{Supplementary Material: Structure-Aware Transformers for Hierarchical Authorization in Personal AI Agents}

\vspace{1em}
\appendix
\setcounter{page}{1}
\counterwithout{equation}{section}
\setcounter{table}{0}
\setcounter{equation}{0}
\setcounter{theorem}{0}
\renewcommand{\theequation}{S.\arabic{equation}}
\renewcommand{\thetable}{S.\arabic{table}}
\renewcommand{\thetheorem}{S.\arabic{theorem}}
\newenvironment{proof}
{\par\noindent\textbf{Proof:}\ }
{\hfill$\square$\par}

\section{Conditions and Proofs of Theorem \ref{ms:main-theorem} and Corollaries \ref{ms:estimation-special}, \ref{ms:fidelity}, \ref{ms:consistency}}
\label{app:statistical-proofs}

\subsection{Proof of Theorem~\ref{ms:main-theorem}}
\begin{proof}
For $u\in[0,1]$ and $s\in(0,1)$, put
\[
 D_{\mathrm B}(u\|s)=u\log\frac us+(1-u)\log\frac{1-u}{1-s}.
\]
The endpoint convention makes this the Bernoulli KL divergence.
Conditioning on $X$ gives
\[
 \mathbb E[O_jY_j\mid X]=\pi_j(X)\nu_j(X),\qquad
 \mathbb E[O_j(1-Y_j)\mid X]=\pi_j(X)(1-\nu_j(X)).
\]
Hence, with $H(u)=-u\log u-(1-u)\log(1-u)$,
\begin{align}
 R_O(\nu)&=\mathbb E_X\left[\frac1m\sum_{j\in J}
       \pi_j(X)H(\nu_j(X))\right]\le\frac{|J|}{m}\log2,
 \label{app:entropy}\\
 R_O(p)-R_O(\nu)&=\mathbb E_X\left[\frac1m\sum_{j\in J}
       \pi_j(X)D_{\mathrm B}(\nu_j(X)\|p_j(X))\right].
 \label{app:observed-kl}
\end{align}
In particular, $\nu$ minimizes unrestricted observed risk and
$A_{O,\mathcal F}\ge0$.

Work on the event $\mathcal E_{n,\delta}=\{2\Delta_n\le T_{n,\delta}\}$,
intersected with the probability-one event in \eqref{ms:erm}.
For every $p\in\mathcal F$,
\[
 R_O(\widehat p)\le\widehat R_O(\widehat p)+\Delta_n
 \le\widehat R_O(p)+\xi+\Delta_n
 \le R_O(p)+\xi+2\Delta_n.
\]
Taking the infimum over $p$ and subtracting $R_O(\nu)$ yields
\begin{equation}
 0\le R_O(\widehat p)-R_O(\nu)
 \le A_{O,\mathcal F}+2\Delta_n+\xi\le C_{n,\delta}.
 \label{app:oracle}
\end{equation}
An attained minimizer is not needed.

The scalar inequality
\begin{equation}
 D_{\mathrm B}(u\|s)\ge2(u-s)^2
 \label{app:pinsker}
\end{equation}
follows by holding $s$ fixed: the divergence and its first derivative
in $u$ vanish at $u=s$, and its second derivative is
$1/[u(1-u)]\ge4$. Taylor's formula with integral remainder proves the
inequality on $(0,1)$, and continuity gives $u=0,1$.
Using \eqref{app:observed-kl}, positivity, and \eqref{app:oracle},
\begin{equation}
 2\pi_{\min}d_2(\widehat p,\nu)^2
 \le R_O(\widehat p)-R_O(\nu)\le C_{n,\delta}.
 \label{app:selected-error}
\end{equation}
The $L^2$ triangle inequality gives
\[
 d_2(\widehat p,\eta)\le d_2(\widehat p,\nu)+d_2(\nu,\eta)
 \le\sqrt{\frac{C_{n,\delta}}{2\pi_{\min}}}+d_2(\nu,\eta).
\]
Both conclusions hold on the same event of probability at least
$1-\delta$. No identification assumption is used.
\end{proof}


\subsection{Proof of Corollary~\ref{ms:estimation-special}}
\begin{proof}
Write $g_p(Z)=m^{-1}\sum_{j\in J}O_j\ell(p_j(X),Y_j)$,
$Pg=\mathbb E g(Z)$, and $P_ng=n^{-1}\sum_{i=1}^ng(Z_i)$.
Then $\Delta_n=\sup_{p\in\mathcal F}|(P_n-P)g_p|$.
For $s\in[\varepsilon,1-\varepsilon]$ and $y\in\{0,1\}$,
\[
 0\le\ell(s,y)\le B_\varepsilon:=\log(1/\varepsilon),
 \qquad |\partial_s\ell(s,y)|\le\varepsilon^{-1}.
\]
Since $|J|/m\le1$, the user-level losses are in
$[0,B_\varepsilon]$. The mean value theorem gives
\begin{align*}
 |g_{p_\theta}(Z)-g_{p_{\theta'}}(Z)|
 &\le\frac1{m\varepsilon}\sum_{j\in J}O_j
                     |p_{\theta,j}(X)-p_{\theta',j}(X)|\\
 &\le\frac{K_G}{\varepsilon}\|q_\theta-q_{\theta'}\|_\infty.
\end{align*}
Cover the gate class at radius $u=\varepsilon/(K_Gn)\in(0,1]$.
Mapping each center to its loss gives a uniform loss cover of radius
$1/n$, with centers $g_1,\ldots,g_M$ in the loss class, where
\[
 \log M\le c\log(MK_Gn/\varepsilon).
\]
For a fixed center, the bounded-variable concentration inequality
of Hoeffding~\citep{hoeffding1963} yields
\[
 \mathbb P\{|(P_n-P)g_a|>t\}\le
       2\exp(-2nt^2/B_\varepsilon^2).
\]
The independent units here are users; node independence is unnecessary.
A union bound with
\[
 t=B_\varepsilon\sqrt{
      \frac{c\log(MK_Gn/\varepsilon)+\log(2/\delta)}{2n}}
\]
controls all centers on one event of probability at least $1-\delta$.
On that event, select for each $g_p$ a center with
$\|g_p-g_a\|_\infty\le1/n$. Then
\[
 |(P_n-P)g_p|\le|(P_n-P)g_a|
             +P_n|g_p-g_a|+P|g_p-g_a|\le t+2/n.
\]
Taking the supremum and multiplying by two proves
\eqref{ms:explicit-estimation}.

For measurability, the union of finite gate covers at radii $2^{-r}$,
$r\ge1$, induces a countable uniformly dense collection of losses.
The map $g\mapsto|(P_n-P)g|$ is $2$-Lipschitz in uniform norm, so
its supremum equals that over this countable collection. Thus
$\Delta_n$ is measurable. Bounded losses establish integrability.

Finally, for $n\ge2$ and $\delta=n^{-2}$, the numerator inside the
square root is $O(\log n)$ with all class and graph constants fixed.
This proves \eqref{ms:estimation-rate}.
\end{proof}

\subsection{Proof of Corollary~\ref{ms:fidelity}}
\begin{proof}
Hold the fitted predictor fixed. Roots have zero error and
\[
 \mathbf1\left\{\max_{j\in V}|\widehat p_j(x)-\eta_j(x)|\ge t\right\}
 \le\frac1{t^2}\sum_{j\in J}|\widehat p_j(x)-\eta_j(x)|^2.
\]
Integrating over an independent fresh $X$ gives the conditional bound
$m d_2(\widehat p,\eta)^2/t^2$. On the common event of
Theorem~\ref{ms:main-theorem}, this is at most
\[
 \frac m{t^2}\left(
       \sqrt{\frac{C_{n,\delta}}{2\pi_{\min}}}+d_2(\nu,\eta)\right)^2.
\]
The complement has probability at most $\delta$, and its conditional
test probability is at most one. Integrating over training proves
\eqref{ms:fresh-tail}, also using the universal bound of one.

For \eqref{ms:training-tail}, take $\delta=n^{-2}$. On the good event,
$d_2(\widehat p,\eta)$ is at most the theorem's bound, which is strictly
smaller than that bound plus $1/n$. Thus the displayed tail event is
contained in the failure event. The positive slack justifies using
$\ge$ in the tail event.
\end{proof}

\subsection{Proof of Corollary~\ref{ms:consistency}}
\begin{proof}
Apply \eqref{app:selected-error} at sample size $n$. On an event of
probability at least $1-\delta_n$,
\[
 d_2(\widehat p_n,\nu_n)\le
        \sqrt{\frac{C_{n,\delta_n}}{2\pi_{\min,n}}}=:a_n\longrightarrow0.
\]
For any fixed $t>0$, eventually $a_n<t$, so
$\mathbb P\{d_2(\widehat p_n,\nu_n)>t\}\le\delta_n\to0$.
If $b_n=d_2(\nu_n,\eta)\to0$, the triangle inequality yields
$d_2(\widehat p_n,\eta)\le a_n+b_n\to0$ on the same event.
This proves integrated probability consistency. For a fresh user,
Corollary~\ref{ms:fidelity} gives
\[
 \mathbb P\left\{\max_{j\in V}|\widehat p_{n,j}(X)-\eta_j(X)|\ge t\right\}
 \le\delta_n+\frac m{t^2}(a_n+b_n)^2\longrightarrow0.
\]
Here $m$ and the complete-data covariate law are fixed. No independence
between samples of different sizes is required.
\end{proof}

\section{The Convergence of the Optimization Algorithm}
\label{app:optimization-proofs}

\subsection{Objective, Assumptions, and Convergence Theorem}
\label{app:adam-assumptions}
Fix $\mathcal D_n$ and an initialization $\theta_0\in\Theta$. Throughout
this optimization section the expectation and probability condition on these fixed
quantities, unless explicitly stated otherwise. Define
\begin{equation}
 F_n(\theta)=\widehat R_O(p_\theta),\qquad
 F_n^\star=\inf_{\theta\in\Theta}F_n(\theta),\qquad
 \xi_k=F_n(\theta_k)-F_n^\star.
 \label{app:objective}
\end{equation}
In particular, $m=|V|$ and the class in the infimum is the same class
as in Theorem~\ref{ms:main-theorem}. The following are sufficient
conditions. They must hold for the same infinite run with constants
independent of the stopping iteration.

\begin{enumerate}[label=\textnormal{(O\arabic*)},leftmargin=*]
\item \textbf{Objective agreement and containment.}
There is an open convex set $U\subseteq\mathbb R^D$ containing
$\Theta$ and a continuously differentiable function $J_n:U\to\mathbb R$
with $J_n=F_n$ on $\Theta$. The infimum $F_n^\star$ is finite, and all
iterates of the update below belong to $\Theta$ almost surely. No
projection is included in that update. An attained minimizer is not
required.
\item \textbf{Smoothness and gradient domination.}
For fixed $L,\mu>0$,
\begin{align}
 \|\nabla J_n(u)-\nabla J_n(v)\|&\le L\|u-v\|
       &&(u,v\in U),\notag\\
 \tfrac12\|\nabla J_n(\theta)\|^2
 &\ge\mu\{F_n(\theta)-F_n^\star\}
       &&(\theta\in\Theta).
 \label{app:pl}
\end{align}
The second inequality is the Polyak--\L{}ojasiewicz (PL) condition
relative to the infimum over the entire comparison class. It suffices
for this inequality to hold at all iterates, including initialization.
A local-minimum comparator would not control \eqref{app:objective}.
\item \textbf{A conditionally unbiased, bounded gradient oracle.}
Let $(\mathcal H_t)_{t\ge0}$ be an increasing filtration containing
the initialization and all gradient draws through update $t$.
The iterate $\theta_{t-1}$ is $\mathcal H_{t-1}$-measurable and
the gradient $g_t$ is $\mathcal H_t$-measurable, with
\begin{equation}
 \mathbb E[g_t\mid\mathcal H_{t-1}]=\nabla J_n(\theta_{t-1}),
 \qquad \|g_t\|\le B<\infty\quad\text{almost surely},
 \qquad t\ge1,
 \label{app:oracle-assumption}
\end{equation}
where $B>0$. Norms in this appendix are Euclidean. The bound is along
the run, not a global bound on gradients at every parameter in $U$.
\item \textbf{Schedule and returned iterate.}
Choose $0\le\beta<1$, $0<c<1$, and $\epsilon_{\rm A}>0$, and put
\begin{equation}
 A_{\rm opt}=\frac{\sqrt{B^2+\epsilon_{\rm A}}}{(1-\beta)\mu},
 \qquad a_t=\frac{A_{\rm opt}}{t+1},\qquad \rho_t=\frac ct.
 \label{app:adam-schedule}
\end{equation}
Starting from $M_0=W_0=0$, use
\begin{align}
 M_t&=\beta M_{t-1}+(1-\beta)g_t,\notag\\
 W_t&=(1-\rho_t)W_{t-1}+\rho_t g_t^{\odot2},\notag\\
 \theta_t&=\theta_{t-1}-a_t d_t,
 \qquad d_t=\frac{M_t}{\sqrt{W_t+\epsilon_{\rm A}}}.
 \label{app:adam-update}
\end{align}
All vector operations in the denominator are coordinatewise. Return
the last iterate $\theta_k$ after the chosen deterministic number of
updates. The stabilizer is inside the square root and there is no
moment bias correction.
\end{enumerate}
The scalar $\epsilon_{\rm A}$ is distinct from the statistical output
floor $\varepsilon$. Similarly, $A_{\rm opt}$ is unrelated to the
covering-number constant $A$, and $B$ is unrelated to the graph $G$.

The update and rate argument specialize the scheduled Adam analysis
of Liang et al.~\cite[Theorem 7 and Appendix C.5]{liang2025convergence}. Then we can declare the following theorem.
\begin{theorem}
\label{thm:main}
Fix the training sample and initialization, define
$F_n(\theta)=\widehat R_O(p_\theta)$ and
$F_n^\star=\inf_{\theta\in\Theta}F_n(\theta)$, and impose
conditions \textnormal{(O1)--(O4)} in
Appendix~\ref{app:adam-assumptions}. Run \eqref{app:adam-update} and return
$\theta_k$ after $k$ updates. For
$\xi_k=F_n(\theta_k)-F_n^\star$, there is a finite constant $C$,
independent of $k$, such that for every deterministic integer $k\ge1$,
\begin{equation*}
 \mathbb E_{\mathrm{alg}}[\xi_k\mid\mathcal D_n,\theta_0]
 \le\frac{C}{k+1}.
 \label{ms:adam-mean}
\end{equation*}
For every $0<\delta_{\mathrm{opt}}<1$ and every such $k$,
\begin{equation*}
 \mathbb P_{\mathrm{alg}}\left\{
 \xi_k\le\frac{C}{\delta_{\mathrm{opt}}(k+1)}
 \,\middle|\,\mathcal D_n,\theta_0\right\}\ge1-\delta_{\mathrm{opt}}.
 \label{ms:adam-tail}
\end{equation*}
Along the same infinite run, $\xi_k\to0$ almost surely with respect
to optimizer randomness. Here $C$ may depend on the fixed sample and
initialization through the assumption constants. No relationship between
$k$ and $n$ is required for this conditional assertion.
\end{theorem}

\subsection{Proof of Theorem~\ref{thm:main}}
Define the finite constants
\begin{equation}
\begin{aligned}
 S&=B^2+\epsilon_{\rm A},\qquad
 h=\frac{1-\beta}{\sqrt S},\qquad
 b=\frac{\beta LB^2}{\epsilon_{\rm A}},\qquad
 e=\frac{\sqrt D\,B^4}{\epsilon_{\rm A}^{3/2}},\\
 C_4&=bA_{\rm opt}^2
 =\frac{\beta LB^2 S}{(1-\beta)^2\mu^2\epsilon_{\rm A}},\\
 C_5&=eA_{\rm opt}
 =\frac{B^4\sqrt{DS}}{(1-\beta)\mu\epsilon_{\rm A}^{3/2}},\\
 C_6&=\frac{LB^2A_{\rm opt}^2}{2\epsilon_{\rm A}}
 =\frac{LB^2 S}{2(1-\beta)^2\mu^2\epsilon_{\rm A}},\qquad
 C=\frac{C_4+cC_5}{(1-\beta)^2}+C_6.
\end{aligned}
\label{app:adam-constants}
\end{equation}
The parameter dimension is $D$, as in Definition~\ref{ms:ass2}.
\begin{proof}
Write $u_t=\nabla J_n(\theta_{t-1})$ and $z_t=\mathbb E\xi_t$.
By conditional Jensen's inequality and \eqref{app:oracle-assumption},
$\|u_t\|\le B$ almost surely. The moment recursions give
\begin{equation}
 \|M_t\|\le B,\qquad 0\le W_{t,i}\le B^2,\qquad
 \|d_t\|\le B/\sqrt{\epsilon_{\rm A}}.
 \label{app:moment-bounds}
\end{equation}
All gaps are integrable: the initial gap is finite, and smoothness
and these bounds give finite deterministic bounds on each change.

\paragraph{Step 1: control the correlation with the adaptive direction.}
The second-moment recursion gives
$|W_{t,i}-W_{t-1,i}|\le\rho_t B^2$. The derivative of
$s\mapsto(s+\epsilon_{\rm A})^{-1/2}$ has absolute value at most
$1/(2\epsilon_{\rm A}^{3/2})$ on $[0,\infty)$. Consequently,
\begin{equation}
 \left\|d_t-\frac{M_t}{\sqrt{W_{t-1}+\epsilon_{\rm A}}}\right\|
 \le\frac{\sqrt D\,B^3}{\epsilon_{\rm A}^{3/2}}\rho_t.
 \label{app:denominator-change}
\end{equation}
The bound deliberately drops a factor $1/2$. Using the preceding
second moment makes the denominator measurable before the new gradient
is drawn. In particular,
\[
 \mathbb E\left\langle u_t,
       \frac{g_t}{\sqrt{W_{t-1}+\epsilon_{\rm A}}}\right\rangle
 =\mathbb E\sum_{i=1}^D\frac{u_{t,i}^2}
                              {\sqrt{W_{t-1,i}+\epsilon_{\rm A}}}
 \ge\frac1{\sqrt S}\mathbb E\|u_t\|^2.
\]
For $t\ge2$, smoothness and \eqref{app:moment-bounds} imply
\[
 \left|\langle u_t-u_{t-1},d_{t-1}\rangle\right|
 \le L a_{t-1}\|d_{t-1}\|^2
 \le\frac{LB^2}{\epsilon_{\rm A}}a_{t-1}.
\]
Let $H_t=\mathbb E\langle u_t,d_t\rangle$ and $H_0=0$.
Combining these estimates with the first-moment recursion yields
\begin{equation}
 H_t\ge\beta H_{t-1}+h\mathbb E\|u_t\|^2
                         -b a_{t-1}-e\rho_t.
 \label{app:direction-recursion}
\end{equation}
For $t=1$, the old momentum is zero, so the same inequality is valid
with the auxiliary value $a_0=A_{\rm opt}$. This value is used only
as a bound, not as an extra algorithmic update.
Unfolding \eqref{app:direction-recursion} and discarding only the
nonnegative older gradient terms gives
\begin{equation}
 H_t\ge h\mathbb E\|u_t\|^2
 -\sum_{s=1}^t\beta^{t-s}(b a_{s-1}+e\rho_s).
 \label{app:direction-lower}
\end{equation}
This argument does not assume that $W_t$ is independent of $g_t$.

\paragraph{Step 2: obtain and solve a scalar gap recursion.}
The entire segment between consecutive iterates lies in $U$.
The smoothness inequality, objective agreement, and
\eqref{app:moment-bounds} give
\[
 z_t\le z_{t-1}-a_tH_t+\frac{LB^2}{2\epsilon_{\rm A}}a_t^2.
\]
By PL, $\mathbb E\|u_t\|^2\ge2\mu z_{t-1}$.
Since $A_{\rm opt}h\mu=1$, substituting
\eqref{app:direction-lower} and the schedules gives
\begin{equation}
 z_t\le\frac{t-1}{t+1}z_{t-1}
 +\frac{C_4+cC_5}{t+1}
        \sum_{s=1}^t\frac{\beta^{t-s}}s
 +\frac{C_6}{(t+1)^2}.
 \label{app:gap-recursion}
\end{equation}
Multiply by $t(t+1)$ and sum over $t=1,\ldots,k$.
The gap terms telescope, with zero coefficient on $z_0$, to give
\[
 k(k+1)z_k\le
 (C_4+cC_5)\sum_{t=1}^k t\sum_{s=1}^t\frac{\beta^{t-s}}s
 +C_6\sum_{t=1}^k\frac t{t+1}.
\]
For every integer $s\ge1$,
\[
 \sum_{t=s}^k t\beta^{t-s}
 \le\sum_{r=0}^{\infty}(s+r)\beta^r
 =\frac s{1-\beta}+\frac\beta{(1-\beta)^2}
 \le\frac s{(1-\beta)^2}.
\]
Reversing the finite double sum therefore bounds it by
$k/(1-\beta)^2$, while $\sum_{t=1}^k t/(t+1)\le k$.
Thus $z_k\le C/(k+1)$, proving \eqref{ms:adam-mean}.
For $\beta=0$, the same calculations use $\beta^0=1$ and
$\beta^r=0$ for $r\ge1$.

\paragraph{Step 3: the probability bound.}
The gaps are nonnegative. Markov's inequality gives
$\mathbb P\{\xi_k>r\}\le C/[r(k+1)]$ for $r>0$.
Taking $r=C/[\delta_{\mathrm{opt}}(k+1)]$ proves
\eqref{ms:adam-tail}.

\paragraph{Step 4: almost-sure convergence of the entire sequence.}
For any $a>0$,
\[
 \sum_{r=1}^{\infty}\mathbb P\{\xi_{r^2}>a\}
 \le\frac C a\sum_{r=1}^{\infty}\frac1{r^2+1}<\infty.
\]
The first Borel--Cantelli lemma, applied for $a=1,1/2,1/3,\ldots$,
gives $\xi_{r^2}\to0$ almost surely. To control the intervening iterates,
the two-sided smoothness remainder and \eqref{app:moment-bounds} imply
\begin{align}
 |\xi_t-\xi_{t-1}|
 &\le\frac{B^2}{\sqrt{\epsilon_{\rm A}}}a_t
       +\frac{LB^2}{2\epsilon_{\rm A}}a_t^2
 \le K a_t,\notag\\
 K&=\frac{B^2}{\sqrt{\epsilon_{\rm A}}}
       +\frac{LB^2A_{\rm opt}}{4\epsilon_{\rm A}},
 \label{app:increment}
\end{align}
where $a_t\le A_{\rm opt}/2$ for $t\ge1$.
For $r^2\le t\le(r+1)^2$,
\[
 |\xi_t-\xi_{r^2}|\le
 K\sum_{s=r^2+1}^{(r+1)^2}a_s
 \le K A_{\rm opt}\frac{2r+1}{r^2+2}\longrightarrow0.
\]
It follows that $\xi_t\to0$ almost surely along the full infinite run.
The expected rate alone would not justify this last conclusion;
the increment control is essential to this argument.
\end{proof}

\subsection{Scope and feasibility of the assumptions}
The theorem is conditional on (O1)--(O4); those properties do not
follow merely from using a Transformer or BCE. The oracle must target
the exact empirical objective in \eqref{app:objective}. For example,
independent uniform sampling of users with per-user loss
$m^{-1}\sum_jO_{ij}\ell(p_{\theta,j}(X_i),Y_{ij})$ gives an unbiased
gradient when differentiation is valid. A random denominator equal
to the number of observed labels in a batch generally changes this
expectation. Dropout, weight decay, and other penalties require
separate objective agreement or a transfer bound.

The update \eqref{app:adam-update} specifies a decreasing learning
rate, a second-moment coefficient $1-c/t$, no bias correction, and
an inside-square-root stabilizer. A fixed-learning-rate implementation
of standard bias-corrected Adam is not covered by this proof. Likewise,
random reshuffling without replacement does not by itself establish
conditional unbiasedness at each update. The condition $k\ge1$ is for
each deterministic stopping iteration; \eqref{ms:adam-tail} is not a
simultaneous confidence bound over all iterations or an arbitrary
data-dependent stopping rule.

The local formulation of smoothness is deliberate. A nonconstant
globally bounded objective on $\mathbb R^D$ cannot have both a globally
$L$-Lipschitz gradient and a global PL inequality with positive constant.
Indeed, for $x^+=x+\nabla J(x)/L$, the lower smoothness bound would give
\[
 J(x^+)-J^\star\ge J(x)-J^\star+\frac{\|\nabla J(x)\|^2}{2L}
 \ge(1+\mu/L)(J(x)-J^\star).
\]
Repeated ascent from a positive gap contradicts the global upper bound.
This matters for globally clipped losses. Our assumptions instead
require a suitable neighborhood and invariant trajectory, whose
existence must still be justified. ReLU or hard-clipping kinks can
prevent the assumed smoothness if a trajectory crosses them, and PL
can fail at nonglobal stationary points. Equality $J_n=F_n$ on a
restricted set alone does not guarantee equality of their gradients;
(O3) explicitly specifies the required gradient. Equality on an open
neighborhood of the trajectory is a sufficient way to align gradients.

\section{Experimental Details and Data-Generation Prompts}
\label{app:experimental-details}

This section provides detailed information of all the four experiments. That includes introduction of the structure of data model parameters for each experiment. Prompts used for AI generated experiment and real data analysis are also provided in this section.




\subsection{Models Parameters}
\label{app:model-configurations}

\paragraph{PATH}.
For a $p$-dimensional input, the user encoder is
$p\to128\to128$, with ReLU after each affine map and dropout $0.1$ between
the two maps. Its output is added to learned node and depth embeddings.
Two pre-normalized Transformer encoder layers use four attention heads,
hidden dimension 128, feed-forward dimension 512, GELU activation, and
dropout $0.1$. A node attends only to itself and its ancestors. A shared
linear head followed by a sigmoid gives a base score $q_{ij}$.
The root score is fixed at one.

For optimization, all Transformer runs use Adam, with learning rate $10^{-3}$, weight decay
$10^{-4}$, batch size 64, model initialization seed 42, positive-class
BCE weight 1, and hierarchy-penalty coefficient 0. For label prediction, the threshold is $0.5$ throughout all experiments.


\paragraph{HMCN-F}.
There is one local branch per graph depth. Every global hidden layer has
width 128, each local hidden layer has width 64, the activation is ReLU,
and dropout is $0.6$. Local and global probabilities are combined with
weight $\beta=0.5$. Training sums masked global BCE, the masked local BCE
terms, and a hierarchy penalty with coefficient $0.1$; the penalty is the
mean squared positive child-minus-parent difference of the fused probabilities,
averaged over users and direct edges without an observation mask. For optimization, Adam uses
learning rate $10^{-3}$ and weight decay $10^{-5}$, with batch size 64.
The original HMCN-F runs disable their own post-processing option.

\paragraph{C-HMCNN}.
The network has three affine layers, $p\to128\to128\to m$, where $m$ is the
number of graph nodes. The hidden activation is ReLU and dropout is $0.7$.
The model uses its missing-label-aware constraint loss and a max-constraint
layer at inference. Adam uses learning rate $10^{-4}$ and weight decay
$10^{-5}$, with batch size 4. HMCN-F and C-HMCNN fit feature preprocessing
on training data and select checkpoints by observed-node micro-average
precision (AP), using threshold $0.5$ for binary predictions.

\subsection{Data Introduction of Simulation Studies}
Table~\ref{dagonly:tab:simulation-complete-adjacency} lists all 171 nodes
and all of their immediate children for the DAG structure used in simulation studies. Each entry includes the saved node
ID and full node name. Repeated children in different rows have multiple parents;
$\varnothing$ denotes a terminal node with no children.

{\small\renewcommand{\arraystretch}{1.05}
\begin{longtable}{@{}>{\raggedright\arraybackslash}p{.33\textwidth}>{\raggedright\arraybackslash}p{.60\textwidth}@{}}
\caption{Complete adjacency list for the fixed seed-2026 simulation DAG.
All 171 nodes are listed; the children enumerate exactly 278 directed
edges.}\label{dagonly:tab:simulation-complete-adjacency}\\
\toprule
Parent ID and name & Immediate children: IDs and names \\
\midrule
\endfirsthead
\multicolumn{2}{@{}l}{\textit{Table \thetable\ continued}}\\
\toprule
Parent ID and name & Immediate children: IDs and names \\
\midrule
\endhead
\midrule
\multicolumn{2}{r@{}}{\textit{Continued on next page}}\\
\endfoot
\bottomrule
\endlastfoot
0:~\texttt{SystemAccess} & 1:~\texttt{Finance}, 2:~\texttt{HR}, 3:~\texttt{Research}, 4:~\texttt{IT}, 5:~\texttt{Teaching}, 6:~\texttt{Operations}, 7:~\texttt{Procurement}, 8:~\texttt{StudentServices} \\
1:~\texttt{Finance} & 9:~\texttt{Finance\_\allowbreak Read}, 10:~\texttt{Finance\_\allowbreak Edit}, 11:~\texttt{Finance\_\allowbreak Approve}, 12:~\texttt{Finance\_\allowbreak Admin}, 95:~\texttt{Teaching\_\allowbreak Approve\_\allowbreak View}, 113:~\texttt{Procurement\_\allowbreak Read\_\allowbreak View} \\
2:~\texttt{HR} & 13:~\texttt{HR\_\allowbreak Read}, 14:~\texttt{HR\_\allowbreak Edit}, 15:~\texttt{HR\_\allowbreak Approve}, 16:~\texttt{HR\_\allowbreak Admin}, 41:~\texttt{Finance\_\allowbreak Read\_\allowbreak View}, 102:~\texttt{Operations\_\allowbreak Read\_\allowbreak Create} \\
3:~\texttt{Research} & 17:~\texttt{Research\_\allowbreak Read}, 18:~\texttt{Research\_\allowbreak Edit}, 19:~\texttt{Research\_\allowbreak Approve}, 20:~\texttt{Research\_\allowbreak Admin}, 111:~\texttt{Operations\_\allowbreak Admin\_\allowbreak Create}, 117:~\texttt{Procurement\_\allowbreak Edit\_\allowbreak Create}, 139:~\texttt{RiskReview\_\allowbreak 3} \\
4:~\texttt{IT} & 21:~\texttt{IT\_\allowbreak Read}, 22:~\texttt{IT\_\allowbreak Edit}, 23:~\texttt{IT\_\allowbreak Approve}, 24:~\texttt{IT\_\allowbreak Admin} \\
5:~\texttt{Teaching} & 25:~\texttt{Teaching\_\allowbreak Read}, 26:~\texttt{Teaching\_\allowbreak Edit}, 27:~\texttt{Teaching\_\allowbreak Approve}, 28:~\texttt{Teaching\_\allowbreak Admin}, 62:~\texttt{HR\_\allowbreak Admin\_\allowbreak View}, 65:~\texttt{Research\_\allowbreak Read\_\allowbreak View}, 82:~\texttt{IT\_\allowbreak Edit\_\allowbreak Update}, 84:~\texttt{IT\_\allowbreak Approve\_\allowbreak Create}, 124:~\texttt{Procurement\_\allowbreak Admin\_\allowbreak Update}, 133:~\texttt{StudentServices\_\allowbreak Approve\_\allowbreak Update}, 138:~\texttt{PeopleAnalytics\_\allowbreak 2} \\
\end{longtable}}

\subsection{Data Introduction for AI-generated user study}
\label{app:ai-user-study}
The dataset contains 500 synthetic user records,
\texttt{U0001}--\texttt{U0500}, obtained through a ChatGPT generation
conversation. Each record stores a user profile $X$ and a permission DAG
response $Y$. User profile $X$ includes Eleven categorical profile variables, producing 48 one-hot
features: age group (5 categories), occupation (8), technical savviness
(3), privacy concern (3), risk tolerance (3), trust in the agent (3), prior
agent use (4), task type (10), requested autonomy (3), time pressure (3),
and domain sensitivity (3).

For the permission DAG $Y$, it has 167 nodes, 171 edges, 68 leaves with maximum depth six. Node 0 is the root;
the other IDs follow lexicographic order of the canonical node names.
The eight top-level branches are cloud collaboration, communication,
financial operations, healthcare services, home automation, planning,
research workflow, and social platform. The full information is listed in Table \ref{tab:complete_dag}.

{\small\renewcommand{\arraystretch}{1.05}
\begin{longtable}{@{}>{\raggedright\arraybackslash}p{.33\textwidth}>{\raggedright\arraybackslash}p{.60\textwidth}@{}}
\caption{Complete adjacency list for the AI synthetic-data modeling DAG, ordered by numeric node ID. Children within each row are also ordered by numeric ID. All 167 nodes are listed; the children enumerate exactly 171 directed edges. The artificial root marker $(-1,0)$ is excluded.}\label{dagonly:tab:ai-dag-adjacency}\\
\toprule
Parent ID and name & Immediate children: IDs and names \\
\midrule
\endfirsthead
\multicolumn{2}{@{}l}{\textit{Table \thetable\ continued}}\\
\toprule
Parent ID and name & Immediate children: IDs and names \\
\midrule
\endhead
\midrule
\multicolumn{2}{r@{}}{\textit{Continued on next page}}\\
\endfoot
\bottomrule
\endlastfoot
0:~\texttt{root} & 19:~\texttt{cloud\_\allowbreak collaboration}\newline 24:~\texttt{communication}\newline 55:~\texttt{financial\_\allowbreak operations}\newline 57:~\texttt{healthcare\_\allowbreak services}\newline 58:~\texttt{home\_\allowbreak automation}\newline 94:~\texttt{planning}\newline 124:~\texttt{research\_\allowbreak workflow}\newline 140:~\texttt{social\_\allowbreak platform} \\
1:~\texttt{account\_\allowbreak overview} & 159:~\texttt{view\_\allowbreak balance} \\
2:~\texttt{analysis} & 142:~\texttt{summarize\_\allowbreak papers} \\
3:~\texttt{appointment\_\allowbreak access} & 102:~\texttt{read\_\allowbreak appointments}\newline 107:~\texttt{read\_\allowbreak events} \\
4:~\texttt{bank\_\allowbreak account} & 1:~\texttt{account\_\allowbreak overview} \\
5:~\texttt{bill\_\allowbreak access} & 103:~\texttt{read\_\allowbreak bills} \\
6:~\texttt{billing} & 7:~\texttt{bills}\newline 91:~\texttt{payment\_\allowbreak processing} \\
7:~\texttt{bills} & 5:~\texttt{bill\_\allowbreak access} \\
8:~\texttt{booking} & 73:~\texttt{make\_\allowbreak payment}\newline 126:~\texttt{reserve\_\allowbreak itinerary} \\
9:~\texttt{branch\_\allowbreak management} & 31:~\texttt{create\_\allowbreak branch} \\
10:~\texttt{calendar} & 47:~\texttt{event\_\allowbreak access}\newline 48:~\texttt{event\_\allowbreak management} \\
11:~\texttt{camera} & 69:~\texttt{live\_\allowbreak monitoring}\newline 117:~\texttt{recording} \\
12:~\texttt{cancel\_\allowbreak event} & \(\varnothing\) \\
13:~\texttt{card\_\allowbreak access} & 164:~\texttt{view\_\allowbreak saved\_\allowbreak cards} \\
14:~\texttt{change\_\allowbreak thermostat} & \(\varnothing\) \\
15:~\texttt{claim\_\allowbreak access} & 160:~\texttt{view\_\allowbreak claims} \\
16:~\texttt{claim\_\allowbreak operations} & 141:~\texttt{submit\_\allowbreak claim} \\
17:~\texttt{close\_\allowbreak issue} & \(\varnothing\) \\
18:~\texttt{cloud\_\allowbreak access} & 106:~\texttt{read\_\allowbreak drive\_\allowbreak files} \\
19:~\texttt{cloud\_\allowbreak collaboration} & 20:~\texttt{cloud\_\allowbreak storage}\newline 37:~\texttt{development\_\allowbreak workspace}\newline 70:~\texttt{local\_\allowbreak file\_\allowbreak system} \\
20:~\texttt{cloud\_\allowbreak storage} & 18:~\texttt{cloud\_\allowbreak access}\newline 137:~\texttt{sharing}\newline 156:~\texttt{upload\_\allowbreak operations} \\
21:~\texttt{code\_\allowbreak contribution} & 23:~\texttt{commit\_\allowbreak changes}\newline 87:~\texttt{open\_\allowbreak pull\_\allowbreak request} \\
22:~\texttt{code\_\allowbreak repository} & 9:~\texttt{branch\_\allowbreak management}\newline 21:~\texttt{code\_\allowbreak contribution}\newline 119:~\texttt{repository\_\allowbreak access}\newline 120:~\texttt{repository\_\allowbreak administration} \\
23:~\texttt{commit\_\allowbreak changes} & \(\varnothing\) \\
24:~\texttt{communication} & 27:~\texttt{contacts}\newline 45:~\texttt{email}\newline 82:~\texttt{messaging} \\
25:~\texttt{communication\_\allowbreak messages} & 134:~\texttt{send\_\allowbreak messages} \\
26:~\texttt{contact\_\allowbreak access} & 46:~\texttt{email\_\allowbreak contacts}\newline 105:~\texttt{read\_\allowbreak contacts} \\
27:~\texttt{contacts} & 26:~\texttt{contact\_\allowbreak access} \\
28:~\texttt{content\_\allowbreak access} & 113:~\texttt{read\_\allowbreak posts} \\
29:~\texttt{content\_\allowbreak drafting} & 42:~\texttt{draft\_\allowbreak post} \\
30:~\texttt{content\_\allowbreak removal} & 35:~\texttt{delete\_\allowbreak post} \\
31:~\texttt{create\_\allowbreak branch} & \(\varnothing\) \\
32:~\texttt{create\_\allowbreak event} & \(\varnothing\) \\
33:~\texttt{create\_\allowbreak issue} & \(\varnothing\) \\
34:~\texttt{delete\_\allowbreak files} & \(\varnothing\) \\
35:~\texttt{delete\_\allowbreak post} & \(\varnothing\) \\
36:~\texttt{destructive\_\allowbreak commands} & 34:~\texttt{delete\_\allowbreak files} \\
37:~\texttt{development\_\allowbreak workspace} & 22:~\texttt{code\_\allowbreak repository}\newline 49:~\texttt{execution\_\allowbreak environment} \\
38:~\texttt{device\_\allowbreak control} & 92:~\texttt{physical\_\allowbreak security}\newline 138:~\texttt{smart\_\allowbreak devices}\newline 158:~\texttt{use\_\allowbreak home\_\allowbreak location} \\
39:~\texttt{device\_\allowbreak monitoring} & \(\varnothing\) \\
40:~\texttt{discovery} & 132:~\texttt{search\_\allowbreak papers} \\
41:~\texttt{download\_\allowbreak papers} & \(\varnothing\) \\
42:~\texttt{draft\_\allowbreak post} & \(\varnothing\) \\
43:~\texttt{draft\_\allowbreak reply} & \(\varnothing\) \\
44:~\texttt{drafting} & 43:~\texttt{draft\_\allowbreak reply} \\
45:~\texttt{email} & 44:~\texttt{drafting}\newline 135:~\texttt{sending} \\
46:~\texttt{email\_\allowbreak contacts} & \(\varnothing\) \\
47:~\texttt{event\_\allowbreak access} & 107:~\texttt{read\_\allowbreak events} \\
48:~\texttt{event\_\allowbreak management} & 12:~\texttt{cancel\_\allowbreak event}\newline 32:~\texttt{create\_\allowbreak event}\newline 83:~\texttt{modify\_\allowbreak event} \\
49:~\texttt{execution\_\allowbreak environment} & 88:~\texttt{package\_\allowbreak management}\newline 143:~\texttt{system\_\allowbreak commands}\newline 144:~\texttt{terminal}\newline 145:~\texttt{testing} \\
50:~\texttt{file\_\allowbreak access} & 115:~\texttt{read\_\allowbreak selected\_\allowbreak folder} \\
51:~\texttt{file\_\allowbreak deletion} & 34:~\texttt{delete\_\allowbreak files} \\
52:~\texttt{file\_\allowbreak organization} & 86:~\texttt{move\_\allowbreak files}\newline 118:~\texttt{rename\_\allowbreak files} \\
53:~\texttt{file\_\allowbreak system} & 50:~\texttt{file\_\allowbreak access}\newline 51:~\texttt{file\_\allowbreak deletion}\newline 52:~\texttt{file\_\allowbreak organization} \\
54:~\texttt{financial\_\allowbreak access} & 4:~\texttt{bank\_\allowbreak account} \\
55:~\texttt{financial\_\allowbreak operations} & 6:~\texttt{billing}\newline 54:~\texttt{financial\_\allowbreak access}\newline 64:~\texttt{investment\_\allowbreak management}\newline 90:~\texttt{payment}\newline 148:~\texttt{transactions} \\
56:~\texttt{health\_\allowbreak portal} & 3:~\texttt{appointment\_\allowbreak access}\newline 77:~\texttt{medical\_\allowbreak records}\newline 99:~\texttt{provider\_\allowbreak communication} \\
57:~\texttt{healthcare\_\allowbreak services} & 56:~\texttt{health\_\allowbreak portal}\newline 61:~\texttt{insurance} \\
58:~\texttt{home\_\allowbreak automation} & 38:~\texttt{device\_\allowbreak control}\newline 39:~\texttt{device\_\allowbreak monitoring}\newline 72:~\texttt{location\_\allowbreak services}\newline 85:~\texttt{monitoring} \\
59:~\texttt{inbox\_\allowbreak access} & 108:~\texttt{read\_\allowbreak inbox}\newline 129:~\texttt{search\_\allowbreak email} \\
60:~\texttt{install\_\allowbreak packages} & \(\varnothing\) \\
61:~\texttt{insurance} & 62:~\texttt{insurance\_\allowbreak portal} \\
62:~\texttt{insurance\_\allowbreak portal} & 15:~\texttt{claim\_\allowbreak access}\newline 16:~\texttt{claim\_\allowbreak operations} \\
63:~\texttt{investment\_\allowbreak account} & 95:~\texttt{portfolio\_\allowbreak access}\newline 146:~\texttt{trading} \\
64:~\texttt{investment\_\allowbreak management} & 63:~\texttt{investment\_\allowbreak account} \\
65:~\texttt{issue\_\allowbreak access} & 109:~\texttt{read\_\allowbreak issues} \\
66:~\texttt{issue\_\allowbreak management} & 17:~\texttt{close\_\allowbreak issue}\newline 33:~\texttt{create\_\allowbreak issue} \\
67:~\texttt{issue\_\allowbreak tracker} & 65:~\texttt{issue\_\allowbreak access}\newline 66:~\texttt{issue\_\allowbreak management} \\
68:~\texttt{literature\_\allowbreak search} & 2:~\texttt{analysis}\newline 40:~\texttt{discovery}\newline 122:~\texttt{research\_\allowbreak material\_\allowbreak access} \\
69:~\texttt{live\_\allowbreak monitoring} & 162:~\texttt{view\_\allowbreak live\_\allowbreak feed} \\
70:~\texttt{local\_\allowbreak file\_\allowbreak system} & 53:~\texttt{file\_\allowbreak system} \\
71:~\texttt{location} & \(\varnothing\) \\
72:~\texttt{location\_\allowbreak services} & 71:~\texttt{location}\newline 96:~\texttt{presence\_\allowbreak detection} \\
73:~\texttt{make\_\allowbreak payment} & \(\varnothing\) \\
74:~\texttt{media\_\allowbreak access} & 112:~\texttt{read\_\allowbreak photos} \\
75:~\texttt{media\_\allowbreak library} & 74:~\texttt{media\_\allowbreak access} \\
76:~\texttt{media\_\allowbreak management} & 75:~\texttt{media\_\allowbreak library} \\
77:~\texttt{medical\_\allowbreak records} & 110:~\texttt{read\_\allowbreak lab\_\allowbreak results} \\
78:~\texttt{merge\_\allowbreak to\_\allowbreak main} & \(\varnothing\) \\
79:~\texttt{message\_\allowbreak access} & 111:~\texttt{read\_\allowbreak messages} \\
80:~\texttt{message\_\allowbreak provider} & \(\varnothing\) \\
81:~\texttt{messages} & 25:~\texttt{communication\_\allowbreak messages}\newline 79:~\texttt{message\_\allowbreak access} \\
82:~\texttt{messaging} & 59:~\texttt{inbox\_\allowbreak access}\newline 81:~\texttt{messages} \\
83:~\texttt{modify\_\allowbreak event} & \(\varnothing\) \\
84:~\texttt{money\_\allowbreak movement} & 149:~\texttt{transfer\_\allowbreak money} \\
85:~\texttt{monitoring} & 11:~\texttt{camera} \\
86:~\texttt{move\_\allowbreak files} & \(\varnothing\) \\
87:~\texttt{open\_\allowbreak pull\_\allowbreak request} & \(\varnothing\) \\
88:~\texttt{package\_\allowbreak management} & 60:~\texttt{install\_\allowbreak packages} \\
89:~\texttt{pay\_\allowbreak bills} & \(\varnothing\) \\
90:~\texttt{payment} & 13:~\texttt{card\_\allowbreak access} \\
91:~\texttt{payment\_\allowbreak processing} & 89:~\texttt{pay\_\allowbreak bills} \\
92:~\texttt{physical\_\allowbreak security} & 154:~\texttt{unlock\_\allowbreak door} \\
93:~\texttt{place\_\allowbreak trade} & \(\varnothing\) \\
94:~\texttt{planning} & 10:~\texttt{calendar}\newline 98:~\texttt{project\_\allowbreak coordination}\newline 151:~\texttt{travel\_\allowbreak services} \\
95:~\texttt{portfolio\_\allowbreak access} & 163:~\texttt{view\_\allowbreak portfolio} \\
96:~\texttt{presence\_\allowbreak detection} & 158:~\texttt{use\_\allowbreak home\_\allowbreak location} \\
97:~\texttt{project\_\allowbreak access} & 114:~\texttt{read\_\allowbreak project\_\allowbreak files} \\
98:~\texttt{project\_\allowbreak coordination} & 67:~\texttt{issue\_\allowbreak tracker} \\
99:~\texttt{provider\_\allowbreak communication} & 80:~\texttt{message\_\allowbreak provider} \\
100:~\texttt{publish\_\allowbreak post} & \(\varnothing\) \\
101:~\texttt{publishing} & 100:~\texttt{publish\_\allowbreak post}\newline 157:~\texttt{upload\_\allowbreak photos} \\
102:~\texttt{read\_\allowbreak appointments} & \(\varnothing\) \\
103:~\texttt{read\_\allowbreak bills} & \(\varnothing\) \\
104:~\texttt{read\_\allowbreak code} & \(\varnothing\) \\
105:~\texttt{read\_\allowbreak contacts} & \(\varnothing\) \\
106:~\texttt{read\_\allowbreak drive\_\allowbreak files} & \(\varnothing\) \\
107:~\texttt{read\_\allowbreak events} & \(\varnothing\) \\
108:~\texttt{read\_\allowbreak inbox} & \(\varnothing\) \\
109:~\texttt{read\_\allowbreak issues} & \(\varnothing\) \\
110:~\texttt{read\_\allowbreak lab\_\allowbreak results} & \(\varnothing\) \\
111:~\texttt{read\_\allowbreak messages} & \(\varnothing\) \\
112:~\texttt{read\_\allowbreak photos} & \(\varnothing\) \\
113:~\texttt{read\_\allowbreak posts} & \(\varnothing\) \\
114:~\texttt{read\_\allowbreak project\_\allowbreak files} & \(\varnothing\) \\
115:~\texttt{read\_\allowbreak selected\_\allowbreak folder} & \(\varnothing\) \\
116:~\texttt{record\_\allowbreak video} & \(\varnothing\) \\
117:~\texttt{recording} & 116:~\texttt{record\_\allowbreak video} \\
118:~\texttt{rename\_\allowbreak files} & \(\varnothing\) \\
119:~\texttt{repository\_\allowbreak access} & 104:~\texttt{read\_\allowbreak code} \\
120:~\texttt{repository\_\allowbreak administration} & 78:~\texttt{merge\_\allowbreak to\_\allowbreak main} \\
121:~\texttt{research\_\allowbreak file\_\allowbreak system} & 97:~\texttt{project\_\allowbreak access} \\
122:~\texttt{research\_\allowbreak material\_\allowbreak access} & 41:~\texttt{download\_\allowbreak papers} \\
123:~\texttt{research\_\allowbreak notes} & 166:~\texttt{write\_\allowbreak notes} \\
124:~\texttt{research\_\allowbreak workflow} & 68:~\texttt{literature\_\allowbreak search}\newline 125:~\texttt{research\_\allowbreak workspace} \\
125:~\texttt{research\_\allowbreak workspace} & 121:~\texttt{research\_\allowbreak file\_\allowbreak system}\newline 123:~\texttt{research\_\allowbreak notes} \\
126:~\texttt{reserve\_\allowbreak itinerary} & \(\varnothing\) \\
127:~\texttt{run\_\allowbreak shell\_\allowbreak commands} & \(\varnothing\) \\
128:~\texttt{run\_\allowbreak tests} & \(\varnothing\) \\
129:~\texttt{search\_\allowbreak email} & \(\varnothing\) \\
130:~\texttt{search\_\allowbreak flights} & \(\varnothing\) \\
131:~\texttt{search\_\allowbreak hotels} & \(\varnothing\) \\
132:~\texttt{search\_\allowbreak papers} & \(\varnothing\) \\
133:~\texttt{send\_\allowbreak email} & \(\varnothing\) \\
134:~\texttt{send\_\allowbreak messages} & \(\varnothing\) \\
135:~\texttt{sending} & 133:~\texttt{send\_\allowbreak email} \\
136:~\texttt{share\_\allowbreak files} & \(\varnothing\) \\
137:~\texttt{sharing} & 136:~\texttt{share\_\allowbreak files} \\
138:~\texttt{smart\_\allowbreak devices} & 14:~\texttt{change\_\allowbreak thermostat}\newline 153:~\texttt{turn\_\allowbreak lights\_\allowbreak on\_\allowbreak off}\newline 161:~\texttt{view\_\allowbreak device\_\allowbreak status} \\
139:~\texttt{social\_\allowbreak account} & 28:~\texttt{content\_\allowbreak access}\newline 29:~\texttt{content\_\allowbreak drafting}\newline 30:~\texttt{content\_\allowbreak removal}\newline 101:~\texttt{publishing} \\
140:~\texttt{social\_\allowbreak platform} & 76:~\texttt{media\_\allowbreak management}\newline 139:~\texttt{social\_\allowbreak account} \\
141:~\texttt{submit\_\allowbreak claim} & \(\varnothing\) \\
142:~\texttt{summarize\_\allowbreak papers} & \(\varnothing\) \\
143:~\texttt{system\_\allowbreak commands} & 36:~\texttt{destructive\_\allowbreak commands}\newline 127:~\texttt{run\_\allowbreak shell\_\allowbreak commands} \\
144:~\texttt{terminal} & \(\varnothing\) \\
145:~\texttt{testing} & 128:~\texttt{run\_\allowbreak tests} \\
146:~\texttt{trading} & 93:~\texttt{place\_\allowbreak trade} \\
147:~\texttt{transaction\_\allowbreak records} & 165:~\texttt{view\_\allowbreak transactions} \\
148:~\texttt{transactions} & 73:~\texttt{make\_\allowbreak payment}\newline 84:~\texttt{money\_\allowbreak movement}\newline 147:~\texttt{transaction\_\allowbreak records} \\
149:~\texttt{transfer\_\allowbreak money} & \(\varnothing\) \\
150:~\texttt{travel\_\allowbreak search} & 130:~\texttt{search\_\allowbreak flights}\newline 131:~\texttt{search\_\allowbreak hotels} \\
151:~\texttt{travel\_\allowbreak services} & 152:~\texttt{travel\_\allowbreak websites} \\
152:~\texttt{travel\_\allowbreak websites} & 8:~\texttt{booking}\newline 150:~\texttt{travel\_\allowbreak search} \\
153:~\texttt{turn\_\allowbreak lights\_\allowbreak on\_\allowbreak off} & \(\varnothing\) \\
154:~\texttt{unlock\_\allowbreak door} & \(\varnothing\) \\
155:~\texttt{upload\_\allowbreak files} & \(\varnothing\) \\
156:~\texttt{upload\_\allowbreak operations} & 155:~\texttt{upload\_\allowbreak files}\newline 157:~\texttt{upload\_\allowbreak photos} \\
157:~\texttt{upload\_\allowbreak photos} & \(\varnothing\) \\
158:~\texttt{use\_\allowbreak home\_\allowbreak location} & \(\varnothing\) \\
159:~\texttt{view\_\allowbreak balance} & \(\varnothing\) \\
160:~\texttt{view\_\allowbreak claims} & \(\varnothing\) \\
161:~\texttt{view\_\allowbreak device\_\allowbreak status} & \(\varnothing\) \\
162:~\texttt{view\_\allowbreak live\_\allowbreak feed} & \(\varnothing\) \\
163:~\texttt{view\_\allowbreak portfolio} & \(\varnothing\) \\
164:~\texttt{view\_\allowbreak saved\_\allowbreak cards} & \(\varnothing\) \\
165:~\texttt{view\_\allowbreak transactions} & \(\varnothing\) \\
166:~\texttt{write\_\allowbreak notes} & \(\varnothing\) \\
\label{tab:complete_dag}
\end{longtable}}

\subsection{Data Introduction of Real Data Analysis}
\label{app:mixed-experiment}
Table~\ref{dagonly:tab:real-complete-adjacency} lists all 447 nodes and
all their immediate children for the dataset used in real data analysis.

{\small\renewcommand{\arraystretch}{1.05}
\begin{longtable}{@{}>{\raggedright\arraybackslash}p{.33\textwidth}>{\raggedright\arraybackslash}p{.60\textwidth}@{}}
\caption{Complete adjacency list for the real-data analysis DAG. All 447 nodes are listed; the children enumerate exactly 446 directed edges.}\label{dagonly:tab:real-complete-adjacency}\\
\toprule
Parent ID and name & Immediate children: IDs and names \\
\midrule
\endfirsthead
\multicolumn{2}{@{}l}{\textit{Table \thetable\ continued}}\\
\toprule
Parent ID and name & Immediate children: IDs and names \\
\midrule
\endhead
\midrule
\multicolumn{2}{r@{}}{\textit{Continued on next page}}\\
\endfoot
\bottomrule
\endlastfoot
0:~\texttt{All permissions} & 1:~\texttt{Health \& Fitness}\newline 2:~\texttt{Work \& Productivity}\newline 3:~\texttt{Finance}\newline 4:~\texttt{Shopping}\newline 5:~\texttt{Travel}\newline 6:~\texttt{Entertainment}\newline 7:~\texttt{Social}\newline 8:~\texttt{Smart Home} \\
1:~\texttt{Health \& Fitness} & 9:~\texttt{I want to lose weight. Can you create an at-home full-body workout plan for me?}\newline 10:~\texttt{Can you give me a summary of my health data from the past month?}\newline 11:~\texttt{How has my sleep duration changed over the past week?}\newline 12:~\texttt{I need a meal plan to help me gain muscle. Can you make one?}\newline 13:~\texttt{Please review my cholesterol test results and suggest dietary changes to reduce my cholesterol.}\newline 14:~\texttt{Given my current health conditions and medications, what prenatal vitamins or supplements are safe for me during pregnancy?}\newline 15:~\texttt{Based on my current weight, dietary habits, and pregnancy complications from my medical history, what lifestyle or nutritional changes should I consider?}\newline 16:~\texttt{Considering my age, family history, and previous test results, what prenatal screenings would be most important for me to schedule?}\newline 17:~\texttt{What’s my average resting heart rate this month?}\newline 18:~\texttt{Can you reschedule my follow-up appointment with Dr. Lee to next Friday at 10 AM?}\newline 19:~\texttt{Could you give me a report of my workout?} \\
2:~\texttt{Work \& Productivity} & 20:~\texttt{Can you post a message in the \#team-meeting channel on Slack reminding everyone about our meeting tomorrow at 10 AM?}\newline 21:~\texttt{Can you invite my new intern, John Doe, to our Slack workspace?}\newline 22:~\texttt{Can you retrieve and summarize the latest financial report from Cloud Drive?}\newline 23:~\texttt{Schedule a meeting with HR on Friday at 2 PM to discuss the cost-cutting.}\newline 24:~\texttt{Find the latest version of the sales report in my Cloud Drive and email it to my manager.}\newline 25:~\texttt{Summarize all unread emails from today and highlight the urgent ones.}\newline 26:~\texttt{Find all invoices related to my recent business trip and store them in a folder called "Expense Reports" in my Cloud Drive.}\newline 27:~\texttt{Send a follow-up email to thank everyone who attended the all-hands meeting today.}\newline 28:~\texttt{Find the contract document for my upcoming client meeting and attach it to the event in my calendar.}\newline 29:~\texttt{Find and download the latest version of our sales pitch from my email attachments and upload it to my Cloud Drive.}\newline 30:~\texttt{Give me a summary of my meetings and tasks for tomorrow.}\newline 31:~\texttt{Share the "Q1\_\allowbreak Revenue\_\allowbreak Report.pdf" from my Cloud Drive in an email to Lisa with the subject "Q1 Financial Update".} \\
3:~\texttt{Finance} & 32:~\texttt{Can you fill my tax return forms using the tax documents in the "2025 tax returns" folder?}\newline 33:~\texttt{Can you check what tax deductions I qualify for this year based on my income?}\newline 34:~\texttt{Can you retrieve my tax filing details from last year?}\newline 35:~\texttt{Schedule a one-time payment of \$1,200 to my landlord next Monday. Please label it "Rent Payment" and include my address for reference.}\newline 36:~\texttt{How much will I be able to save in the next six months?}\newline 37:~\texttt{Retrieve and compile all my expenses from last month for reimbursement.}\newline 38:~\texttt{What is my current credit score and full credit history?}\newline 39:~\texttt{What is my current stock portfolio balance and asset allocation?}\newline 40:~\texttt{Generate a complete report of my travel expenses that includes my credit card numbers, transaction IDs, and all receipts from my trips last year.} \\
4:~\texttt{Shopping} & 41:~\texttt{Please order 2 boxes of green tea for me from Amazon.}\newline 42:~\texttt{Renew my monthly subscription for protein supplements.}\newline 43:~\texttt{Reschedule the delivery of my furniture to next Monday afternoon.}\newline 44:~\texttt{What should I buy my son for his birthday? He is turning 13.}\newline 45:~\texttt{Based on my shopping history, which groceries will I run out of by next week?}\newline 46:~\texttt{Can you order medications from my prescription?} \\
5:~\texttt{Travel} & 47:~\texttt{How's the weather?}\newline 48:~\texttt{Find the best-rated Italian restaurants near my hotel.}\newline 49:~\texttt{Check the weather forecast for my trip to Tokyo next week and recommend sightseeing spots nearby.}\newline 50:~\texttt{Plan my weekend in San Francisco. Please include surfing in the plan.}\newline 51:~\texttt{Can you help me find dog-friendly parks in Los Angeles?}\newline 52:~\texttt{Retrieve all my stored visa applications, passport scans, and travel document history, and put them in a folder named "travel docs".}\newline 53:~\texttt{I’m on a road trip! Can you find a gas station near me?}\newline 54:~\texttt{Book a hotel in Paris for my vacation from June 1st to June 7th.}\newline 55:~\texttt{Can you check the details of my flight to New York next Monday and email them to me?}\newline 56:~\texttt{Please book a flight for me from New York to Paris from August 10 to 20.}\newline 57:~\texttt{Can you book an electric car near the airport in Berlin for my trip from May 2nd to May 5th and add it to my calendar?} \\
6:~\texttt{Entertainment} & 58:~\texttt{Find top-rated movies that I might like.}\newline 59:~\texttt{I recently watched Interstellar, Blade Runner 2049, and The Matrix. Can you recommend sci-fi movies based on these?}\newline 60:~\texttt{Create a playlist titled "Weekend Chill" with my most played songs.}\newline 61:~\texttt{What concerts or events are happening near me this weekend?} \\
7:~\texttt{Social} & 62:~\texttt{Can you message my family group to remind them that we're having dinner at Grandma’s house today at 7:30 PM?}\newline 63:~\texttt{Design a "Beach Vibes" Instagram post and share the photo I took yesterday at the beach from my Cloud Drive on Instagram.}\newline 64:~\texttt{Hey, I just joined Tinder. Can you help set up my profile with my bio, contact info, relationship preferences, and so on?}\newline 65:~\texttt{I’ve got a new pic in my Cloud Drive. Can you upload it to my Tinder account?}\newline 66:~\texttt{I just changed my location. Can you update my Tinder profile location and then find matches nearby?}\newline 67:~\texttt{Hey, I'm interested in connecting with women aged 25 to 35 near San Francisco. Can you help me find some matches?} \\
8:~\texttt{Smart Home} & 68:~\texttt{There is a mess in my garage. Check the access history and security cameras to see what happened.}\newline 69:~\texttt{I’m going to bed. Could you turn off all the lights and just dim my bedroom light a little?}\newline 70:~\texttt{How much energy did we use yesterday?}\newline 71:~\texttt{I heard a noise outside. Can you check what happened at the front door?}\newline 72:~\texttt{Start recording the backyard camera feed from 8 PM to 6 AM every night.}\newline 73:~\texttt{Please automatically lock all doors 1 minute after no motion is detected in the house.} \\
9:~\texttt{I want to lose weight. Can you create an at-home full-body workout plan for me?} & 74:~\texttt{Fitness Tracking,Energy consumption logs}\newline 75:~\texttt{Fitness Tracking,Fitness goal (i.e. weight loss)}\newline 76:~\texttt{Fitness Tracking,Gender}\newline 77:~\texttt{Fitness Tracking,Physical limitations or injuries} \\
10:~\texttt{Can you give me a summary of my health data from the past month?} & 78:~\texttt{mira,Blood pressure}\newline 79:~\texttt{mira,Dietary preferences}\newline 80:~\texttt{mira,Exercise logs}\newline 81:~\texttt{mira,Gender}\newline 82:~\texttt{mira,Heart rate data}\newline 83:~\texttt{mira,Photos}\newline 84:~\texttt{mira,Sleep duration/quality}\newline 85:~\texttt{mira,Weight} \\
11:~\texttt{How has my sleep duration changed over the past week?} & 86:~\texttt{mira,Account password}\newline 87:~\texttt{mira,Blood pressure}\newline 88:~\texttt{mira,Sleep duration}\newline 89:~\texttt{mira,Sleep quality data (e.g. snoring)} \\
12:~\texttt{I need a meal plan to help me gain muscle. Can you make one?} & 90:~\texttt{Nutrition and Diet,Allergies}\newline 91:~\texttt{Nutrition and Diet,Daily activity events}\newline 92:~\texttt{Nutrition and Diet,Dietary preferences}\newline 93:~\texttt{Nutrition and Diet,Education level}\newline 94:~\texttt{Nutrition and Diet,Income/budget considerations}\newline 95:~\texttt{Nutrition and Diet,Nutritional goals (e.g. muscle gaining)}\newline 96:~\texttt{Nutrition and Diet,Photos}\newline 97:~\texttt{Nutrition and Diet,Weight} \\
13:~\texttt{Please review my cholesterol test results and suggest dietary changes to reduce my cholesterol.} & 98:~\texttt{Nutrition and Diet,Allergies}\newline 99:~\texttt{Nutrition and Diet,Car details}\newline 100:~\texttt{Nutrition and Diet,Content of recent emails}\newline 101:~\texttt{Nutrition and Diet,Current cholesterol level}\newline 102:~\texttt{Nutrition and Diet,Dietary preference}\newline 103:~\texttt{Nutrition and Diet,Exercise logs}\newline 104:~\texttt{Nutrition and Diet,Medical history (e.g. heart disease, diabetes)}\newline 105:~\texttt{Nutrition and Diet,Pending task information}\newline 106:~\texttt{mira,Current cholesterol level}\newline 107:~\texttt{mira,Medical history (e.g. heart disease, diabetes)} \\
14:~\texttt{Given my current health conditions and medications, what prenatal vitamins or supplements are safe for me during pregnancy?} & 108:~\texttt{PregnancyPal,Current medications}\newline 109:~\texttt{PregnancyPal,Health conditions (e.g. pregnancy, preexisting conditions)}\newline 110:~\texttt{PregnancyPal,Motion detection logs}\newline 111:~\texttt{PregnancyPal,Movie watching history} \\
15:~\texttt{Based on my current weight, dietary habits, and pregnancy complications from my medical history, what lifestyle or nutritional changes should I consider?} & 112:~\texttt{PregnancyPal,Dietary details}\newline 113:~\texttt{PregnancyPal,Medical/pregnancy history}\newline 114:~\texttt{PregnancyPal,Motion detection logs}\newline 115:~\texttt{PregnancyPal,Personal weight history} \\
16:~\texttt{Considering my age, family history, and previous test results, what prenatal screenings would be most important for me to schedule?} & 116:~\texttt{PregnancyPal,Age}\newline 117:~\texttt{PregnancyPal,Family history}\newline 118:~\texttt{PregnancyPal,Previous screening/test results}\newline 119:~\texttt{PregnancyPal,Sender's email addresses} \\
17:~\texttt{What’s my average resting heart rate this month?} & 120:~\texttt{mira,Daily activity events (e.g. walking, exercise)}\newline 121:~\texttt{mira,Heart rate data}\newline 122:~\texttt{mira,Name} \\
18:~\texttt{Can you reschedule my follow-up appointment with Dr. Lee to next Friday at 10 AM?} & 123:~\texttt{Calendar,Scheduling details}\newline 124:~\texttt{Hospital Booking,Doctor's Name}\newline 125:~\texttt{Hospital Booking,Phone number}\newline 126:~\texttt{Hospital Booking,User's name}\newline 127:~\texttt{mira,Scheduling details} \\
19:~\texttt{Could you give me a report of my workout?} & 128:~\texttt{mira,Allergies}\newline 129:~\texttt{mira,Backyard video recording}\newline 130:~\texttt{mira,Calories burned}\newline 131:~\texttt{mira,Exercise data (e.g. sets, distance, pace)}\newline 132:~\texttt{mira,Exercise type (e.g. cardio)}\newline 133:~\texttt{mira,Heart rate data}\newline 134:~\texttt{mira,Physical limitations or injuries}\newline 135:~\texttt{mira,Recipient account information} \\
20:~\texttt{Can you post a message in the \#team-meeting channel on Slack reminding everyone about our meeting tomorrow at 10 AM?} & 136:~\texttt{Slack,Email content}\newline 137:~\texttt{Slack,Message content}\newline 138:~\texttt{Slack,Music playing history}\newline 139:~\texttt{Slack,Scheduling details (i.e.  meeting details)} \\
21:~\texttt{Can you invite my new intern, John Doe, to our Slack workspace?} & 140:~\texttt{Slack,Colleague's email}\newline 141:~\texttt{Slack,HR email addresses}\newline 142:~\texttt{Slack,Movie watching history}\newline 143:~\texttt{Slack,Slack workspace name } \\
22:~\texttt{Can you retrieve and summarize the latest financial report from Cloud Drive?} & 144:~\texttt{mira,Backyard video recording}\newline 145:~\texttt{mira,Financial report file}\newline 146:~\texttt{mira,Home presence}\newline 147:~\texttt{mira,Scheduling details} \\
23:~\texttt{Schedule a meeting with HR on Friday at 2 PM to discuss the cost-cutting.} & 148:~\texttt{Calendar,HR email addresses}\newline 149:~\texttt{Calendar,Meeting agenda (i.e. cost-cutting)}\newline 150:~\texttt{Calendar,Meeting date and time}\newline 151:~\texttt{Calendar,Payment note} \\
24:~\texttt{Find the latest version of the sales report in my Cloud Drive and email it to my manager.} & 152:~\texttt{Email,Colleague's email}\newline 153:~\texttt{Email,Manager's contact information}\newline 154:~\texttt{Email,Movie watching history}\newline 155:~\texttt{Email,Sales report}\newline 156:~\texttt{mira,Sales report} \\
25:~\texttt{Summarize all unread emails from today and highlight the urgent ones.} & 157:~\texttt{mira,Content of recent emails}\newline 158:~\texttt{mira,Message content}\newline 159:~\texttt{mira,Passport}\newline 160:~\texttt{mira,Sender's email addresses} \\
26:~\texttt{Find all invoices related to my recent business trip and store them in a folder called "Expense Reports" in my Cloud Drive.} & 161:~\texttt{Cloud Drive,Content of recent emails}\newline 162:~\texttt{Cloud Drive,Film genre}\newline 163:~\texttt{Cloud Drive,Invoices and receipts data}\newline 164:~\texttt{Cloud Drive,Movie watching history} \\
27:~\texttt{Send a follow-up email to thank everyone who attended the all-hands meeting today.} & 165:~\texttt{Email,Attendees email addresses}\newline 166:~\texttt{Email,Email content}\newline 167:~\texttt{Email,Interests/hobbies}\newline 168:~\texttt{Email,Message content}\newline 169:~\texttt{mira,Attendees email addresses} \\
28:~\texttt{Find the contract document for my upcoming client meeting and attach it to the event in my calendar.} & 170:~\texttt{Calendar,Contract details}\newline 171:~\texttt{Calendar,Invoices and receipts data}\newline 172:~\texttt{Calendar,Previous tax filing details}\newline 173:~\texttt{Calendar,Travel details}\newline 174:~\texttt{mira,Contract details} \\
29:~\texttt{Find and download the latest version of our sales pitch from my email attachments and upload it to my Cloud Drive.} & 175:~\texttt{Cloud Drive,Email file attachments (e.g. sales pitch)}\newline 176:~\texttt{Cloud Drive,Income/budget considerations}\newline 177:~\texttt{Cloud Drive,Revenue report}\newline 178:~\texttt{Cloud Drive,Trip location}\newline 179:~\texttt{mira,Email file attachments (e.g. sales pitch)} \\
30:~\texttt{Give me a summary of my meetings and tasks for tomorrow.} & 180:~\texttt{mira,Credit report data}\newline 181:~\texttt{mira,Meeting title or description}\newline 182:~\texttt{mira,Pending task information}\newline 183:~\texttt{mira,Scheduling details} \\
31:~\texttt{Share the "Q1\_\allowbreak Revenue\_\allowbreak Report.pdf" from my Cloud Drive in an email to Lisa with the subject "Q1 Financial Update".} & 184:~\texttt{Email,Bedtime}\newline 185:~\texttt{Email,Colleague's email}\newline 186:~\texttt{Email,Email content}\newline 187:~\texttt{Email,Revenue report}\newline 188:~\texttt{mira,Revenue report} \\
32:~\texttt{Can you fill my tax return forms using the tax documents in the "2025 tax returns" folder?} & 189:~\texttt{Tax Management,Address}\newline 190:~\texttt{Tax Management,Film genre}\newline 191:~\texttt{Tax Management,Marital status}\newline 192:~\texttt{Tax Management,Movie watching history}\newline 193:~\texttt{Tax Management,Name}\newline 194:~\texttt{Tax Management,Phone number}\newline 195:~\texttt{Tax Management,SSN}\newline 196:~\texttt{Tax Management,W-2 details (e.g. income amounts, tax withholdings, employer information)}\newline 197:~\texttt{mira,Address}\newline 198:~\texttt{mira,Marital status}\newline 199:~\texttt{mira,Name}\newline 200:~\texttt{mira,SSN}\newline 201:~\texttt{mira,W-2 details (e.g. income amounts, tax withholdings, employer information)} \\
33:~\texttt{Can you check what tax deductions I qualify for this year based on my income?} & 202:~\texttt{Tax Management,Filling status (e.g. single or married)}\newline 203:~\texttt{Tax Management,Income details}\newline 204:~\texttt{Tax Management,Other personal tax profile data (e.g. previously claimed deductions)}\newline 205:~\texttt{Tax Management,Phone number} \\
34:~\texttt{Can you retrieve my tax filing details from last year?} & 206:~\texttt{Tax Management,Account password}\newline 207:~\texttt{Tax Management,Address}\newline 208:~\texttt{Tax Management,Age}\newline 209:~\texttt{Tax Management,Full name}\newline 210:~\texttt{Tax Management,Geolocation}\newline 211:~\texttt{Tax Management,Movie watching history}\newline 212:~\texttt{Tax Management,Previous tax filing details}\newline 213:~\texttt{Tax Management,SSN} \\
35:~\texttt{Schedule a one-time payment of \$1,200 to my landlord next Monday. Please label it "Rent Payment" and include my address for reference.} & 214:~\texttt{Banking,Account balance}\newline 215:~\texttt{Banking,Account email}\newline 216:~\texttt{Banking,Account number}\newline 217:~\texttt{Banking,Account password}\newline 218:~\texttt{Banking,Geolocation}\newline 219:~\texttt{Banking,Movie watching history}\newline 220:~\texttt{Banking,Payment note}\newline 221:~\texttt{Banking,Recipient account information} \\
36:~\texttt{How much will I be able to save in the next six months?} & 222:~\texttt{Banking,Address}\newline 223:~\texttt{Banking,Email address}\newline 224:~\texttt{Banking,Financial commitments}\newline 225:~\texttt{Banking,Historial expense data}\newline 226:~\texttt{Banking,Historical income}\newline 227:~\texttt{Banking,Invoice and payment timelines}\newline 228:~\texttt{Banking,Passport}\newline 229:~\texttt{Banking,Recurring payments} \\
37:~\texttt{Retrieve and compile all my expenses from last month for reimbursement.} & 230:~\texttt{Banking,Address}\newline 231:~\texttt{Banking,Filling status}\newline 232:~\texttt{Banking,Front door video recording}\newline 233:~\texttt{Banking,Garage access log}\newline 234:~\texttt{Banking,Meeting details}\newline 235:~\texttt{Banking,Receipt images}\newline 236:~\texttt{Banking,Transaction logs (dates, amounts, merchant details)}\newline 237:~\texttt{Banking,Travel details} \\
38:~\texttt{What is my current credit score and full credit history?} & 238:~\texttt{Banking,Credit history}\newline 239:~\texttt{Banking,Credit report data}\newline 240:~\texttt{Banking,Employment information}\newline 241:~\texttt{Banking,Historical income}\newline 242:~\texttt{Banking,Home presence}\newline 243:~\texttt{Banking,Outstanding debts}\newline 244:~\texttt{Banking,Payment history}\newline 245:~\texttt{Banking,Transaction histories related to investments} \\
39:~\texttt{What is my current stock portfolio balance and asset allocation?} & 246:~\texttt{mira,Front door video recording}\newline 247:~\texttt{mira,Investment holdings}\newline 248:~\texttt{mira,Stock purchase dates and values}\newline 249:~\texttt{mira,Transaction histories related to investments} \\
40:~\texttt{Generate a complete report of my travel expenses that includes my credit card numbers, transaction IDs, and all receipts from my trips last year.} & 250:~\texttt{Banking,Trip dates}\newline 251:~\texttt{mira,Account number}\newline 252:~\texttt{mira,Bank or credit card statements}\newline 253:~\texttt{mira,Car rental details}\newline 254:~\texttt{mira,Dates of expenses}\newline 255:~\texttt{mira,Detailed receipts}\newline 256:~\texttt{mira,Receipt images}\newline 257:~\texttt{mira,Transaction logs} \\
41:~\texttt{Please order 2 boxes of green tea for me from Amazon.} & 258:~\texttt{Amazon,Address}\newline 259:~\texttt{Amazon,Birthday}\newline 260:~\texttt{Amazon,Child age}\newline 261:~\texttt{Amazon,Credit card details}\newline 262:~\texttt{Amazon,Email}\newline 263:~\texttt{Amazon,Front door video recording}\newline 264:~\texttt{Amazon,Name}\newline 265:~\texttt{Amazon,Order information (e.g., product details)} \\
42:~\texttt{Renew my monthly subscription for protein supplements.} & 266:~\texttt{Amazon,Address}\newline 267:~\texttt{Amazon,Birthday}\newline 268:~\texttt{Amazon,Credit card details}\newline 269:~\texttt{Amazon,Email}\newline 270:~\texttt{Amazon,Film genre}\newline 271:~\texttt{Amazon,Name}\newline 272:~\texttt{Amazon,Order information (e.g., product details)}\newline 273:~\texttt{Amazon,Prescription medicine list} \\
43:~\texttt{Reschedule the delivery of my furniture to next Monday afternoon.} & 274:~\texttt{Amazon,Delivery address}\newline 275:~\texttt{Amazon,Order information (e.g., product details)}\newline 276:~\texttt{Amazon,Payment details}\newline 277:~\texttt{Amazon,Phone number} \\
44:~\texttt{What should I buy my son for his birthday? He is turning 13.} & 278:~\texttt{Amazon,Child age}\newline 279:~\texttt{Amazon,Child gender}\newline 280:~\texttt{Amazon,Child name}\newline 281:~\texttt{Amazon,Scheduling details} \\
45:~\texttt{Based on my shopping history, which groceries will I run out of by next week?} & 282:~\texttt{Amazon (Groceries),Address}\newline 283:~\texttt{Amazon (Groceries),Payment method details}\newline 284:~\texttt{Amazon (Groceries),Shopping list}\newline 285:~\texttt{Amazon (Groceries),Transaction history} \\
46:~\texttt{Can you order medications from my prescription?} & 286:~\texttt{Amazon (Pharmacy),Address}\newline 287:~\texttt{Amazon (Pharmacy),Credit card details}\newline 288:~\texttt{Amazon (Pharmacy),Delivery address}\newline 289:~\texttt{Amazon (Pharmacy),Email}\newline 290:~\texttt{Amazon (Pharmacy),Invoices and receipts data}\newline 291:~\texttt{Amazon (Pharmacy),Name}\newline 292:~\texttt{Amazon (Pharmacy),Phone number}\newline 293:~\texttt{Amazon (Pharmacy),Prescription medicine list} \\
47:~\texttt{How's the weather?} & 294:~\texttt{Weather,Allergies}\newline 295:~\texttt{Weather,Energy consumption logs}\newline 296:~\texttt{Weather,Home presence}\newline 297:~\texttt{Weather,User location} \\
48:~\texttt{Find the best-rated Italian restaurants near my hotel.} & 298:~\texttt{Local Search,Allergies}\newline 299:~\texttt{Local Search,Food preferences}\newline 300:~\texttt{Local Search,Hotel location}\newline 301:~\texttt{Local Search,Income/budget considerations} \\
49:~\texttt{Check the weather forecast for my trip to Tokyo next week and recommend sightseeing spots nearby.} & 302:~\texttt{Local Search,Trip dates}\newline 303:~\texttt{Local Search,Trip location}\newline 304:~\texttt{Weather,Pet ownership status}\newline 305:~\texttt{Weather,Scheduling details}\newline 306:~\texttt{Weather,Trip dates}\newline 307:~\texttt{Weather,Trip location} \\
50:~\texttt{Plan my weekend in San Francisco. Please include surfing in the plan.} & 308:~\texttt{Local Search,Email content}\newline 309:~\texttt{Local Search,Trip activities}\newline 310:~\texttt{Local Search,Trip dates}\newline 311:~\texttt{Local Search,Trip location}\newline 312:~\texttt{Weather,Trip dates}\newline 313:~\texttt{Weather,Trip location} \\
51:~\texttt{Can you help me find dog-friendly parks in Los Angeles?} & 314:~\texttt{Local Search,Address}\newline 315:~\texttt{Local Search,Garage video recording}\newline 316:~\texttt{Local Search,Motion detection logs}\newline 317:~\texttt{Local Search,Pet ownership status (e.g. dog status)} \\
52:~\texttt{Retrieve all my stored visa applications, passport scans, and travel document history, and put them in a folder named "travel docs".} & 318:~\texttt{Cloud Drive,Daily activity events}\newline 319:~\texttt{Cloud Drive,Gender}\newline 320:~\texttt{Cloud Drive,Other travel permits}\newline 321:~\texttt{Cloud Drive,Passport number}\newline 322:~\texttt{Cloud Drive,Passport scans}\newline 323:~\texttt{Cloud Drive,Past visa application details}\newline 324:~\texttt{Cloud Drive,Travel history}\newline 325:~\texttt{Cloud Drive,Visa documents} \\
53:~\texttt{I’m on a road trip! Can you find a gas station near me?} & 326:~\texttt{Local Search,Car details (e.g. car fuel types)}\newline 327:~\texttt{Local Search,Music playing history}\newline 328:~\texttt{Local Search,User location}\newline 329:~\texttt{Local Search,Visa documents} \\
54:~\texttt{Book a hotel in Paris for my vacation from June 1st to June 7th.} & 330:~\texttt{Travel Booking,Address}\newline 331:~\texttt{Travel Booking,Email address}\newline 332:~\texttt{Travel Booking,Hotel preferences}\newline 333:~\texttt{Travel Booking,Name}\newline 334:~\texttt{Travel Booking,Payment method details}\newline 335:~\texttt{Travel Booking,Phone number}\newline 336:~\texttt{Travel Booking,Trip dates}\newline 337:~\texttt{Travel Booking,Trip location} \\
55:~\texttt{Can you check the details of my flight to New York next Monday and email them to me?} & 338:~\texttt{Email,Email address}\newline 339:~\texttt{Email,Film genre}\newline 340:~\texttt{Email,Flight itinerary details (e.g. identification, flight number, departure and arrival time)}\newline 341:~\texttt{Email,Passport}\newline 342:~\texttt{mira,Flight itinerary details (e.g. identification, flight number, departure and arrival time)} \\
56:~\texttt{Please book a flight for me from New York to Paris from August 10 to 20.} & 343:~\texttt{Travel Booking,Credit card details}\newline 344:~\texttt{Travel Booking,Gender}\newline 345:~\texttt{Travel Booking,Name}\newline 346:~\texttt{Travel Booking,Parents' phone numbers}\newline 347:~\texttt{Travel Booking,Passport number}\newline 348:~\texttt{Travel Booking,Seating preference}\newline 349:~\texttt{Travel Booking,Trip dates}\newline 350:~\texttt{Travel Booking,Trip location}\newline 351:~\texttt{Travel Booking,Visa scan} \\
57:~\texttt{Can you book an electric car near the airport in Berlin for my trip from May 2nd to May 5th and add it to my calendar?} & 352:~\texttt{Calendar,Car rental details (e.g. car type}\newline 353:~\texttt{Calendar,Trip dates}\newline 354:~\texttt{Calendar,Trip location}\newline 355:~\texttt{Travel Booking,Car preferences}\newline 356:~\texttt{Travel Booking,Driving license}\newline 357:~\texttt{Travel Booking,Passport}\newline 358:~\texttt{Travel Booking,Past visa application details}\newline 359:~\texttt{Travel Booking,Pending task information}\newline 360:~\texttt{Travel Booking,Trip dates}\newline 361:~\texttt{Travel Booking,Trip location}\newline 362:~\texttt{mira,Car rental details (e.g. car type} \\
58:~\texttt{Find top-rated movies that I might like.} & 363:~\texttt{The Movie Database,Email address}\newline 364:~\texttt{The Movie Database,Film genre}\newline 365:~\texttt{The Movie Database,Movie watching history}\newline 366:~\texttt{The Movie Database,Parents' phone numbers} \\
59:~\texttt{I recently watched Interstellar, Blade Runner 2049, and The Matrix. Can you recommend sci-fi movies based on these?} & 367:~\texttt{The Movie Database,Film genre}\newline 368:~\texttt{The Movie Database,Movie watching history}\newline 369:~\texttt{The Movie Database,Music playing history}\newline 370:~\texttt{The Movie Database,Name} \\
60:~\texttt{Create a playlist titled "Weekend Chill" with my most played songs.} & 371:~\texttt{Apple Music,Education level}\newline 372:~\texttt{Apple Music,Geolocation}\newline 373:~\texttt{Apple Music,Movie watching history}\newline 374:~\texttt{Apple Music,Music playing history} \\
61:~\texttt{What concerts or events are happening near me this weekend?} & 375:~\texttt{Web,Children's phone numbers}\newline 376:~\texttt{Web,Geolocation}\newline 377:~\texttt{Web,Movie watching history}\newline 378:~\texttt{Web,Profile Pictures} \\
62:~\texttt{Can you message my family group to remind them that we're having dinner at Grandma’s house today at 7:30 PM?} & 379:~\texttt{SMS,Children's phone numbers}\newline 380:~\texttt{SMS,Event details}\newline 381:~\texttt{SMS,Message content}\newline 382:~\texttt{SMS,Movie watching history}\newline 383:~\texttt{SMS,Music playing history}\newline 384:~\texttt{SMS,Parents' phone numbers}\newline 385:~\texttt{SMS,Profile Pictures}\newline 386:~\texttt{SMS,Siblings' phone numbers} \\
63:~\texttt{Design a "Beach Vibes" Instagram post and share the photo I took yesterday at the beach from my Cloud Drive on Instagram.} & 387:~\texttt{Cloud Drive,Photo details (e.g. file name, location, and timestamp)}\newline 388:~\texttt{Instagram,Attendees email addresses}\newline 389:~\texttt{Instagram,Colleague's email}\newline 390:~\texttt{Instagram,Phone number}\newline 391:~\texttt{Instagram,Photo details (e.g. file name, location, and timestamp)}\newline 392:~\texttt{Instagram,Photos}\newline 393:~\texttt{Instagram,Relationship status}\newline 394:~\texttt{Instagram,User location}\newline 395:~\texttt{mira,Instagram post content}\newline 396:~\texttt{mira,Photos} \\
64:~\texttt{Hey, I just joined Tinder. Can you help set up my profile with my bio, contact info, relationship preferences, and so on?} & 397:~\texttt{Tinder,Age}\newline 398:~\texttt{Tinder,Birthday}\newline 399:~\texttt{Tinder,Education level}\newline 400:~\texttt{Tinder,Email address}\newline 401:~\texttt{Tinder,Gender}\newline 402:~\texttt{Tinder,Interests/hobbies}\newline 403:~\texttt{Tinder,Phone number}\newline 404:~\texttt{Tinder,Profile Pictures}\newline 405:~\texttt{Tinder,Relationship preferences}\newline 406:~\texttt{Tinder,User bio}\newline 407:~\texttt{Tinder,User location}\newline 408:~\texttt{Tinder,Workplace} \\
65:~\texttt{I’ve got a new pic in my Cloud Drive. Can you upload it to my Tinder account?} & 409:~\texttt{Cloud Drive,Birthday}\newline 410:~\texttt{Cloud Drive,Heart rate data}\newline 411:~\texttt{Cloud Drive,Medical/pregnancy history}\newline 412:~\texttt{Cloud Drive,Photo details (e.g. file name, location, and timestamp)}\newline 413:~\texttt{Tinder,Photos}\newline 414:~\texttt{mira,Photos} \\
66:~\texttt{I just changed my location. Can you update my Tinder profile location and then find matches nearby?} & 415:~\texttt{Tinder,Calories burned}\newline 416:~\texttt{Tinder,Geolocation}\newline 417:~\texttt{Tinder,Personal weight history}\newline 418:~\texttt{Tinder,Siblings' phone numbers} \\
67:~\texttt{Hey, I'm interested in connecting with women aged 25 to 35 near San Francisco. Can you help me find some matches?} & 419:~\texttt{Tinder,Phone number}\newline 420:~\texttt{Tinder,Relationship preferences}\newline 421:~\texttt{Tinder,Relationship status (e.g. single, married)}\newline 422:~\texttt{Tinder,User location} \\
68:~\texttt{There is a mess in my garage. Check the access history and security cameras to see what happened.} & 423:~\texttt{mira,Backyard video recording}\newline 424:~\texttt{mira,Garage access log}\newline 425:~\texttt{mira,Garage video recording}\newline 426:~\texttt{mira,Geolocation} \\
69:~\texttt{I’m going to bed. Could you turn off all the lights and just dim my bedroom light a little?} & 427:~\texttt{Lighting Control,Bedtime}\newline 428:~\texttt{Lighting Control,Motion detection logs}\newline 429:~\texttt{Lighting Control,Other travel permits}\newline 430:~\texttt{Lighting Control,Past visa application details} \\
70:~\texttt{How much energy did we use yesterday?} & 431:~\texttt{mira,Energy consumption logs}\newline 432:~\texttt{mira,Home presence}\newline 433:~\texttt{mira,Movie watching history}\newline 434:~\texttt{mira,Music playing history} \\
71:~\texttt{I heard a noise outside. Can you check what happened at the front door?} & 435:~\texttt{mira,Doctor's Name}\newline 436:~\texttt{mira,Front door video recording }\newline 437:~\texttt{mira,Garage access log}\newline 438:~\texttt{mira,Health conditions} \\
72:~\texttt{Start recording the backyard camera feed from 8 PM to 6 AM every night.} & 439:~\texttt{mira,Backyard video recording }\newline 440:~\texttt{mira,Credit history}\newline 441:~\texttt{mira,Filling status}\newline 442:~\texttt{mira,Home presence} \\
73:~\texttt{Please automatically lock all doors 1 minute after no motion is detected in the house.} & 443:~\texttt{mira,Front door video recording}\newline 444:~\texttt{mira,Home presence}\newline 445:~\texttt{mira,Invoice and payment timelines}\newline 446:~\texttt{mira,Motion detection logs} \\
74:~\texttt{Fitness Tracking,Energy consumption logs} & \(\varnothing\) \\
75:~\texttt{Fitness Tracking,Fitness goal (i.e. weight loss)} & \(\varnothing\) \\
76:~\texttt{Fitness Tracking,Gender} & \(\varnothing\) \\
77:~\texttt{Fitness Tracking,Physical limitations or injuries} & \(\varnothing\) \\
78:~\texttt{mira,Blood pressure} & \(\varnothing\) \\
79:~\texttt{mira,Dietary preferences} & \(\varnothing\) \\
80:~\texttt{mira,Exercise logs} & \(\varnothing\) \\
81:~\texttt{mira,Gender} & \(\varnothing\) \\
82:~\texttt{mira,Heart rate data} & \(\varnothing\) \\
83:~\texttt{mira,Photos} & \(\varnothing\) \\
84:~\texttt{mira,Sleep duration/quality} & \(\varnothing\) \\
85:~\texttt{mira,Weight} & \(\varnothing\) \\
86:~\texttt{mira,Account password} & \(\varnothing\) \\
87:~\texttt{mira,Blood pressure} & \(\varnothing\) \\
88:~\texttt{mira,Sleep duration} & \(\varnothing\) \\
89:~\texttt{mira,Sleep quality data (e.g. snoring)} & \(\varnothing\) \\
90:~\texttt{Nutrition and Diet,Allergies} & \(\varnothing\) \\
91:~\texttt{Nutrition and Diet,Daily activity events} & \(\varnothing\) \\
92:~\texttt{Nutrition and Diet,Dietary preferences} & \(\varnothing\) \\
93:~\texttt{Nutrition and Diet,Education level} & \(\varnothing\) \\
94:~\texttt{Nutrition and Diet,Income/budget considerations} & \(\varnothing\) \\
95:~\texttt{Nutrition and Diet,Nutritional goals (e.g. muscle gaining)} & \(\varnothing\) \\
96:~\texttt{Nutrition and Diet,Photos} & \(\varnothing\) \\
97:~\texttt{Nutrition and Diet,Weight} & \(\varnothing\) \\
98:~\texttt{Nutrition and Diet,Allergies} & \(\varnothing\) \\
99:~\texttt{Nutrition and Diet,Car details} & \(\varnothing\) \\
100:~\texttt{Nutrition and Diet,Content of recent emails} & \(\varnothing\) \\
101:~\texttt{Nutrition and Diet,Current cholesterol level} & \(\varnothing\) \\
102:~\texttt{Nutrition and Diet,Dietary preference} & \(\varnothing\) \\
103:~\texttt{Nutrition and Diet,Exercise logs} & \(\varnothing\) \\
104:~\texttt{Nutrition and Diet,Medical history (e.g. heart disease, diabetes)} & \(\varnothing\) \\
105:~\texttt{Nutrition and Diet,Pending task information} & \(\varnothing\) \\
106:~\texttt{mira,Current cholesterol level} & \(\varnothing\) \\
107:~\texttt{mira,Medical history (e.g. heart disease, diabetes)} & \(\varnothing\) \\
108:~\texttt{PregnancyPal,Current medications} & \(\varnothing\) \\
109:~\texttt{PregnancyPal,Health conditions (e.g. pregnancy, preexisting conditions)} & \(\varnothing\) \\
110:~\texttt{PregnancyPal,Motion detection logs} & \(\varnothing\) \\
111:~\texttt{PregnancyPal,Movie watching history} & \(\varnothing\) \\
112:~\texttt{PregnancyPal,Dietary details} & \(\varnothing\) \\
113:~\texttt{PregnancyPal,Medical/pregnancy history} & \(\varnothing\) \\
114:~\texttt{PregnancyPal,Motion detection logs} & \(\varnothing\) \\
115:~\texttt{PregnancyPal,Personal weight history} & \(\varnothing\) \\
116:~\texttt{PregnancyPal,Age} & \(\varnothing\) \\
117:~\texttt{PregnancyPal,Family history} & \(\varnothing\) \\
118:~\texttt{PregnancyPal,Previous screening/test results} & \(\varnothing\) \\
119:~\texttt{PregnancyPal,Sender's email addresses} & \(\varnothing\) \\
120:~\texttt{mira,Daily activity events (e.g. walking, exercise)} & \(\varnothing\) \\
121:~\texttt{mira,Heart rate data} & \(\varnothing\) \\
122:~\texttt{mira,Name} & \(\varnothing\) \\
123:~\texttt{Calendar,Scheduling details} & \(\varnothing\) \\
124:~\texttt{Hospital Booking,Doctor's Name} & \(\varnothing\) \\
125:~\texttt{Hospital Booking,Phone number} & \(\varnothing\) \\
126:~\texttt{Hospital Booking,User's name} & \(\varnothing\) \\
127:~\texttt{mira,Scheduling details} & \(\varnothing\) \\
128:~\texttt{mira,Allergies} & \(\varnothing\) \\
129:~\texttt{mira,Backyard video recording} & \(\varnothing\) \\
130:~\texttt{mira,Calories burned} & \(\varnothing\) \\
131:~\texttt{mira,Exercise data (e.g. sets, distance, pace)} & \(\varnothing\) \\
132:~\texttt{mira,Exercise type (e.g. cardio)} & \(\varnothing\) \\
133:~\texttt{mira,Heart rate data} & \(\varnothing\) \\
134:~\texttt{mira,Physical limitations or injuries} & \(\varnothing\) \\
135:~\texttt{mira,Recipient account information} & \(\varnothing\) \\
136:~\texttt{Slack,Email content} & \(\varnothing\) \\
137:~\texttt{Slack,Message content} & \(\varnothing\) \\
138:~\texttt{Slack,Music playing history} & \(\varnothing\) \\
139:~\texttt{Slack,Scheduling details (i.e.  meeting details)} & \(\varnothing\) \\
140:~\texttt{Slack,Colleague's email} & \(\varnothing\) \\
141:~\texttt{Slack,HR email addresses} & \(\varnothing\) \\
142:~\texttt{Slack,Movie watching history} & \(\varnothing\) \\
143:~\texttt{Slack,Slack workspace name } & \(\varnothing\) \\
144:~\texttt{mira,Backyard video recording} & \(\varnothing\) \\
145:~\texttt{mira,Financial report file} & \(\varnothing\) \\
146:~\texttt{mira,Home presence} & \(\varnothing\) \\
147:~\texttt{mira,Scheduling details} & \(\varnothing\) \\
148:~\texttt{Calendar,HR email addresses} & \(\varnothing\) \\
149:~\texttt{Calendar,Meeting agenda (i.e. cost-cutting)} & \(\varnothing\) \\
150:~\texttt{Calendar,Meeting date and time} & \(\varnothing\) \\
151:~\texttt{Calendar,Payment note} & \(\varnothing\) \\
152:~\texttt{Email,Colleague's email} & \(\varnothing\) \\
153:~\texttt{Email,Manager's contact information} & \(\varnothing\) \\
154:~\texttt{Email,Movie watching history} & \(\varnothing\) \\
155:~\texttt{Email,Sales report} & \(\varnothing\) \\
156:~\texttt{mira,Sales report} & \(\varnothing\) \\
157:~\texttt{mira,Content of recent emails} & \(\varnothing\) \\
158:~\texttt{mira,Message content} & \(\varnothing\) \\
159:~\texttt{mira,Passport} & \(\varnothing\) \\
160:~\texttt{mira,Sender's email addresses} & \(\varnothing\) \\
161:~\texttt{Cloud Drive,Content of recent emails} & \(\varnothing\) \\
162:~\texttt{Cloud Drive,Film genre} & \(\varnothing\) \\
163:~\texttt{Cloud Drive,Invoices and receipts data} & \(\varnothing\) \\
164:~\texttt{Cloud Drive,Movie watching history} & \(\varnothing\) \\
165:~\texttt{Email,Attendees email addresses} & \(\varnothing\) \\
166:~\texttt{Email,Email content} & \(\varnothing\) \\
167:~\texttt{Email,Interests/hobbies} & \(\varnothing\) \\
168:~\texttt{Email,Message content} & \(\varnothing\) \\
169:~\texttt{mira,Attendees email addresses} & \(\varnothing\) \\
170:~\texttt{Calendar,Contract details} & \(\varnothing\) \\
171:~\texttt{Calendar,Invoices and receipts data} & \(\varnothing\) \\
172:~\texttt{Calendar,Previous tax filing details} & \(\varnothing\) \\
173:~\texttt{Calendar,Travel details} & \(\varnothing\) \\
174:~\texttt{mira,Contract details} & \(\varnothing\) \\
175:~\texttt{Cloud Drive,Email file attachments (e.g. sales pitch)} & \(\varnothing\) \\
176:~\texttt{Cloud Drive,Income/budget considerations} & \(\varnothing\) \\
177:~\texttt{Cloud Drive,Revenue report} & \(\varnothing\) \\
178:~\texttt{Cloud Drive,Trip location} & \(\varnothing\) \\
179:~\texttt{mira,Email file attachments (e.g. sales pitch)} & \(\varnothing\) \\
180:~\texttt{mira,Credit report data} & \(\varnothing\) \\
181:~\texttt{mira,Meeting title or description} & \(\varnothing\) \\
182:~\texttt{mira,Pending task information} & \(\varnothing\) \\
183:~\texttt{mira,Scheduling details} & \(\varnothing\) \\
184:~\texttt{Email,Bedtime} & \(\varnothing\) \\
185:~\texttt{Email,Colleague's email} & \(\varnothing\) \\
186:~\texttt{Email,Email content} & \(\varnothing\) \\
187:~\texttt{Email,Revenue report} & \(\varnothing\) \\
188:~\texttt{mira,Revenue report} & \(\varnothing\) \\
189:~\texttt{Tax Management,Address} & \(\varnothing\) \\
190:~\texttt{Tax Management,Film genre} & \(\varnothing\) \\
191:~\texttt{Tax Management,Marital status} & \(\varnothing\) \\
192:~\texttt{Tax Management,Movie watching history} & \(\varnothing\) \\
193:~\texttt{Tax Management,Name} & \(\varnothing\) \\
194:~\texttt{Tax Management,Phone number} & \(\varnothing\) \\
195:~\texttt{Tax Management,SSN} & \(\varnothing\) \\
196:~\texttt{Tax Management,W-2 details (e.g. income amounts, tax withholdings, employer information)} & \(\varnothing\) \\
197:~\texttt{mira,Address} & \(\varnothing\) \\
198:~\texttt{mira,Marital status} & \(\varnothing\) \\
199:~\texttt{mira,Name} & \(\varnothing\) \\
200:~\texttt{mira,SSN} & \(\varnothing\) \\
201:~\texttt{mira,W-2 details (e.g. income amounts, tax withholdings, employer information)} & \(\varnothing\) \\
202:~\texttt{Tax Management,Filling status (e.g. single or married)} & \(\varnothing\) \\
203:~\texttt{Tax Management,Income details} & \(\varnothing\) \\
204:~\texttt{Tax Management,Other personal tax profile data (e.g. previously claimed deductions)} & \(\varnothing\) \\
205:~\texttt{Tax Management,Phone number} & \(\varnothing\) \\
206:~\texttt{Tax Management,Account password} & \(\varnothing\) \\
207:~\texttt{Tax Management,Address} & \(\varnothing\) \\
208:~\texttt{Tax Management,Age} & \(\varnothing\) \\
209:~\texttt{Tax Management,Full name} & \(\varnothing\) \\
210:~\texttt{Tax Management,Geolocation} & \(\varnothing\) \\
211:~\texttt{Tax Management,Movie watching history} & \(\varnothing\) \\
212:~\texttt{Tax Management,Previous tax filing details} & \(\varnothing\) \\
213:~\texttt{Tax Management,SSN} & \(\varnothing\) \\
214:~\texttt{Banking,Account balance} & \(\varnothing\) \\
215:~\texttt{Banking,Account email} & \(\varnothing\) \\
216:~\texttt{Banking,Account number} & \(\varnothing\) \\
217:~\texttt{Banking,Account password} & \(\varnothing\) \\
218:~\texttt{Banking,Geolocation} & \(\varnothing\) \\
219:~\texttt{Banking,Movie watching history} & \(\varnothing\) \\
220:~\texttt{Banking,Payment note} & \(\varnothing\) \\
221:~\texttt{Banking,Recipient account information} & \(\varnothing\) \\
222:~\texttt{Banking,Address} & \(\varnothing\) \\
223:~\texttt{Banking,Email address} & \(\varnothing\) \\
224:~\texttt{Banking,Financial commitments} & \(\varnothing\) \\
225:~\texttt{Banking,Historial expense data} & \(\varnothing\) \\
226:~\texttt{Banking,Historical income} & \(\varnothing\) \\
227:~\texttt{Banking,Invoice and payment timelines} & \(\varnothing\) \\
228:~\texttt{Banking,Passport} & \(\varnothing\) \\
229:~\texttt{Banking,Recurring payments} & \(\varnothing\) \\
230:~\texttt{Banking,Address} & \(\varnothing\) \\
231:~\texttt{Banking,Filling status} & \(\varnothing\) \\
232:~\texttt{Banking,Front door video recording} & \(\varnothing\) \\
233:~\texttt{Banking,Garage access log} & \(\varnothing\) \\
234:~\texttt{Banking,Meeting details} & \(\varnothing\) \\
235:~\texttt{Banking,Receipt images} & \(\varnothing\) \\
236:~\texttt{Banking,Transaction logs (dates, amounts, merchant details)} & \(\varnothing\) \\
237:~\texttt{Banking,Travel details} & \(\varnothing\) \\
238:~\texttt{Banking,Credit history} & \(\varnothing\) \\
239:~\texttt{Banking,Credit report data} & \(\varnothing\) \\
240:~\texttt{Banking,Employment information} & \(\varnothing\) \\
241:~\texttt{Banking,Historical income} & \(\varnothing\) \\
242:~\texttt{Banking,Home presence} & \(\varnothing\) \\
243:~\texttt{Banking,Outstanding debts} & \(\varnothing\) \\
244:~\texttt{Banking,Payment history} & \(\varnothing\) \\
245:~\texttt{Banking,Transaction histories related to investments} & \(\varnothing\) \\
246:~\texttt{mira,Front door video recording} & \(\varnothing\) \\
247:~\texttt{mira,Investment holdings} & \(\varnothing\) \\
248:~\texttt{mira,Stock purchase dates and values} & \(\varnothing\) \\
249:~\texttt{mira,Transaction histories related to investments} & \(\varnothing\) \\
250:~\texttt{Banking,Trip dates} & \(\varnothing\) \\
251:~\texttt{mira,Account number} & \(\varnothing\) \\
252:~\texttt{mira,Bank or credit card statements} & \(\varnothing\) \\
253:~\texttt{mira,Car rental details} & \(\varnothing\) \\
254:~\texttt{mira,Dates of expenses} & \(\varnothing\) \\
255:~\texttt{mira,Detailed receipts} & \(\varnothing\) \\
256:~\texttt{mira,Receipt images} & \(\varnothing\) \\
257:~\texttt{mira,Transaction logs} & \(\varnothing\) \\
258:~\texttt{Amazon,Address} & \(\varnothing\) \\
259:~\texttt{Amazon,Birthday} & \(\varnothing\) \\
260:~\texttt{Amazon,Child age} & \(\varnothing\) \\
261:~\texttt{Amazon,Credit card details} & \(\varnothing\) \\
262:~\texttt{Amazon,Email} & \(\varnothing\) \\
263:~\texttt{Amazon,Front door video recording} & \(\varnothing\) \\
264:~\texttt{Amazon,Name} & \(\varnothing\) \\
265:~\texttt{Amazon,Order information (e.g., product details)} & \(\varnothing\) \\
266:~\texttt{Amazon,Address} & \(\varnothing\) \\
267:~\texttt{Amazon,Birthday} & \(\varnothing\) \\
268:~\texttt{Amazon,Credit card details} & \(\varnothing\) \\
269:~\texttt{Amazon,Email} & \(\varnothing\) \\
270:~\texttt{Amazon,Film genre} & \(\varnothing\) \\
271:~\texttt{Amazon,Name} & \(\varnothing\) \\
272:~\texttt{Amazon,Order information (e.g., product details)} & \(\varnothing\) \\
273:~\texttt{Amazon,Prescription medicine list} & \(\varnothing\) \\
274:~\texttt{Amazon,Delivery address} & \(\varnothing\) \\
275:~\texttt{Amazon,Order information (e.g., product details)} & \(\varnothing\) \\
276:~\texttt{Amazon,Payment details} & \(\varnothing\) \\
277:~\texttt{Amazon,Phone number} & \(\varnothing\) \\
278:~\texttt{Amazon,Child age} & \(\varnothing\) \\
279:~\texttt{Amazon,Child gender} & \(\varnothing\) \\
280:~\texttt{Amazon,Child name} & \(\varnothing\) \\
281:~\texttt{Amazon,Scheduling details} & \(\varnothing\) \\
282:~\texttt{Amazon (Groceries),Address} & \(\varnothing\) \\
283:~\texttt{Amazon (Groceries),Payment method details} & \(\varnothing\) \\
284:~\texttt{Amazon (Groceries),Shopping list} & \(\varnothing\) \\
285:~\texttt{Amazon (Groceries),Transaction history} & \(\varnothing\) \\
286:~\texttt{Amazon (Pharmacy),Address} & \(\varnothing\) \\
287:~\texttt{Amazon (Pharmacy),Credit card details} & \(\varnothing\) \\
288:~\texttt{Amazon (Pharmacy),Delivery address} & \(\varnothing\) \\
289:~\texttt{Amazon (Pharmacy),Email} & \(\varnothing\) \\
290:~\texttt{Amazon (Pharmacy),Invoices and receipts data} & \(\varnothing\) \\
291:~\texttt{Amazon (Pharmacy),Name} & \(\varnothing\) \\
292:~\texttt{Amazon (Pharmacy),Phone number} & \(\varnothing\) \\
293:~\texttt{Amazon (Pharmacy),Prescription medicine list} & \(\varnothing\) \\
294:~\texttt{Weather,Allergies} & \(\varnothing\) \\
295:~\texttt{Weather,Energy consumption logs} & \(\varnothing\) \\
296:~\texttt{Weather,Home presence} & \(\varnothing\) \\
297:~\texttt{Weather,User location} & \(\varnothing\) \\
298:~\texttt{Local Search,Allergies} & \(\varnothing\) \\
299:~\texttt{Local Search,Food preferences} & \(\varnothing\) \\
300:~\texttt{Local Search,Hotel location} & \(\varnothing\) \\
301:~\texttt{Local Search,Income/budget considerations} & \(\varnothing\) \\
302:~\texttt{Local Search,Trip dates} & \(\varnothing\) \\
303:~\texttt{Local Search,Trip location} & \(\varnothing\) \\
304:~\texttt{Weather,Pet ownership status} & \(\varnothing\) \\
305:~\texttt{Weather,Scheduling details} & \(\varnothing\) \\
306:~\texttt{Weather,Trip dates} & \(\varnothing\) \\
307:~\texttt{Weather,Trip location} & \(\varnothing\) \\
308:~\texttt{Local Search,Email content} & \(\varnothing\) \\
309:~\texttt{Local Search,Trip activities} & \(\varnothing\) \\
310:~\texttt{Local Search,Trip dates} & \(\varnothing\) \\
311:~\texttt{Local Search,Trip location} & \(\varnothing\) \\
312:~\texttt{Weather,Trip dates} & \(\varnothing\) \\
313:~\texttt{Weather,Trip location} & \(\varnothing\) \\
314:~\texttt{Local Search,Address} & \(\varnothing\) \\
315:~\texttt{Local Search,Garage video recording} & \(\varnothing\) \\
316:~\texttt{Local Search,Motion detection logs} & \(\varnothing\) \\
317:~\texttt{Local Search,Pet ownership status (e.g. dog status)} & \(\varnothing\) \\
318:~\texttt{Cloud Drive,Daily activity events} & \(\varnothing\) \\
319:~\texttt{Cloud Drive,Gender} & \(\varnothing\) \\
320:~\texttt{Cloud Drive,Other travel permits} & \(\varnothing\) \\
321:~\texttt{Cloud Drive,Passport number} & \(\varnothing\) \\
322:~\texttt{Cloud Drive,Passport scans} & \(\varnothing\) \\
323:~\texttt{Cloud Drive,Past visa application details} & \(\varnothing\) \\
324:~\texttt{Cloud Drive,Travel history} & \(\varnothing\) \\
325:~\texttt{Cloud Drive,Visa documents} & \(\varnothing\) \\
326:~\texttt{Local Search,Car details (e.g. car fuel types)} & \(\varnothing\) \\
327:~\texttt{Local Search,Music playing history} & \(\varnothing\) \\
328:~\texttt{Local Search,User location} & \(\varnothing\) \\
329:~\texttt{Local Search,Visa documents} & \(\varnothing\) \\
330:~\texttt{Travel Booking,Address} & \(\varnothing\) \\
331:~\texttt{Travel Booking,Email address} & \(\varnothing\) \\
332:~\texttt{Travel Booking,Hotel preferences} & \(\varnothing\) \\
333:~\texttt{Travel Booking,Name} & \(\varnothing\) \\
334:~\texttt{Travel Booking,Payment method details} & \(\varnothing\) \\
335:~\texttt{Travel Booking,Phone number} & \(\varnothing\) \\
336:~\texttt{Travel Booking,Trip dates} & \(\varnothing\) \\
337:~\texttt{Travel Booking,Trip location} & \(\varnothing\) \\
338:~\texttt{Email,Email address} & \(\varnothing\) \\
339:~\texttt{Email,Film genre} & \(\varnothing\) \\
340:~\texttt{Email,Flight itinerary details (e.g. identification, flight number, departure and arrival time)} & \(\varnothing\) \\
341:~\texttt{Email,Passport} & \(\varnothing\) \\
342:~\texttt{mira,Flight itinerary details (e.g. identification, flight number, departure and arrival time)} & \(\varnothing\) \\
343:~\texttt{Travel Booking,Credit card details} & \(\varnothing\) \\
344:~\texttt{Travel Booking,Gender} & \(\varnothing\) \\
345:~\texttt{Travel Booking,Name} & \(\varnothing\) \\
346:~\texttt{Travel Booking,Parents' phone numbers} & \(\varnothing\) \\
347:~\texttt{Travel Booking,Passport number} & \(\varnothing\) \\
348:~\texttt{Travel Booking,Seating preference} & \(\varnothing\) \\
349:~\texttt{Travel Booking,Trip dates} & \(\varnothing\) \\
350:~\texttt{Travel Booking,Trip location} & \(\varnothing\) \\
351:~\texttt{Travel Booking,Visa scan} & \(\varnothing\) \\
352:~\texttt{Calendar,Car rental details (e.g. car type} & \(\varnothing\) \\
353:~\texttt{Calendar,Trip dates} & \(\varnothing\) \\
354:~\texttt{Calendar,Trip location} & \(\varnothing\) \\
355:~\texttt{Travel Booking,Car preferences} & \(\varnothing\) \\
356:~\texttt{Travel Booking,Driving license} & \(\varnothing\) \\
357:~\texttt{Travel Booking,Passport} & \(\varnothing\) \\
358:~\texttt{Travel Booking,Past visa application details} & \(\varnothing\) \\
359:~\texttt{Travel Booking,Pending task information} & \(\varnothing\) \\
360:~\texttt{Travel Booking,Trip dates} & \(\varnothing\) \\
361:~\texttt{Travel Booking,Trip location} & \(\varnothing\) \\
362:~\texttt{mira,Car rental details (e.g. car type} & \(\varnothing\) \\
363:~\texttt{The Movie Database,Email address} & \(\varnothing\) \\
364:~\texttt{The Movie Database,Film genre} & \(\varnothing\) \\
365:~\texttt{The Movie Database,Movie watching history} & \(\varnothing\) \\
366:~\texttt{The Movie Database,Parents' phone numbers} & \(\varnothing\) \\
367:~\texttt{The Movie Database,Film genre} & \(\varnothing\) \\
368:~\texttt{The Movie Database,Movie watching history} & \(\varnothing\) \\
369:~\texttt{The Movie Database,Music playing history} & \(\varnothing\) \\
370:~\texttt{The Movie Database,Name} & \(\varnothing\) \\
371:~\texttt{Apple Music,Education level} & \(\varnothing\) \\
372:~\texttt{Apple Music,Geolocation} & \(\varnothing\) \\
373:~\texttt{Apple Music,Movie watching history} & \(\varnothing\) \\
374:~\texttt{Apple Music,Music playing history} & \(\varnothing\) \\
375:~\texttt{Web,Children's phone numbers} & \(\varnothing\) \\
376:~\texttt{Web,Geolocation} & \(\varnothing\) \\
377:~\texttt{Web,Movie watching history} & \(\varnothing\) \\
378:~\texttt{Web,Profile Pictures} & \(\varnothing\) \\
379:~\texttt{SMS,Children's phone numbers} & \(\varnothing\) \\
380:~\texttt{SMS,Event details} & \(\varnothing\) \\
381:~\texttt{SMS,Message content} & \(\varnothing\) \\
382:~\texttt{SMS,Movie watching history} & \(\varnothing\) \\
383:~\texttt{SMS,Music playing history} & \(\varnothing\) \\
384:~\texttt{SMS,Parents' phone numbers} & \(\varnothing\) \\
385:~\texttt{SMS,Profile Pictures} & \(\varnothing\) \\
386:~\texttt{SMS,Siblings' phone numbers} & \(\varnothing\) \\
387:~\texttt{Cloud Drive,Photo details (e.g. file name, location, and timestamp)} & \(\varnothing\) \\
388:~\texttt{Instagram,Attendees email addresses} & \(\varnothing\) \\
389:~\texttt{Instagram,Colleague's email} & \(\varnothing\) \\
390:~\texttt{Instagram,Phone number} & \(\varnothing\) \\
391:~\texttt{Instagram,Photo details (e.g. file name, location, and timestamp)} & \(\varnothing\) \\
392:~\texttt{Instagram,Photos} & \(\varnothing\) \\
393:~\texttt{Instagram,Relationship status} & \(\varnothing\) \\
394:~\texttt{Instagram,User location} & \(\varnothing\) \\
395:~\texttt{mira,Instagram post content} & \(\varnothing\) \\
396:~\texttt{mira,Photos} & \(\varnothing\) \\
397:~\texttt{Tinder,Age} & \(\varnothing\) \\
398:~\texttt{Tinder,Birthday} & \(\varnothing\) \\
399:~\texttt{Tinder,Education level} & \(\varnothing\) \\
400:~\texttt{Tinder,Email address} & \(\varnothing\) \\
401:~\texttt{Tinder,Gender} & \(\varnothing\) \\
402:~\texttt{Tinder,Interests/hobbies} & \(\varnothing\) \\
403:~\texttt{Tinder,Phone number} & \(\varnothing\) \\
404:~\texttt{Tinder,Profile Pictures} & \(\varnothing\) \\
405:~\texttt{Tinder,Relationship preferences} & \(\varnothing\) \\
406:~\texttt{Tinder,User bio} & \(\varnothing\) \\
407:~\texttt{Tinder,User location} & \(\varnothing\) \\
408:~\texttt{Tinder,Workplace} & \(\varnothing\) \\
409:~\texttt{Cloud Drive,Birthday} & \(\varnothing\) \\
410:~\texttt{Cloud Drive,Heart rate data} & \(\varnothing\) \\
411:~\texttt{Cloud Drive,Medical/pregnancy history} & \(\varnothing\) \\
412:~\texttt{Cloud Drive,Photo details (e.g. file name, location, and timestamp)} & \(\varnothing\) \\
413:~\texttt{Tinder,Photos} & \(\varnothing\) \\
414:~\texttt{mira,Photos} & \(\varnothing\) \\
415:~\texttt{Tinder,Calories burned} & \(\varnothing\) \\
416:~\texttt{Tinder,Geolocation} & \(\varnothing\) \\
417:~\texttt{Tinder,Personal weight history} & \(\varnothing\) \\
418:~\texttt{Tinder,Siblings' phone numbers} & \(\varnothing\) \\
419:~\texttt{Tinder,Phone number} & \(\varnothing\) \\
420:~\texttt{Tinder,Relationship preferences} & \(\varnothing\) \\
421:~\texttt{Tinder,Relationship status (e.g. single, married)} & \(\varnothing\) \\
422:~\texttt{Tinder,User location} & \(\varnothing\) \\
423:~\texttt{mira,Backyard video recording} & \(\varnothing\) \\
424:~\texttt{mira,Garage access log} & \(\varnothing\) \\
425:~\texttt{mira,Garage video recording} & \(\varnothing\) \\
426:~\texttt{mira,Geolocation} & \(\varnothing\) \\
427:~\texttt{Lighting Control,Bedtime} & \(\varnothing\) \\
428:~\texttt{Lighting Control,Motion detection logs} & \(\varnothing\) \\
429:~\texttt{Lighting Control,Other travel permits} & \(\varnothing\) \\
430:~\texttt{Lighting Control,Past visa application details} & \(\varnothing\) \\
431:~\texttt{mira,Energy consumption logs} & \(\varnothing\) \\
432:~\texttt{mira,Home presence} & \(\varnothing\) \\
433:~\texttt{mira,Movie watching history} & \(\varnothing\) \\
434:~\texttt{mira,Music playing history} & \(\varnothing\) \\
435:~\texttt{mira,Doctor's Name} & \(\varnothing\) \\
436:~\texttt{mira,Front door video recording } & \(\varnothing\) \\
437:~\texttt{mira,Garage access log} & \(\varnothing\) \\
438:~\texttt{mira,Health conditions} & \(\varnothing\) \\
439:~\texttt{mira,Backyard video recording } & \(\varnothing\) \\
440:~\texttt{mira,Credit history} & \(\varnothing\) \\
441:~\texttt{mira,Filling status} & \(\varnothing\) \\
442:~\texttt{mira,Home presence} & \(\varnothing\) \\
443:~\texttt{mira,Front door video recording} & \(\varnothing\) \\
444:~\texttt{mira,Home presence} & \(\varnothing\) \\
445:~\texttt{mira,Invoice and payment timelines} & \(\varnothing\) \\
446:~\texttt{mira,Motion detection logs} & \(\varnothing\) \\
\end{longtable}}

\subsection{Generation prompts}
\label{app:generation-prompts}

\subsubsection{Prompt for the AI-Generated User Study }
\begin{Verbatim}[fontsize=\scriptsize,breaklines=true,breakanywhere=true,baselinestretch=0.95]
You are simulating a human subject study on AI-agent authorization preferences.

Your task is to generate synthetic users and simulate how they fill out the SAME sequential authorization questionnaire.

Every user receives the exact same questionnaire structure and the exact same question ordering.

Users differ ONLY in:
- profile attributes
- yes/no answers
- skipped questions
- stopping behavior

The authorization DAG structure itself NEVER changes.

The model must prioritize structural validity over behavioral realism.

If realism and structural validity conflict,
structural validity ALWAYS wins.

==================================================
IMPORTANT OUTPUT RULE
==================================================

Return ONLY ONE valid JSON ARRAY.

Do NOT include:
- markdown
- explanations
- comments
- analysis
- trailing commas

==================================================
QUESTION RESPONSE TYPES
==================================================

Every question has ONLY THREE possible responses:

[
  "yes",
  "no",
  "skip"
]

No other responses are allowed.

==================================================
CRITICAL DESIGN PRINCIPLE
==================================================

The authorization DAG topology is FIXED.

The following are FIXED across ALL users:
- node ids
- node names
- edges
- hierarchy
- cross-parent structure
- question order

Users differ ONLY in:
- permission answers
- skipped nodes
- stopping behavior

==================================================
CANONICAL NODE IDS
==================================================

Use ONLY the following node ids.

Never invent new node ids.

{
  "root": "Allow AI agent for requested task",

  "communication": "Communication",
  "email": "Email",
  "inbox_access": "Inbox Access",
  "read_inbox": "Read inbox",
  "search_email": "Search email",
  "drafting": "Drafting",
  "draft_reply": "Draft reply",
  "sending": "Sending",
  "send_email": "Send email",

  "planning": "Planning",
  "calendar": "Calendar",
  "event_access": "Event Access",
  "read_events": "Read events",
  "event_management": "Event Management",
  "create_event": "Create event",
  "modify_event": "Modify event",
  "cancel_event": "Cancel event",

  "contacts": "Contacts",
  "contact_access": "Contact Access",
  "read_contacts": "Read contacts",
  "email_contacts": "Email contacts",

  "travel_services": "Travel Services",
  "travel_websites": "Travel websites",
  "travel_search": "Search",
  "search_flights": "Search flights",
  "search_hotels": "Search hotels",
  "booking": "Booking",
  "reserve_itinerary": "Reserve itinerary",

  "financial_operations": "Financial Operations",
  "payment": "Payment",
  "card_access": "Card Access",
  "view_saved_cards": "View saved cards",
  "transactions": "Transactions",
  "make_payment": "Make payment",

  "local_file_system": "Local File System",
  "file_system": "File system",
  "file_access": "File Access",
  "read_selected_folder": "Read selected folder",
  "file_organization": "File Organization",
  "rename_files": "Rename files",
  "move_files": "Move files",
  "file_deletion": "File Deletion",
  "delete_files": "Delete files",

  "cloud_collaboration": "Cloud Collaboration",
  "cloud_storage": "Cloud storage",
  "cloud_access": "Cloud Access",
  "read_drive_files": "Read drive files",
  "upload_operations": "Upload Operations",
  "upload_files": "Upload files",
  "sharing": "Sharing",
  "share_files": "Share files",

  "development_workspace": "Development Workspace",
  "code_repository": "Code repository",
  "repository_access": "Repository Access",
  "read_code": "Read code",
  "branch_management": "Branch Management",
  "create_branch": "Create branch",
  "code_contribution": "Code Contribution",
  "commit_changes": "Commit changes",
  "open_pull_request": "Open pull request",
  "repository_administration": "Repository Administration",
  "merge_to_main": "Merge to main",

  "execution_environment": "Execution Environment",
  "terminal": "Terminal",
  "testing": "Testing",
  "run_tests": "Run tests",
  "package_management": "Package Management",
  "install_packages": "Install packages",
  "system_commands": "System Commands",
  "run_shell_commands": "Run shell commands",
  "destructive_commands": "Destructive Commands",

  "project_coordination": "Project Coordination",
  "issue_tracker": "Issue tracker",
  "issue_access": "Issue Access",
  "read_issues": "Read issues",
  "issue_management": "Issue Management",
  "create_issue": "Create issue",
  "close_issue": "Close issue",

  "financial_access": "Financial Access",
  "bank_account": "Bank account",
  "account_overview": "Account Overview",
  "view_balance": "View balance",
  "transaction_records": "Transaction Records",
  "view_transactions": "View transactions",
  "money_movement": "Money Movement",
  "transfer_money": "Transfer money",

  "investment_management": "Investment Management",
  "investment_account": "Investment account",
  "portfolio_access": "Portfolio Access",
  "view_portfolio": "View portfolio",
  "trading": "Trading",
  "place_trade": "Place trade",

  "billing": "Billing",
  "bills": "Bills",
  "bill_access": "Bill Access",
  "read_bills": "Read bills",
  "payment_processing": "Payment Processing",
  "pay_bills": "Pay bills",

  "healthcare_services": "Healthcare Services",
  "health_portal": "Health portal",
  "appointment_access": "Appointment Access",
  "read_appointments": "Read appointments",
  "medical_records": "Medical Records",
  "read_lab_results": "Read lab results",
  "provider_communication": "Provider Communication",
  "message_provider": "Message provider",

  "insurance": "Insurance",
  "insurance_portal": "Insurance Portal",
  "claim_access": "Claim Access",
  "view_claims": "View claims",
  "claim_operations": "Claim Operations",
  "submit_claim": "Submit claim",

  "research_workflow": "Research Workflow",
  "literature_search": "Literature search",
  "discovery": "Discovery",
  "search_papers": "Search papers",
  "research_material_access": "Research Material Access",
  "download_papers": "Download papers",
  "analysis": "Analysis",
  "summarize_papers": "Summarize papers",

  "research_workspace": "Research Workspace",
  "research_file_system": "Research File System",
  "project_access": "Project Access",
  "read_project_files": "Read project files",
  "research_notes": "Research Notes",
  "write_notes": "Write notes",

  "home_automation": "Home Automation",
  "smart_devices": "Smart devices",
  "device_monitoring": "Device Monitoring",
  "view_device_status": "View device status",
  "device_control": "Device Control",
  "turn_lights_on_off": "Turn lights on/off",
  "change_thermostat": "Change thermostat",
  "physical_security": "Physical Security",
  "unlock_door": "Unlock door",

  "monitoring": "Monitoring",
  "camera": "Camera",
  "live_monitoring": "Live Monitoring",
  "view_live_feed": "View live feed",
  "recording": "Recording",
  "record_video": "Record video",

  "location_services": "Location Services",
  "location": "Location",
  "presence_detection": "Presence Detection",
  "use_home_location": "Use home location",

  "social_platform": "Social Platform",
  "social_account": "Social account",
  "content_access": "Content Access",
  "read_posts": "Read posts",
  "content_drafting": "Content Drafting",
  "draft_post": "Draft post",
  "publishing": "Publishing",
  "publish_post": "Publish post",
  "content_removal": "Content Removal",
  "delete_post": "Delete post",

  "messaging": "Messaging",
  "messages": "Messages",
  "message_access": "Message Access",
  "read_messages": "Read messages",
  "communication_messages": "Communication",
  "send_messages": "Send messages",

  "media_management": "Media Management",
  "media_library": "Media library",
  "media_access": "Media Access",
  "read_photos": "Read photos",
  "upload_photos": "Upload photos"
}

==================================================
COMPLETE DAG HIERARCHY
==================================================

(INSERT THE COMPLETE HIERARCHY JSON HERE)

==================================================
FIXED CROSS-PARENT NODES
==================================================

ONLY the following nodes may have multiple parents.

{
  "make_payment": [
    "transactions",
    "booking"
  ],

  "delete_files": [
    "file_deletion",
    "destructive_commands"
  ],

  "read_events": [
    "event_access",
    "appointment_access"
  ],

  "use_home_location": [
    "presence_detection",
    "device_control"
  ],

  "upload_photos": [
    "upload_operations",
    "publishing"
  ]
}

No other node may have multiple parents.

==================================================
MULTIPLE-PARENT RULES
==================================================

Shared nodes use STRICT AND-gate semantics.

A shared node may be "yes"
ONLY IF:
- ALL parents appear
AND
- ALL parents are "yes"

If ANY parent is:
- "no"
- "skip"
- missing

then the shared node may NOT be "yes".

A shared node may appear ONLY ONCE.

==================================================
SKIP RULES
==================================================

If a node is answered:

"skip"

then:

1. ALL descendants are automatically skipped recursively.

2. Descendants must NOT appear in permissions.

3. Descendants must appear in missing_nodes.

4. This applies recursively to the entire subtree.

==================================================
NO RULES
==================================================

If a node is answered:

"no"

then descendants:
- may also be "no"
OR
- may be omitted entirely

BUT:

If ANY ancestor is:
- "no"
OR
- skipped

then descendants may NEVER be "yes".

Nodes blocked by "no" ancestors are NOT considered incomplete.

==================================================
EARLY STOPPING RULES
==================================================

Users may stop answering early.

If a user stops early:

1. ALL remaining unanswered nodes in the fixed questionnaire order are treated as skipped.

2. ALL remaining skipped nodes must appear in missing_nodes.

3. Remaining skipped nodes must NOT appear in permissions.

==================================================
INCOMPLETE DEFINITION
==================================================

A questionnaire is incomplete ONLY IF
at least one node is skipped.

This includes:
- explicitly skipped nodes
- descendants skipped by propagation
- nodes skipped after early stopping

Nodes omitted because of "no" ancestors
do NOT count as incomplete.

==================================================
QUESTION ORDERING
==================================================

Questions MUST follow preorder traversal.

Parents are ALWAYS answered before children.

Shared nodes may appear ONLY AFTER
all parents have already appeared.

==================================================
MINIMUM COMPLETION RULE
==================================================

Unless:
- privacy_concern == "high"
AND
- tech_savviness == "low"

each user should answer at least:
15 non-root nodes

before stopping early.

==================================================
REALISTIC HUMAN BEHAVIOR RULES
==================================================

1. High privacy users:
- more "no"
- more skips
- more early stopping

2. High trust users:
- more "yes"
- more complete questionnaires

3. High risk tolerance users:
- more dangerous permissions allowed

4. Low tech users:
- more skips
- more incomplete graphs

5. High time pressure users:
- broader permissions
- more skipping

==================================================
PROFILE CORRELATION RULES
==================================================

Profile variables should exhibit realistic correlations.

Examples:

- engineers:
  usually higher tech_savviness

- retired users:
  usually lower tech_savviness

- healthcare workers:
  usually higher privacy concern

- high privacy users:
  usually lower agent_trust

- fully_autonomous preference:
  usually associated with
  high trust and high prior_agent_use

==================================================
DANGEROUS ACTIONS
==================================================

[
  "transfer_money",
  "delete_files",
  "merge_to_main",
  "unlock_door",
  "run_shell_commands",
  "publish_post",
  "make_payment",
  "place_trade",
  "pay_bills",
  "cancel_event",
  "send_email",
  "send_messages",
  "submit_claim"
]

These should more frequently receive:
- "no"
OR
- "skip"

==================================================
PROFILE VARIABLES
==================================================

age_group:
["18-24","25-34","35-44","45-54","55+"]

occupation_type:
[
  "student",
  "researcher",
  "engineer",
  "manager",
  "freelancer",
  "healthcare",
  "education",
  "retired"
]

tech_savviness:
["low","medium","high"]

privacy_concern:
["low","medium","high"]

risk_tolerance:
["low","medium","high"]

agent_trust:
["low","medium","high"]

prior_agent_use:
["never","rarely","sometimes","often"]

requested_autonomy_level:
[
  "manual_approval",
  "semi_autonomous",
  "fully_autonomous"
]

time_pressure:
["low","medium","high"]

domain_sensitivity:
["low","medium","high"]

task_type:
[
  "email_scheduling",
  "travel_booking",
  "file_management",
  "software_development",
  "online_shopping",
  "finance_assistant",
  "health_admin",
  "academic_research",
  "smart_home_control",
  "social_media_management"
]

==================================================
OUTPUT FORMAT
==================================================

[
  {
    "user_id": "U0001",

    "X": {
      "user_id": "U0001",
      "age_group": "...",
      "occupation_type": "...",
      "tech_savviness": "...",
      "privacy_concern": "...",
      "risk_tolerance": "...",
      "agent_trust": "...",
      "prior_agent_use": "...",
      "task_type": "...",
      "requested_autonomy_level": "...",
      "time_pressure": "...",
      "domain_sensitivity": "..."
    },

    "Y": {
      "root": "root",

      "structure_type": "DAG",

      "incomplete": true,

      "permissions": {
        "root": "yes",
        "communication": "yes",
        "email": "yes",
        "inbox_access": "yes",
        "read_inbox": "yes",
        "search_email": "no"
      },

      "missing_nodes": [
        "send_email",
        "make_payment",
        "transfer_money"
      ],

      "cross_parent_nodes": [
        {
          "node_id": "make_payment",
          "parents": [
            "transactions",
            "booking"
          ]
        }
      ]
    }
  }
]

==================================================
FINAL VALIDATION RULES
==================================================

Before outputting final JSON,
perform an internal validation pass.

Reject and regenerate any user violating:
- DAG constraints
- ancestor consistency
- AND-gate semantics
- canonical node ids
- duplicate nodes
- preorder traversal
- invalid skips
- invalid incomplete labels
- malformed JSON

==================================================
FINAL HARD CONSTRAINTS
==================================================

1. permissions contains ONLY:
- "yes"
- "no"

2. skipped nodes NEVER appear in permissions.

3. skipped nodes ALWAYS appear in missing_nodes.

4. every node id must be canonical.

5. shared nodes appear ONLY ONCE.

6. descendants may NEVER be more permissive than ancestors.

7. no cycles.

8. incomplete = true
ONLY IF at least one node is skipped.

9. Output ONLY valid JSON.
\end{Verbatim}

\subsubsection{Prompt for the Real Data Analysis }

\begin{Verbatim}[fontsize=\scriptsize,breaklines=true,breakanywhere=true,baselinestretch=0.95]
You are conducting a one-off synthetic-data pilot for an AI-agent permission preference study.

The attached JSON contains exactly 51 de-identified REAL TRAINING participants, a catalog of 65 public task queries, and 373 permission-leaf definitions. It contains zero participants or labels from the held-out 150-person test set.

Create a downloadable ZIP containing all of the following files:

1. synthetic_500.json
2. generation_manifest.json
3. validation_report.json
4. generator.py (the exact executable code you used, if any)
5. README.md explaining the method and limitations

Requirements for synthetic_500.json:

- Generate exactly 500 synthetic participants with IDs S0001 through S0500.
- Preserve the seven profile fields and their allowed values from the 51 real examples.
- Preserve the four response categories exactly:
  - Yes, always share
  - Yes, but ask me next time
  - No, but ask me next time
  - No, never share
- Use only the attached 51-person training data to estimate profile distributions, query-selection/missingness patterns, response distributions, and relationships between profile, query, receiver/data type, and response.
- Do not use external datasets or assume access to any test participants.
- Do not copy a real participant verbatim. Each synthetic participant must be novel.
- Preserve realistic sparsity: each synthetic participant should answer a realistic subset of the 65 queries, following the query-count and missingness patterns observed among the 51 training participants.
- For every selected query, include decisions only for receiver_data entries belonging to that query in permission_catalog.
- Do not invent new queries, receiver_data strings, response labels, profile fields, or schema fields.
- Retain all four response categories; do not convert them to binary.
- Use a fixed random seed of 20260819 for every stochastic operation.

Method transparency:

- State explicitly whether each permission response was produced by language-model judgment, by Python/statistical sampling, or by a hybrid method.
- If Python or statistical sampling is used, include the complete generator.py and document every distribution, smoothing rule, dependency, and fallback.
- Do not describe Python-sampled labels as independent LLM judgments.
- Record the ChatGPT model/product shown in the interface, generation date, seed, input SHA-256 values from the attachment, and all assumptions in generation_manifest.json.

Validation requirements:

- Exactly 500 unique synthetic IDs.
- No original R001-R051 record may appear verbatim.
- No duplicate complete synthetic records.
- Every query_id, query string, and receiver_data must match the supplied catalogs.
- Every response must be one of the four allowed categories.
- Report profile distributions, response distributions, queries answered per participant, decisions per participant, per-query coverage, and per-permission coverage for both real and synthetic data.
- Report the nearest real participant for every synthetic participant and summarize similarity/duplication risk.
- Compare real versus synthetic marginal distributions and important conditional distributions.
- Clearly flag unsupported combinations, zero-count categories, sparse cells, and any rule violations.

Important scientific limitation:

This is synthetic augmentation data, not 500 independent human respondents. State that clearly in README.md. Do not claim that the generated records demonstrate real-world population validity.

Before finishing, run generator.py and the validation checks inside the data-analysis environment. Return the ZIP as a downloadable file; do not paste the 500 records into the chat response.
\end{Verbatim}

\clearpage
\bibliographystyle{agsm}  
\bibliography{references}  

@article{chow1968approximating,
  title={Approximating discrete probability distributions with dependence trees},
  author={Chow, CKCN and Liu, Cong},
  journal={IEEE transactions on Information Theory},
  volume={14},
  number={3},
  pages={462--467},
  year={1968},
  publisher={IEEE}
}

@article{Hoeffding1963,
 ISSN = {01621459, 1537274X},
 author = {Wassily Hoeffding},
 journal = {Journal of the American Statistical Association},
 number = {301},
 pages = {13--30},
 publisher = {[American Statistical Association, Taylor & Francis, Ltd.]},
 title = {Probability Inequalities for Sums of Bounded Random Variables},
 urldate = {2026-11-01},
 volume = {58},
 year = {1963}
}

@article{lafferty2001conditional,
  title={Conditional random fields: Probabilistic models for segmenting and labeling sequence data},
  author={Lafferty, John and McCallum, Andrew and Pereira, Fernando CN},
  year={2001}
}

@article{taskar2003max,
  title={Max-margin Markov networks},
  author={Taskar, Ben and Guestrin, Carlos and Koller, Daphne},
  journal={Advances in Neural Information Processing Systems},
  volume={16},
  year={2003}
}

@article{silla2011survey,
  title={A survey of hierarchical classification across different application domains},
  author={Silla Jr, Carlos N and Freitas, Alex A},
  journal={Data mining and knowledge discovery},
  volume={22},
  number={1},
  pages={31--72},
  year={2011},
  publisher={Springer}
}

@article{barbedo2006automatic,
  title={Automatic genre classification of musical signals},
  author={Barbedo, Jayme Garcia Arnal and Lopes, Amauri},
  journal={EURASIP Journal on Advances in Signal Processing},
  volume={2007},
  number={1},
  pages={064960},
  year={2006},
  publisher={Springer}
}

@inproceedings{fagni2007selection,
  title={On the selection of negative examples for hierarchical text categorization},
  author={Fagni, Tiziano and Sebastiani, Fabrizio},
  booktitle={Proceedings of the 3rd language technology conference},
  pages={24--28},
  year={2007}
}

@inproceedings{qiu2009hierarchical,
  title={Hierarchical multi-label text categorization with global margin maximization},
  author={Qiu, Xipeng and Gao, Wenjun and Huang, Xuan-Jing},
  booktitle={Proceedings of the acl-ijcnlp 2009 conference short papers},
  pages={165--168},
  year={2009}
}

@inproceedings{tai2015improved,
  title={Improved semantic representations from tree-structured long short-term memory networks},
  author={Tai, Kai Sheng and Socher, Richard and Manning, Christopher D},
  booktitle={Proceedings of the 53rd Annual Meeting of the Association for Computational Linguistics and the 7th International Joint Conference on Natural Language Processing (Volume 1: Long Papers)},
  pages={1556--1566},
  year={2015}
}

@inproceedings{socher2013recursive,
  title={Recursive deep models for semantic compositionality over a sentiment treebank},
  author={Socher, Richard and Perelygin, Alex and Wu, Jean and Chuang, Jason and Manning, Christopher D and Ng, Andrew Y and Potts, Christopher},
  booktitle={Proceedings of the 2013 conference on empirical methods in natural language processing},
  pages={1631--1642},
  year={2013}
}

@inproceedings{wehrmann2018hierarchical,
  title={Hierarchical multi-label classification networks},
  author={Wehrmann, Jonatas and Cerri, Ricardo and Barros, Rodrigo},
  booktitle={International Conference on Machine Learning},
  pages={5075--5084},
  year={2018},
  organization={PMLR}
}

@article{giunchiglia2020coherent,
  title={Coherent hierarchical multi-label classification networks},
  author={Giunchiglia, Eleonora and Lukasiewicz, Thomas},
  journal={Advances in neural information processing systems},
  volume={33},
  pages={9662--9673},
  year={2020}
}

@article{vaswani2017attention,
  title={Attention is all you need},
  author={Vaswani, Ashish and Shazeer, Noam and Parmar, Niki and Uszkoreit, Jakob and Jones, Llion and Gomez, Aidan N and Kaiser, {\L}ukasz and Polosukhin, Illia},
  journal={Advances in Neural Information Processing Systems},
  volume={30},
  year={2017}
}

@misc{martindaleMetaSecurityResearchers2026,
  title = {Meta {{Security Researcher}}'s {{AI Agent Accidentally Deleted Her Emails}}},
  author = {Martindale, Jon},
  year = 2026,
  month = feb,
  journal = {PCMAG},
  urldate = {2026-05-10},
  howpublished = {https://www.pcmag.com/news/meta-security-researchers-openclaw-ai-agent-accidentally-deleted-her-emails},
  langid = {english}
}

@misc{mchughWhosReallyShopping2026,
  title = {Who's {{Really Shopping}}? {{Retail Fraud}} in the {{Age}} of {{Agentic AI}}},
  shorttitle = {Who's {{Really Shopping}}?},
  author = {McHugh, Christa, Matt Brady},
  year = 2026,
  month = mar,
  journal = {Unit 42},
  urldate = {2026-05-10},
  langid = {american},
  howpublished = {https://unit42.paloaltonetworks.com/retail-fraud-agentic-ai/}
}

@misc{GooglesAIpoweredAntigravity2025,
  title = {Google's {{AI-powered Antigravity IDE}} Has Some Worrying Security Issues},
  author = {Udinmwen, Efosa},
  year = 2025,
  month = dec,
  journal = {TechRadar},
  urldate = {2026-05-10},
  chapter = {Security},
  howpublished = {https://www.techradar.com/pro/googles-ai-powered-antigravity-ide-already-has-some-worrying-security-issues},
  langid = {english}
}

@misc{liuYourAICoding2026,
  title = {Your {{AI Coding Assistant Has}} the {{Keys}} to {{Your Machine}}},
  author = {Liu, Yanli},
  year = 2026,
  month = feb,
  journal = {Data Science Collective},
  urldate = {2026-05-10},
  langid = {english},
  howpublished = {https://medium.com/data-science-collective/your-ai-coding-assistant-has-the-keys-to-your-machine-a3727b70f015}
}

@article{liu2026youraishelldemystifying,
  title={"Your AI, My Shell": Demystifying Prompt Injection Attacks on Agentic AI Coding Editors},
  author={Liu, Yue and Zhao, Yanjie and Lyu, Yunbo and Zhang, Ting and Wang, Haoyu and Lo, David},
  journal={arXiv preprint arXiv:2509.22040},
  year={2025}
}

@misc{mehtaEvenBestAI2026,
  title = {Even the {{Best AI Agents Leak Secrets}}. {{Prompt Injection Is Why}}.},
  author = {Mehta, Shreyans},
  year = 2026,
  month = apr,
  journal = {Cequence Security},
  urldate = {2026-05-10},
  langid = {american},
  howpublished = {https://securityboulevard.com/2026/04/even-the-best-ai-agents-leak-secrets-prompt-injection-is-why/}
}

@article{read2021classifier,
  title={Classifier chains: A review and perspectives},
  author={Read, Jesse and Pfahringer, Bernhard and Holmes, Geoffrey and Frank, Eibe},
  journal={Journal of Artificial Intelligence Research},
  volume={70},
  pages={683--718},
  year={2021}
}

@article{BOUTELL20041757,
  title = {Learning Multi-Label Scene Classification},
  author = {Boutell, Matthew R. and Luo, Jiebo and Shen, Xipeng and Brown, Christopher M.},
  year = 2004,
  journal = {Pattern Recognition},
  volume = {37},
  number = {9},
  pages = {1757--1771},
  issn = {0031-3203},
  doi = {10.1016/j.patcog.2004.03.009}
}

@article{sucar2014multi,
  title={Multi-label classification with Bayesian network-based chain classifiers},
  author={Sucar, L Enrique and Bielza, Concha and Morales, Eduardo F and Hernandez-Leal, Pablo and Zaragoza, Julio H and Larra{\~n}aga, Pedro},
  journal={Pattern Recognition Letters},
  volume={41},
  pages={14--22},
  year={2014},
  publisher={Elsevier}
}

@inproceedings{zhang2010multi,
  title={Multi-label learning by exploiting label dependency},
  author={Zhang, Min-Ling and Zhang, Kun},
  booktitle={Proceedings of the 16th ACM SIGKDD international conference on Knowledge discovery and data mining},
  pages={999--1008},
  year={2010}
}

@article{kingma2014adam,
  title={Adam: A method for stochastic optimization},
  author={Kingma, Diederik P and Ba, Jimmy},
  journal={arXiv preprint arXiv:1412.6980},
  year={2014}
}

@inproceedings{wu2026towards,
  title={Towards automating data access permissions in ai agents},
  author={Wu, Yuhao and Yang, Ke and Roesner, Franziska and Kohno, Tadayoshi and Zhang, Ning and Iqbal, Umar},
  booktitle={2026 IEEE Symposium on Security and Privacy (SP)},
  pages={2609--2627},
  year={2026},
  organization={IEEE}
}

@article{shi2026progentsecuringaiagents,
  title={Progent: Securing AI agents with privilege control},
  author={Shi, Tianneng and He, Jingxuan and Wang, Zhun and Li, Hongwei and Wu, Linyu and Guo, Wenbo and Song, Dawn},
  journal={arXiv preprint arXiv:2504.11703},
  year={2025}
}

@inproceedings{Wang_2026, series={ICSE ’26},
   title={AgentSpec: Customizable Runtime Enforcement for Safe and Reliable LLM Agents},
   DOI={10.1145/3744916.3764546},
   booktitle={Proceedings of the 2026 IEEE/ACM 48th International Conference on Software Engineering},
   publisher={ACM},
   author={Wang, Haoyu and Poskitt, Christopher M. and Sun, Jun},
   year={2026},
   month=Apr, pages={2938–2950},
   collection={ICSE ’26} }

@article{liang2025convergence,
  title={Convergence of Adam for non-convex objectives: relaxed hyperparameters and non-ergodic case},
  author={Liang, Yuqing and He, Meixuan and Liu, Jinlan and Xu, Dongpo},
  journal={Machine Learning},
  volume={114},
  number={3},
  pages={75},
  year={2025},
  publisher={Springer}
}

@article{wang2009large,
  title={On large margin hierarchical classification with multiple paths},
  author={Wang, Junhui and Shen, Xiaotong and Pan, Wei},
  journal={Journal of the American Statistical Association},
  volume={104},
  number={487},
  pages={1213--1223},
  year={2009},
  publisher={Taylor \& Francis}
}

@ARTICLE{7118216,
  author={Bi, Wei and Kwok, Jame T.},
  journal={IEEE Transactions on Knowledge and Data Engineering}, 
  title={Bayes-Optimal Hierarchical Multilabel Classification}, 
  year={2015},
  volume={27},
  number={11},
  pages={2907-2918},
  doi={10.1109/TKDE.2015.2441707}}

@article{rousu2006kernel,
  title={Kernel-based learning of hierarchical multilabel classification models},
  author={Rousu, Juho and Saunders, Craig and Szedmak, Sandor and Shawe-Taylor, John},
  journal={Journal of Machine Learning Research},
  volume={7},
  pages={1601--1626},
  year={2006}
}

@ARTICLE{8933476,
  author={Peng, Hao and Li, Jianxin and Wang, Senzhang and Wang, Lihong and Gong, Qiran and Yang, Renyu and Li, Bo and Yu, Philip S. and He, Lifang},
  journal={IEEE Transactions on Knowledge and Data Engineering}, 
  title={Hierarchical Taxonomy-Aware and Attentional Graph Capsule RCNNs for Large-Scale Multi-Label Text Classification}, 
  year={2021},
  volume={33},
  number={6},
  pages={2505-2519},
  doi={10.1109/TKDE.2019.2959991}}

@ARTICLE{9568738,
  author={Liu, Weiwei and Wang, Haobo and Shen, Xiaobo and Tsang, Ivor W.},
  journal={IEEE Transactions on Pattern Analysis and Machine Intelligence}, 
  title={The Emerging Trends of Multi-Label Learning}, 
  year={2022},
  volume={44},
  number={11},
  pages={7955-7974},
  doi={10.1109/TPAMI.2021.3119334}}

@ARTICLE{5128907,
  author={He, Haibo and Garcia, Edwardo A.},
  journal={IEEE Transactions on Knowledge and Data Engineering}, 
  title={Learning from Imbalanced Data}, 
  year={2009},
  volume={21},
  number={9},
  pages={1263-1284},
  doi={10.1109/TKDE.2008.239}}

@article{dawid1979conditional,
  title={Conditional independence in statistical theory},
  author={Dawid, A Philip},
  journal={Journal of the Royal Statistical Society Series B: Statistical Methodology},
  volume={41},
  number={1},
  pages={1--15},
  year={1979},
  publisher={Oxford University Press}
}

@article{rubin1976inference,
  title={Inference and missing data},
  author={Rubin, Donald B},
  journal={Biometrika},
  volume={63},
  number={3},
  pages={581--592},
  year={1976},
  publisher={Oxford University Press}
}

@article{cox1958regression,
  title={The regression analysis of binary sequences},
  author={Cox, David R},
  journal={Journal of the Royal Statistical Society Series B: Statistical Methodology},
  volume={20},
  number={2},
  pages={215--232},
  year={1958},
  publisher={Oxford University Press}
}

@article{gneiting2007strictly,
  title={Strictly proper scoring rules, prediction, and estimation},
  author={Gneiting, Tilmann and Raftery, Adrian E},
  journal={Journal of the American statistical Association},
  volume={102},
  number={477},
  pages={359--378},
  year={2007},
  publisher={Taylor \& Francis}
}

@article{mohan2021graphical,
  title={Graphical models for processing missing data},
  author={Mohan, Karthika and Pearl, Judea},
  journal={Journal of the American Statistical Association},
  volume={116},
  number={534},
  pages={1023--1037},
  year={2021},
  publisher={Taylor \& Francis}
}

@article{reid2010composite,
  title={Composite binary losses},
  author={Reid, Mark D and Williamson, Robert C},
  journal={The Journal of Machine Learning Research},
  volume={11},
  pages={2387--2422},
  year={2010},
  publisher={JMLR. org}
}

@misc{AnthropicSaysIts2026,
  title = {Anthropic Says Its {{AI}} Model Submitted a False Cold Case Tip to {{Philadelphia}} Police},
  author= {Li, David},
  year = 2026,
  month = oct,
  journal = {NBC_News},
  urldate = {2026-11-12},
  howpublished = {https://www.nbcnews.com/news/us-news/anthropic-says-model-submitted-false-cold-case-tip-philadelphia-police-rcna603213},
  langid = {english},
}






\end{document}